\documentclass{article}
\usepackage{salesforce,times}

\usepackage{amsmath,amssymb}
\usepackage{booktabs}
\usepackage{multirow}
\usepackage{graphicx}
\usepackage{wrapfig}
\usepackage{xcolor}
\usepackage{siunitx}
\usepackage{subcaption}
\usepackage{hyperref}
\usepackage{url}
\usepackage{float}
\usepackage{enumitem}
\usepackage{listings}
\usepackage[breakable,skins,most]{tcolorbox}
\newif\ifarxiv                                % \arxivtrue for tight title padding
\newcommand{\promptleft}{4pt}
\newcommand{\promptright}{4pt}
\definecolor{mylightgray}{RGB}{240,240,240}
\usepackage[most]{tcolorbox}
\newtcolorbox{guidance}{
  colback=mylightgray,
  colframe=mylightgray,
  boxrule=0pt,
  arc=3pt,
  left=6pt, right=6pt, top=4pt, bottom=4pt,
  before skip=4pt, after skip=4pt,
}
\DeclareSIUnit{\flop}{FLOP}
\DeclareSIUnit{\gpu}{GPU}
\DeclareSIUnit{\minute}{min}
\newcommand{\pct}[1]{\SI[quantity-product={}]{#1}{\percent}}

\newcommand{\sflogo}{\raisebox{-0.15em}{\includegraphics[height=1.0em]{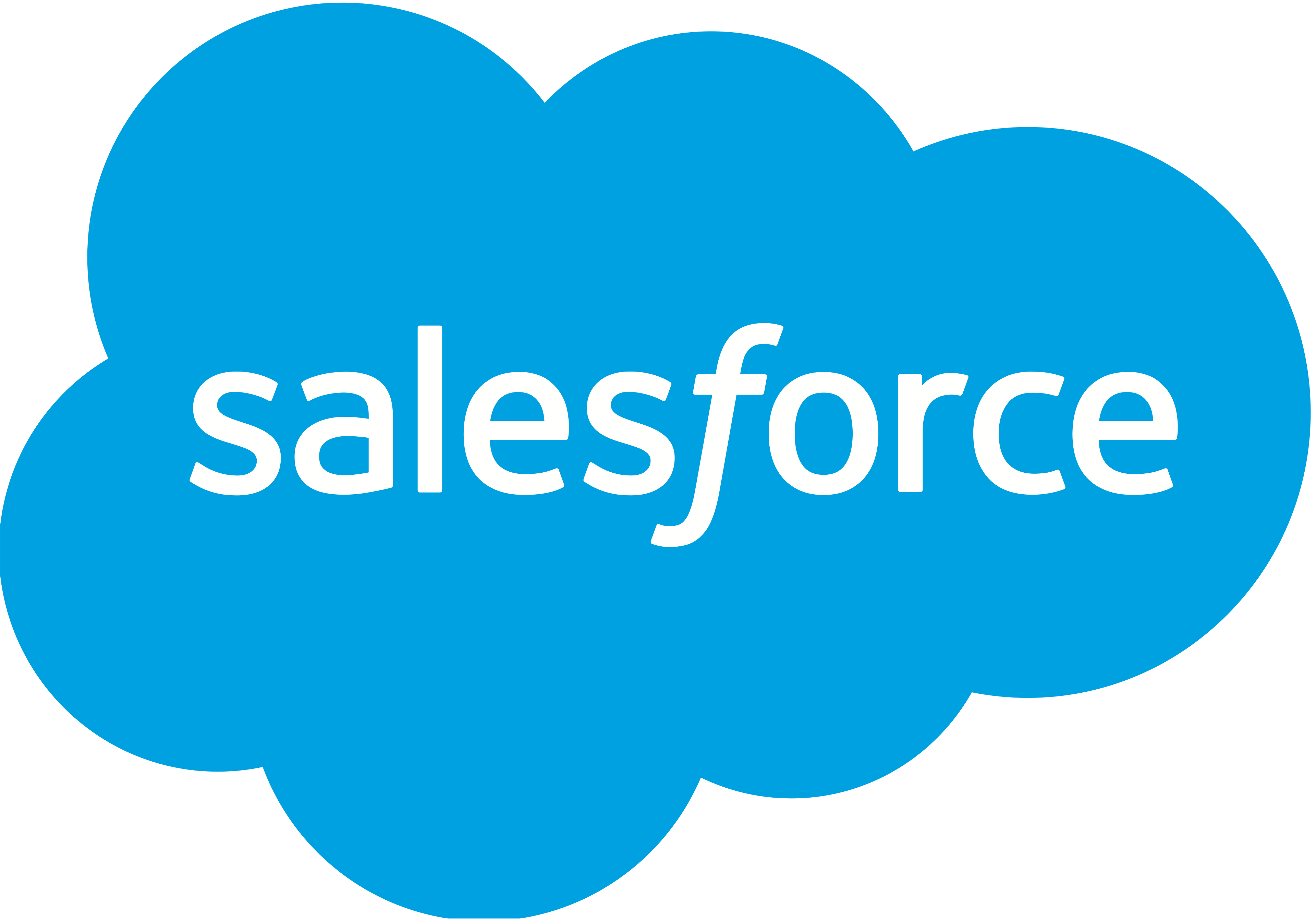}}}
\newcommand{\uiuclogo}{\raisebox{-0.15em}{\includegraphics[height=1.0em]{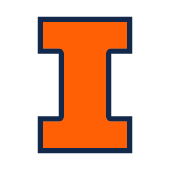}}}

\newcommand{\opd}{OPD}
\newcommand{\best}[1]{\textbf{#1}}

\title{Unpacking the Compute Efficiency of Strong-to-Weak \opd{} against GRPO}

\author{
Zeyu Leo Liu$^{1}$, Yang Li$^{1}$, Bo Pang$^{1}$, Emre Can Acikgoz$^{1,2}$, Xuan-Phi Nguyen$^{1}$, Anurag Koul$^{1}$, Srijan Bansal$^{1}$, Erik Nijkamp$^{1}$, Shafiq Joty$^{1}$, Semih Yavuz$^{1}$
\\
$^{1}$ \sflogo\,Salesforce AI Research, $^{2}$\uiuclogo UIUC
}

\begin{document}
\maketitle

\begin{abstract}

On-policy distillation (\opd{}) trains a student on its own rollouts under a frozen teacher's per-token supervision.
Does this dense signal reduce the training FLOPs needed to reach a target accuracy compared with reinforcement learning?
We systematically compare vanilla reverse-KL \opd{} with Group Relative Policy Optimization (GRPO) across task difficulty, student and teacher size, domain, and optimization choices, measuring held-out accuracy against training FLOPs.
With an external post-trained teacher, \opd{} generally loses to GRPO in accuracy per training FLOP at matched settings, even when the teacher is further trained with GRPO, and often suffers response-length inflation.
Yet a different teacher changes the picture: when the teacher is the student's own GRPO checkpoint, \opd{} recovers that checkpoint's accuracy on math with \num{50}$\times$ fewer distillation FLOPs than GRPO needs to reach the same accuracy.
Following the same teacher setting, we consistently observe this high efficiency with a larger student, on code, and in a second model family.
On a single domain, however, the GRPO checkpoint is already the desired model, so distilling it adds to total compute.
Across domains, separate GRPO checkpoints hold different capabilities, and no single checkpoint serves every domain.
Multi-teacher \opd{}~\citep{deepseek2026v4} consolidates per-domain GRPO experts into a single student that matches each expert's accuracy, with the distillation stage costing only \pct{5.2} (on 2 domains) and \pct{8.0} (on 4 domains) of the specialist training FLOPs.
Counting specialist training and distillation together, this pipeline reaches joint GRPO's performance on the 2-domain mix with \pct{42.0} fewer total FLOPs.

\end{abstract}

\begin{figure}[H]
\centering
% \begin{subfigure}[t]{0.32\textwidth}
%   \centering
%   \includegraphics[width=\textwidth]{figures/fig1_panel_a.pdf}
% \end{subfigure}\hfill
% \begin{subfigure}[t]{0.32\textwidth}
%   \centering
%   \includegraphics[width=\textwidth]{figures/fig1_panel_b.pdf}
% \end{subfigure}\hfill
% \begin{subfigure}[t]{0.32\textwidth}
%   \centering
%   \includegraphics[width=\textwidth]{figures/fig1_panel_c.pdf}
% \end{subfigure}
\includegraphics[width=\textwidth]{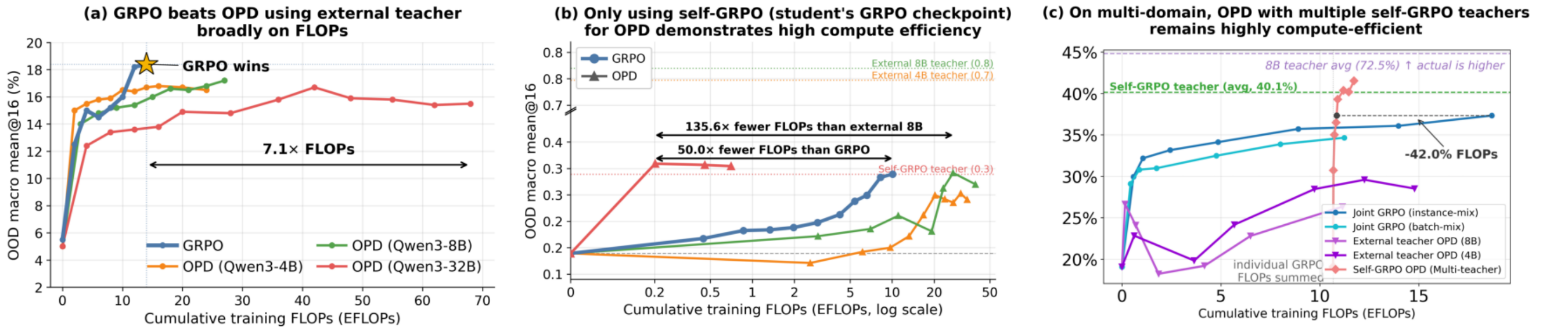}
\caption{\textbf{\opd{} compute efficiency, at a glance.}
(a) GRPO achieves higher performance at fewer FLOPs than \opd{} across external teacher sizes (up to 32B) on the OOD math evaluation (AIME24/25 and AMC23); the same ordering holds across the other settings.
(b) On the OOD math eval with a \texttt{Qwen3-4B-Base} student,
\opd{} with a \emph{self-GRPO teacher} (a GRPO checkpoint of the student) is
the only setup that matches GRPO performance with \num{50.0}$\times$ fewer FLOPs than GRPO, and \num{135.6}$\times$ fewer FLOPs than \opd{} with an external teacher \texttt{Qwen3-8B}.
(c) On a multi-domain data mix with math and code, running multi-teacher \opd{} with self-GRPO teachers (per domain) matches joint GRPO at \pct{42.0} fewer total FLOPs. A single external
\num{8}B teacher substantially underperforms despite its stronger overall performance; the trained student demonstrates severe degenerate behavior like unstoppable repetition, and Chinese characters given an English prompt.
}
\label{fig:teaser}
\end{figure}

% \yli{Overall comments: the content is a bit sparse to fill up 9 pages. we can have more description, analysis, discussion, takeaway etc}

\section{Introduction}

On-policy distillation (\opd{})~\citep{agarwal2024gkd} trains a student on
its own rollouts, with each token scored by a frozen teacher. It combines
the dense per-token supervision of classical knowledge
distillation~\citep{kim2016seqkd,gu2024minillm} with on-policy sampling
that avoids the train/test mismatch of teacher-generated data. 
% Because every token carries gradient rather than one scalar per trajectory, the
% objective is intuitively more sample-efficient than outcome-reward RL
% such as GRPO~\citep{shao2024deepseekmath}. 
Dense teacher feedback suggests that \opd{} may learn from fewer rollouts than outcome-reward RL such as GRPO~\citep{shao2024deepseekmath}. 
Its promise is sharpest on hard problems: when the student rarely finds a correct answer, outcome-reward GRPO has little positive signal, while the teacher can still supervise every generated token.
% This intuition underlies a widely repeated claim about \emph{compute} efficiency: the Qwen3 technical report describes reaching a higher AIME'24 score at roughly one tenth the GPU hours of its RL pipeline~\citep{qwen3}---but GPU hours are an implementation-dependent metric. 
% Yet \opd{} adds a frozen-teacher forward pass on every training token, and longer responses can further increase its cost. 
Yet that supervision requires a teacher forward pass on every training token, and longer responses raise the cost further.
Fewer rollouts or optimizer steps therefore need not mean fewer FLOPs. 
% Another widely-cited industry blog post~\citep{tml2025opd} asserts that \opd{} recovers an RL-trained policy ``in approximately 7--10$\times$ fewer gradient steps, which corresponds to a compute efficiency of 50--100$\times$.'' 
% However, it provides only partial FLOPs-based comparisons and specifies neither the RL baseline nor its hyperparameters, making the ratio hard to verify.
% The practical question is when, and with which teacher, \opd{} reaches a target accuracy with less training compute than GRPO. 
\textbf{Can \opd{} reach the same accuracy as GRPO with fewer training FLOPs?}
Existing efficiency reports use GPU-hours, gradient-step ratios, or partial FLOPs comparisons, leaving the matched-FLOPs comparison unresolved.
% Prominent efficiency claims use GPU-hours or gradient-step ratios, sometimes alongside partial FLOPs comparisons~\citep{qwen3, tml2025opd}.
% \yli{ Our reproduction on the same \textbf{tinker} library does not
% replicate the claimed ordering: at matched settings, GRPO reaches an
% AIME'24 accuracy that \opd{} does not
% approach (\S\ref{app:framework-detail}).
% }
To our knowledge, no prior work systematically compares \opd{} and GRPO
at matched training FLOPs (\S\ref{sec:related-work}; Appendix~\ref{app:audit}).

% To systematically compare the two methods, we sweep nine experimental axes
% that could plausibly affect the compute-efficiency ordering: dataset
% difficulty, student and teacher sizes, training domain, batch size,
% initialization, thinking mode, learning rate, reverse-KL variant, and
% RL framework (\S\ref{sec:main}--\S\ref{sec:ablations}), measuring accuracy
% against training FLOPs. 
In this work, we focus on systematically comparing reverse-KL \opd{} with GRPO by tracking held-out task performance against cumulative training FLOPs.
Our experiments vary data difficulty, student and teacher sizes and initializations, training domain, thinking mode, distillation objective, and the RL framework.
With off-the-shelf post-trained teachers, GRPO generally achieves higher accuracy at a fixed FLOP budget. 
Reducing \opd{} to one rollout per prompt is the one change that lets an external teacher outperform GRPO, and only on single-domain math; the advantage disappears on code and multi-domain training (Appendix~\ref{app:n1-defense}).
% Across all nine axes, GRPO broadly outperforms
% \opd{} on both compute efficiency and best performance with any same-family
% larger teacher. 
To our surprise, this even holds in the subset that should
intuitively favour \opd{} most: a small student that barely solves
(\texttt{mean@32}${=}0$) paired with a large teacher that solves well
(\texttt{mean@32}${>}0.6$), where GRPO's outcome reward is very weak and
dense per-token supervision should matter most (\S\ref{sec:stratification}).
% This suggests the bottleneck for \opd{} with an external teacher is not hyperparameters but the training signal itself.
External-teacher \opd{} consistently inflates response length until it reaches the maximum length, increasing the FLOPs paid per step (Appendix~\ref{app:why-detail}). 

We next ask whether the teacher's training history changes the result.
Recent work reports that GRPO-training the teacher before distillation can
help the student learn better~\citep{li2026a260413016}.
% , but bases this claim on a limited set of teacher choices. 
% To understand this more broadly, we systematically sweep the teacher type across external thinking models, thinking models further GRPO-trained, and GRPO-trained base models at or above the student's size (\S\ref{sec:crossover}). 
Building on this evidence, we compare external thinking models, further-GRPO-trained thinking models, and GRPO-trained base models at or above the student's size.
% This reveals a sharper observation: \opd{} fully recovers the teacher's accuracy only when the teacher is a GRPO checkpoint of the student, and does so at \num{50.0}$\times$ fewer FLOPs than GRPO requires to reach the same accuracy. 
Among the teachers tested, \opd{} consistently recovers the teacher's accuracy only when the teacher is a GRPO checkpoint of the student.
In particular, in our math setting, \opd{} with one rollout per prompt matches that checkpoint's accuracy with \num{50}$\times$ fewer distillation-stage FLOPs (\num{6}$\times$ at the default 8-rollout-per-prompt setting) than the eight-rollout GRPO training needs to reach the same accuracy. 
We further find such observation generalizes across model sizes, domains, and model families.
When using \texttt{Olmo-3-7B-Instruct-SFT} on math, we are surprised to find self-GRPO OPD can lead to a student outperforming teacher substantially by 11\% absolute score.
% We further find that the initial top-\num{16} token overlap between student and teacher, measured on the first batch, strongly correlates with how much of the teacher's performance the student recovers.

However, distilling from the student's own GRPO checkpoint reveals a
paradox: on a single domain, why distill when you already have the
GRPO-trained model? 
The \num{50}$\times$ ratio counts only distillation: the GRPO run has already produced the desired single domain model, so distilling it back into the base student adds cost.
% It resolves in the multi-domain setting adopted by
% recent frontier models~\citep{deepseek2026v4,kimi2026a260724653}: from the
% same base model, one trains a GRPO expert per domain independently, then
% multi-teacher \opd{} distills all experts into a single student, routing
% each domain's tokens to its expert. 
Consider math and code: separate GRPO runs produce two specialists, but neither checkpoint can serve both domains.
This is where multi-teacher \opd{} has a distinct compute-efficiency opportunity: it can consolidate the GRPO specialists into the multi-domain model neither provides, with only a small additional distillation cost.
The decisive compute test is whether the cost of training those experts and consolidating them via distillation beats that of joint GRPO on the same data mix.
Recent open-source models~\citep{yang2026a260319220, coreteam2026mimov2flashtechnicalreport, deepseek2026v4,kimi2026a260724653} use this recipe, but leave that full-pipeline comparison unmeasured.
% We test two specialist case, then extend to four domains.
% The single-domain efficiency carries over: multi-teacher \opd{} recovers all experts' performance at only $\pct{5.2}$ (on 2 domains) and $\pct{8.0}$ (on 4 domains) of the specialist training FLOPs, a level neither joint GRPO nor \opd{} with a larger external teacher reaches on the same data mix (\S\ref{sec:merging}). 
Multi-teacher \opd{} combines the per-domain GRPO experts into a single student model that recovers their performance, with distillation adding only $\pct{5.2}$ (two domains) and $\pct{8.0}$ (four domains) to the specialists' training FLOPs.
Counting both stages, the two-domain pipeline reaches joint GRPO's accuracy on the same data mix with $\pct{42}$ fewer total FLOPs.
On both two-domain and four-domain settings, joint GRPO does not even reach the merged student's performance within the tested budget.

Our contributions identify where \opd{} falls short of GRPO and how distillation gains a distinct compute advantage with multi-teacher \opd{}:
% Our analysis thus gives a compute-efficiency rationale for the choice made in recent frontier reports~\citep{deepseek2026v4,kimi2026a260724653,coreteam2026mimov2flashtechnicalreport}, which describe the recipe without matched-compute measurement. Our
% contributions:

%%%%% OLD contributions (for records in case needed)
% \item A nine-axis ablation showing that no external teacher makes vanilla reverse-KL \opd{} compute-efficient against GRPO, with   GRPO even broadly outperforming reverse-KL \opd{} (\S\ref{sec:main}).
% \item A characterization of the highly compute-efficient regime of strong-to-weak \opd{}: only when the teacher is a GRPO checkpoint of the student does \opd{} achieve \num{50.0}$\times$ compute efficiency over GRPO, and it achieves 135.6$\times$ over \opd{} with an external teacher (\S\ref{sec:crossover}).
% \item A compute-efficiency rationale for the multi-teacher \opd{} recipe that frontier reports~\citep{deepseek2026v4,kimi2026a260724653,coreteam2026mimov2flashtechnicalreport} adopt without justification: distilling per-domain GRPO experts into a single student recovers all expert performances at an additional $\pct{5.2}$ (on 2 domains) and $\pct{8.0}$  (on 4 domains) of the specialist training FLOPs. Its performance is not reached by either joint GRPO or \opd{} with a larger external teacher (\S\ref{sec:merging}).
%%%%% OLD contributions (for records in case needed)

\begin{itemize}[leftmargin=*,itemsep=0pt,topsep=0pt]
  \item A comprehensive set of controlled experiments shows that dense token supervision alone does not reliably make external-teacher \opd{} cheaper than GRPO (\S\ref{sec:main}).
  \item Choose teachers by provenance. Among the teachers tested, a GRPO checkpoint of the student model itself transfers most reliably and \num{50.0}$\times$ more efficiently, but its distillation-stage savings must be weighed against the cost of training it  (\S\ref{sec:crossover}).
  \item Specialize, then consolidate. Train separate GRPO experts on different domains and combine them with routed multi-teacher \opd{}, in which scenario \opd{} clearly offers a more compute-efficient solution than joint GRPO by producing the multi-domain model that no single specialist alone supplies (\S\ref{sec:merging}).
\end{itemize}

\section{Related work}
\label{sec:related-work}

\paragraph{Recent \opd{} work rarely accounts for FLOPs when claiming advantage over GRPO.}
Many recent methods modify the canonical reverse-KL objective for external-teacher \opd{} to outperform GRPO,
but none evaluate their claimed advantage over outcome-reward methods like GRPO in terms of
FLOPs (Appendix~\ref{app:audit}). They report sample efficiency~\citep{ko2026a260311137}, GPU-hours, or
wall-clock~\citep{zhang2026a260510189, feng2026a260705394, chen2026a260517862, rang2026a260505940};
\citet{zhao2026decouplingkltrajectoriesunified} reports FLOPs but only among \opd{} variants,
not against GRPO. Reported wall-clock or GPU-hour wins over GRPO also use different group sizes
for the two methods~\citep{zheng2026a260410688, tian2026vicurvisualcuesrecoverable} or leave it
unreported~\citep{rang2026a260505940, zhang2026a260510189, chen2026a260517862}, so the gap is not
attributable to objective quality alone. 
We isolate this confound by controlling group size and
characterizing external-teacher \opd{} through FLOPs, with a smaller-group-size ablation in
Appendix~\ref{app:n1-defense}.

\paragraph{Negative evidence and behavior of \opd{}}
Several concurrent studies report evidence against a plain \opd{} compute win.
\citet{lee2026a260618216} find \opd{} slower than GRPO in wall-clock time;
\citet{liu2026a260513230} report external-teacher reverse-KL \opd{} undergoing response-length
explosion while a reward-only GRPO baseline stays stable; and \citet{yang2026a260717558} find
\opd{} improves the base student but stays below the GRPO-trained
student, with \opd{} \emph{after} GRPO lowering the average. Further studies document problems
with external-teacher \opd{}---diversity collapse, prefix and support drift, low-information
suffixes, length inflation~\citep{kim2026a260324472, fu2026a260325562, luo2026a260408527,
wang2026a260410674, li2026a260413016, wang2026a260424005, yang2026a260507804, zhu2026a260511182,
yang2026a260606021, xin2026a260609471, ding2026a260612634, nicolicioiu2026a260626091,
wang2026a260717247}.

\paragraph{Concurrent \opd{} analyses also omit a compute-efficiency comparison with GRPO.}
\citet{li2026a260413016} find a small same-family RL-trained teacher yields consistent gains
where a stronger, more divergent one does not, but without a GRPO baseline;
\citet{luo2026a260408527, zhu2026facesonpolicydistillationpitfalls, fu2026rethinkingonpolicydistillationlarge}
likewise do not consider efficiency. In contrast, we compare GRPO and conventional \opd{} in a
controlled setting using FLOPs as the compute measure.  In Appendix~\ref{app:audit}, we conduct a broader, LLM-assisted survey to look for FLOPs comparison but find no qualifying work.

\paragraph{Multi-teacher \opd{} in frontier model reports.}
Several recent frontier-model reports use multi-teacher \opd{} to combine domain-expert teachers
into a shared student, e.g., DeepSeek-V4 and Kimi K3~\citep{yang2026a260319220, coreteam2026mimov2flashtechnicalreport, deepseek2026v4,kimi2026a260724653}. They present the recipe
as an engineering choice; we provide a rationale for it (\S\ref{sec:merging}).

\section{Experimental setup and compute accounting}
\label{sec:setup}

\paragraph{Comparison and compute accounting.} 
We compare held-out task performance against cumulative training FLOPs in our analysis. 
The main external-teacher comparisons match student initialization, prompts, rollout group size, and shared hyperparameters.
Let $N_s$ and $N_t$ indicate the number of student and teacher parameters used in per-token matrix multiplications.
GRPO costs $10N_s$ per generated token: $2N_s$ for generation, $2N_s$ for log-prob recomputation, and $6N_s$ for the update.
\opd{} adds one frozen teacher forward pass with $2N_t$ cost.
Each run pays for its generated tokens, excluding shared prompt context (Appendix~\ref{app:flops}).

\paragraph{GRPO and \opd{} Training Objectives.}
Let $\pi_\theta$ be the student policy, $\pi_T$ a frozen teacher, and, for a
prompt $q$, let $\{o_i\}_{i=1}^{G}$ be a group of $G$ student rollouts with
binary outcome rewards $\{r_i\}_{i=1}^{G} \in \{0,1\}$. We denote $o_{i,t}$ for
the $t$-th token of the $i$-th rollout and $q_{i,<t}$ for its prefix. GRPO
(Group Relative Policy Optimization) maximizes the mean-subtracted
advantage of the outcome reward, following \citet{liu2024understanding}:
\begin{equation*}
\mathcal{L}_{\text{GRPO}}(\theta) = -\,\mathbb{E}_{q,\{o_i\}}\!\left[\frac{1}{G}\sum_{i=1}^{G} \hat{A}_i \sum_{t=1}^{|o_i|}\log\pi_\theta(o_{i,t}\mid q, o_{i,<t})\right], \quad \hat{A}_i = r_i - \mathrm{mean}(\{r_j\}).
\end{equation*}
Our \opd{} objective, following \citet{li2026a260413016}, is the closed-form
reverse KL against $\pi_T$ over the student's top-$k$ support, computed on the
student's own rollouts. At each position $(i,t)$ we pick the student's top-$k$
token set $\mathcal{S}_k(o_{i,t})\subset\mathcal{V}$ ($k=16$), read the
student and teacher probabilities on those $k$ tokens, and renormalize each
to a distribution over $\mathcal{S}_k(o_{i,t})$:
\begin{equation*}
\tilde{\pi}_\theta(v\mid o_{i,<t}) = \frac{\pi_\theta(v\mid o_{i,<t})}{\sum_{u\in\mathcal{S}_k}\pi_\theta(u\mid o_{i,<t})},\quad
\tilde{\pi}_T(v\mid o_{i,<t}) = \frac{\pi_T(v\mid o_{i,<t})}{\sum_{u\in\mathcal{S}_k}\pi_T(u\mid o_{i,<t})},\quad v\in\mathcal{S}_k(o_{i,t}).
\end{equation*}
The loss is the closed-form reverse KL between the renormalized distributions:
\begin{equation*}
\mathcal{L}_{\text{OPD}}(\theta) = \mathbb{E}_{q,\{o_i\}}\!\left[\frac{1}{G}\sum_{i=1}^{G} \sum_{t=1}^{|o_i|} \sum_{v\in\mathcal{S}_k(o_{i,t})} \tilde{\pi}_\theta(v\mid o_{i,<t}) \Big[\log\tilde{\pi}_\theta(v\mid o_{i,<t}) - \log\tilde{\pi}_T(v\mid o_{i,<t})\Big]\right].
\end{equation*}
The outcome reward $r_i$ does \emph{not} enter $\mathcal{L}_{\text{OPD}}$.
\opd{} adds one frozen-teacher forward
pass over every token it trains on; GRPO adds nothing. Our training
hyperparameters are close to the ones in \cite{li2026a260413016} and are detailed in
Appendix~\ref{app:training-hypers}. 
% We discuss how we account FLOPs in Appendix~\ref{app:flops}.

\paragraph{Models and teachers}
Students are Qwen3 models of 1.7B, 4B, and 8B parameters~\citep{qwen3}, mostly base checkpoints (e.g. \texttt{Qwen3-4B-Base}) with  selected SFT cold starts for ablation.
Teachers include external post-trained thinking models (\texttt{Qwen3-}\{\texttt{4B}, \texttt{8B}, \texttt{14B}, \texttt{32B}\}), selected external models after further GRPO, and GRPO-trained base checkpoints at or above the student's size. 
% capable of outputting long chains of thought. With Qwen3 models, one can turn off the thinking mode by appending ``\texttt{<think>\textbackslash n\textbackslash n</think>\textbackslash n\textbackslash n}'' right after the input.
We call the student's own GRPO checkpoint a self-GRPO teacher; multi-domain runs use one such checkpoint per domain.
% \texttt{Qwen3} thinking mode is disabled by appending ``\texttt{<think>\textbackslash n\textbackslash n</think>\textbackslash n\textbackslash n}'' after the input % \sy{feel free to omit}.

\paragraph{Data.}
For \emph{math} we train on DAPO-Math-17K~\citep{yu2025dapo} and
DeepMath-103k~\citep{he2025deepmath}. For
\emph{code} we train on LiveCodeBench~v1--v5~\citep{jain2025livecodebench}
and a high-quality subset of CodeForces \texttt{open-r1/codeforces}. For the multi-domain experiments
(\S\ref{sec:merging}) we add \emph{MMLU-Pro} (knowledge) and
\emph{IFBench-RLVR}~\citep{pyatkin2025generalizingverifiableinstructionfollowing} (instruction following; rule-based constraint reward). Dataset construction details are in
Appendices~\ref{app:codeforces}--\ref{app:ifbench}.

\paragraph{SFT cold start.}
Our cold-started students are full fine-tunes of \texttt{Qwen3-}\{\texttt{1.7B}, \texttt{4B}\}\texttt{-Base}
on OpenThoughts-3~\citep{guha2025openthoughts}, following the SFT recipe
from Thinking Machines Lab~\citep{tml2025opd}. We sweep the SFT learning
rate to pick the initialization we then post-train from
(Tables~\ref{tab:sft-lr-1p7b} and~\ref{tab:sft-lr-4b}, Appendix~\ref{app:sft}); full hypers are
in Table~\ref{tab:sft-hypers}.

\paragraph{Evaluation.}
All evaluations report \texttt{mean@$k$} ($k{=}16$): the mean accuracy over $k$ independent samples per prompt. Math is evaluated on an out-of-distribution (\emph{OOD}) macro, the unweighted average over AIME24 (\num{30} prompts), AIME25 (\num{30}), and AMC23 (\num{83}), all held out from the math training data. Code is evaluated on LiveCodeBench v6~\citep{jain2025livecodebench} (\num{175} prompts). For the multi-domain experiments (\S\ref{sec:merging}), knowledge is evaluated on a \num{512}-prompt held-out subset of MMLU-Pro~\citep{wang2024mmlu}, and instruction following on IFBench~\citep{pyatkin2025generalizingverifiableinstructionfollowing}. Generation hyperparameters and length-maximum conventions are in Appendix~\ref{app:eval-hypers}. 

\section{External teachers do not make \opd{} reliably compute-efficient}
\label{sec:main}

Our initial observation on DAPO-Math-17K contradicts the compute-efficiency
claim. Training a \texttt{Qwen3-1.7B-Base} student against external
\texttt{Qwen3-}\{\texttt{4B}, \texttt{8B}, \texttt{32B}\} teachers at \textit{matched} settings, GRPO sits above every
\opd{} run at every evaluated FLOP count on both the DAPO validation split
and the OOD macro (Figure~\ref{fig:full-dapo}); larger teachers only increase the FLOPs of the \opd{} curve without lifting the student's performance.

\subsection{Dataset surgery}
\label{sec:stratification}

One reading of the negative result is that the dataset may contain instances that the model can solve well and thus favour GRPO unintentionally. We therefore stratify DAPO-Math-17K
by the joint difficulty of the student and a teacher model. This helps us understand how much \opd{} really provides a learning signal.

\paragraph{Stratify the dataset by the performances of student and teacher.} We
score every prompt in DAPO-Math-17K with \texttt{mean@32} under both the
\texttt{Qwen3-1.7B-Base} student and the \texttt{Qwen3-4B} thinking teacher, then bin along
both axes ($=\!0$ and $>\!0$ for the student; $=\!0$, $(0,\num{0.3}]$,
$(\num{0.3},\num{0.6}]$, $>\!\num{0.6}$ for the teacher), yielding eight
cells with matched train- and validation-set sizes
(Appendix~\ref{sec:difficulty}); OOD evaluation is kept as the AIME24/25 + AMC23
macro. We retrain GRPO and \opd{} in each subset, and
evaluate in-domain and OOD separately. Intuitively, the cell most favourable to \opd{} on paper is student~$=\!0$,
teacher~$>\!\num{0.6}$: GRPO's outcome reward is very weak,
so the dense signal should matter most.

\begin{figure}[t]
\centering
\includegraphics[width=\linewidth]{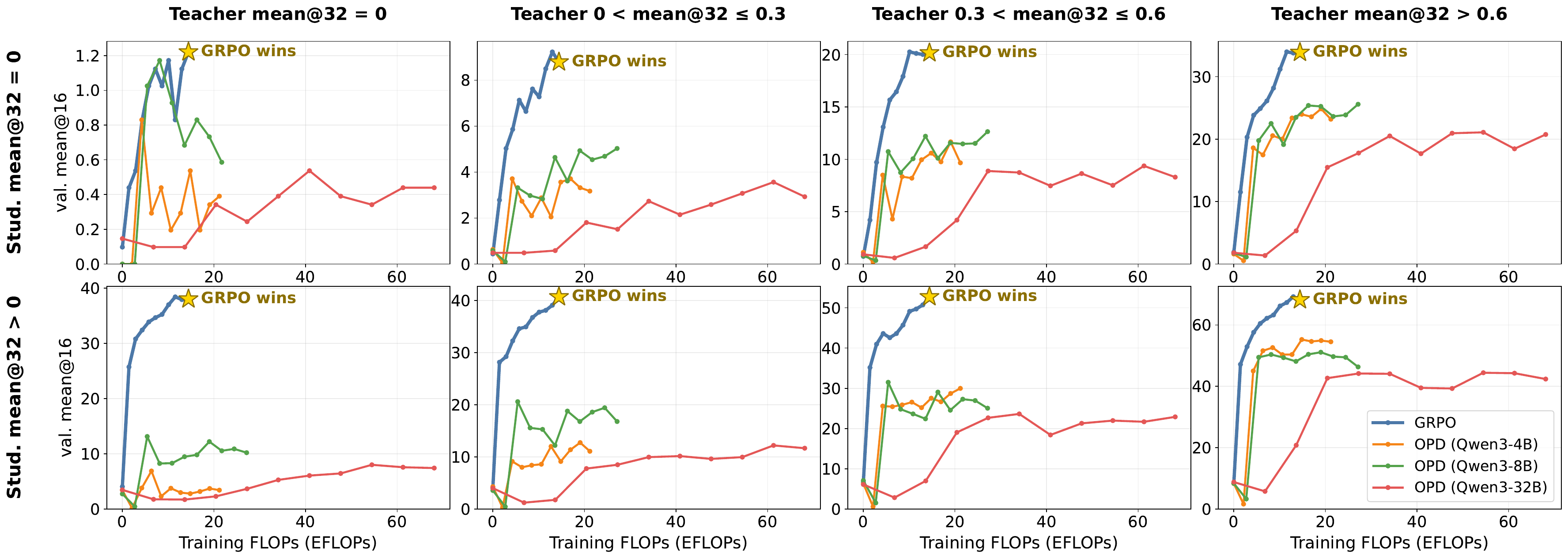}
\caption{\textbf{GRPO vs \opd{} across difficulty levels.} We stratify DAPO-Math-17K by \texttt{mean@32} solve rate of the student (\texttt{Qwen3-1.7B-Base}) and teacher (\texttt{Qwen3-4B}). This yields eight subsets, each corresponding to a performance level by student and teacher; each subplot shows in-domain \texttt{mean@16} against cumulative training FLOPs. GRPO outperforms every \opd{} variant in every cell, including the cell most favourable to \opd{} (top-right: student barely solves, teacher performs very well). OOD results in Figure~\ref{fig:difficulty-grid-ood}.}
\label{fig:difficulty-grid}
\end{figure}

To our surprise, such intuition falls apart. Figure~\ref{fig:difficulty-grid} shows the in-domain result: GRPO's final
accuracy exceeds every \opd{} run's peak in every cell at a smaller FLOP budget throughout. Although an \opd{} run sometimes briefly reaches GRPO's early-training performance at lower compute early in the training, the advantage of GRPO over \opd{} widens as we use stronger and larger teachers. The OOD macro tells a different story (Figure~\ref{fig:difficulty-grid-ood}). In the three cells where the teacher itself performs worst (the teacher \texttt{mean@32}$=0$ column, plus the cell with student $=0$ and teacher $\in(0,\num{0.3}]$), every \opd{} run reaches a higher peak OOD accuracy than GRPO, the opposite of what the dense-signal motivation predicts; GRPO wins the other five cells. Moreover, \opd{}'s final OOD macro stays within \pct{15}--\pct{18} across all eight cells regardless of the training subset, while GRPO's ranges from \pct{11.9} to \pct{20.5}. This suggests that part of what \opd{} learns from an external teacher does not depend on the training prompts; Appendix~\ref{app:extrapolation} examines this further.

Repeating the same comparison on a \texttt{Qwen3-1.7B-Base} student with thinking mode on yields an even stronger pattern: GRPO wins over every \opd{} run on In-Domain and OOD performance
(Appendix~\ref{sec:difficulty}).

\subsection{GRPO's compute efficiency advantage is robust across ablations}
\label{sec:ablations}

A natural objection to \S\ref{sec:main} is that the comparison uses a poor
configuration for \opd{}. We therefore search for a better configuration
among external teachers, measured on the compute axis. We vary nine factors
one at a time against the experiments in Figure~\ref{fig:teaser}. At matched
settings, GRPO is more compute-efficient than every external-teacher \opd{}
run along all nine axes. Appendix~\ref{app:ablation-detail} gives the setup
and results for each axis, and we discuss highlights below.

\paragraph{Student initialization.}
One hypothesis is that the \texttt{1.7B-Base} might be too weak to capture signal from the teacher. We therefore supervised-finetune \texttt{1.7B-Base} on 
OpenThoughts-3. With thinking mode off, GRPO still outperforms every \opd{} run
(Figure~\ref{fig:sft-init-1p7b-nothink}). We conducted a learning rate tuning to obtain the best performing checkpoint for \texttt{1.7B-Base} and \texttt{Qwen3-4B-Base} and we turn on the thinking mode. However, a good student initialization (compared with a sub-optimal one) doesn't change the ordering (Figure~\ref{fig:sft-init-1p7b}). With SFT-ed \texttt{4B-Base} as the student, we also experiment \opd{} with \texttt{QwQ-32B} --- the generator of the OpenThoughts-\num{1.2}M data, \opd{} still underperforms GRPO
(Figure~\ref{fig:sft-init-4b}). See more details in Appendix~\ref{sec:init}.

\paragraph{Training domain.}
To understand whether our observation is dataset specific, we train \texttt{4B-Base} student on two code datasets (LiveCodeBench v1--v5 and a high-quality CodeForces set) with GRPO and \opd{} under two external teachers (\num{8}B and \num{32}B). The ordering observed on math still holds on code: on LiveCodeBench v1--v5, GRPO reaches \pct{46.4} LCBv6 accuracy at \SI{0.6}{\exa\flop}, whereas \opd{}-\num{32}B peaks at only \pct{28.7} at \SI{68.7}{\exa\flop} and \opd{}-\num{8}B peaks at \pct{36.7} at \SI{45.7}{\exa\flop}. Results on CodeForces are similar (Figure~\ref{fig:code-math-extrap} left column). See more details in Appendix~\ref{app:training-domain}.

\paragraph{Additional ablations.}
We also sweep the learning rate
(Appendix~\ref{app:lr-detail}), batch size (Appendix~\ref{app:batch-detail}), reverse-KL estimator (closed-form top-$k$ vs
sampled-token, Appendix~\ref{app:rkl-estimator}),  the thinking mode (no-think vs
think, Appendix~\ref{app:teacher-detail}), the student model
sizes (Appendix~\ref{app:model-sizes}), and the RL framework
(\texttt{verl} and \texttt{tinker}, Appendix~\ref{app:framework-detail}). 
None of these changes the ordering. Reducing the rollout group size to
$G{=}1$ is the one change that helps: on single-domain math, external-teacher
\opd{} then outperforms GRPO at matched FLOPs, using $8\times$ as many
optimizer steps. However, the advantage disappears on code and on the multi-domain
mixes of Section~\ref{sec:merging} (Appendix~\ref{app:n1-defense}).
Across these ablations, vanilla reverse-KL \opd{} with an external teacher,
even one from the same model family, does not offer a reliable
compute-efficiency advantage over GRPO on the same student.
Appendix~\ref{app:why-detail} discusses the response-length inflation that
\opd{}-trained students show.

In the next section (\S\ref{sec:crossover}), following recent work that
GRPO-trains the teacher before distilling, we examine whether that change
to the external teacher alters what we observe here.

\begin{figure}[t!]
\centering
\includegraphics[width=\linewidth]{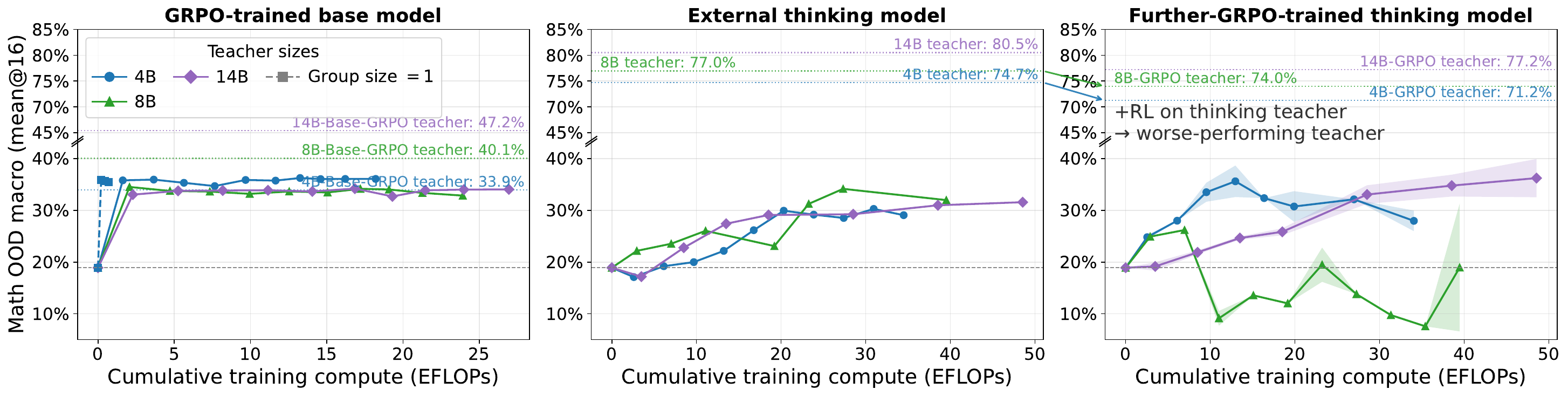}   
\caption{\textbf{Impact of running GRPO on teachers.}
\texttt{Qwen3-4B-Base} student on DAPO-Math-17K; $x$ is cumulative training FLOPs, $y$
is Math OOD macro \texttt{mean@16}; dotted horizontals mark each teacher's own
accuracy. We sweep teachers at \num{4}B, \num{8}B, and \num{14}B across three
types: GRPO-trained base (left), external thinking (centre), and
GRPO on external thinking (right). \textbf{Left:} all three GRPO-trained base teachers quickly converge the
student to similar performance; only the same-size $4$B
teacher is fully recovered, despite larger teacher having a higher performance ceiling; \textbf{Centre:} 
student performances climb much more slowly to the same level with GRPO-trained base; all students stay well below any teacher throughout training. \textbf{Right:} \opd{} with a further-GRPO-trained thinking teacher is more unstable than with a thinking teacher. }
\label{fig:teacher-by-nature}
\end{figure}

\section{Self-GRPO teacher consistently makes \opd{} compute-efficient}
\label{sec:crossover}

A highly cited work reports that GRPO-training the teacher before distillation can
help \opd{}~\citep{li2026a260413016}, but it establishes this claim on a limited experiment: running GRPO on \texttt{Qwen3-4B-Base} leads to a better teacher than \texttt{Qwen3-4B} (Non-thinking) when the student is \texttt{Qwen3-1.7B-Base}. We vary the teacher along three dimensions: 1) initialization --- a base model or an external thinking model, 2) whether it is used as-is or first GRPO-trained on the task, and 3) its size --- same or larger.

\paragraph{Setup.}
We fix the student to \texttt{Qwen3-4B-Base} on DAPO-Math-17K and sweep the teacher across three types: (i) \textit{GRPO-trained base} model: a base model (after pre-training only, e.g., \texttt{Qwen3-8B-Base}) is trained with GRPO on DAPO-Math-17K; (ii) \textit{external thinking} model: the off-the-shelf post-trained Qwen3 checkpoints as used in \S\ref{sec:stratification} (e.g., \texttt{Qwen3-8B}); (iii) \textit{further-GRPO-trained thinking} model: the external thinking model is further trained with GRPO on DAPO-Math-17K. All experiments in this section use thinking mode. Families (i) and (iii) share the same GRPO recipe; the only difference is the response-length maximum (\num{7168} for base, \num{16384} for thinking; full hyperparameters in Table~\ref{tab:shared-hypers}). For each type, we cover the teacher sizes at \num{4}B, \num{8}B, and \num{14}B. Every \opd{} run shares the same recipe; only the teacher varies.

\subsection{Student's GRPO checkpoint is consistently the most efficient teacher}
\label{sec:teacher-nature}

We present our results in Figure~\ref{fig:teacher-by-nature}. With GRPO-trained base
teachers, the student converges to similar performance by the first evaluated checkpoint
($\leq 2.5$~EFLOPs) --- the \num{4}B, \num{8}B, and \num{14}B teachers reach \pct{35.8},
\pct{34.5}, and \pct{33.0} --- and each stays in a narrow band throughout training. The
same-size \num{4}B teacher fully recovers its own GRPO accuracy (\pct{33.9}, which GRPO
reaches at \num{10.1}~EFLOP) by \num{1.62}~EFLOP, an ${\sim}\num{6}{\times}$ compute
reduction; with group size $G{=}1$ the same recovery costs only \num{0.2}~EFLOP
(${\sim}\num{50}{\times}$), matching the efficiency gains reported in the \opd{}
literature~\citep{tml2025opd}. Larger teachers are individually stronger, but their students
plateau at an even slightly lower level --- the extra capacity does not transfer. External thinking teachers improve the student much more slowly: the \num{8}B teacher peaks at \pct{34.2} (\num{27.2}~EFLOP),
while the \num{4}B and \num{14}B teachers stay below \pct{32}, all well below their teacher
accuracy. Further-GRPO-trained thinking teachers give the most unstable curves: the \num{4}B
teacher's student peaks at \pct{38.7} (step \num{80}), near the best GRPO accuracy
(\pct{38.8}), then collapses to \pct{26.0} by step \num{200}; the \num{8}B teacher's student
falls below initialization to \pct{6.6}; we repeat each run to confirm the instability. Only
using GRPO-trained student as teacher avoids the pathological response-length inflation that dominates other teachers (Figure~\ref{fig:resp-len-by-teacher}, Appendix~\ref{app:why-detail});
Appendix~\ref{app:iso-mechanism} discusses a potential explanation via changes in the spectra of model weight during optimization.

\paragraph{More \opd{} experiments with self-GRPO (the student's GRPO checkpoint) teacher}
On math, using \texttt{Qwen3-4B-Base} as student and its GRPO checkpoint as teacher gave both
compute-efficient improvement and full teacher recovery. In
Appendix~\ref{app:samebase-replication} we test this beyond that setting by (i) enlarging the
student to \texttt{Qwen3-8B-Base}, (ii) switching the domain from math to code, and (iii)
replacing the \texttt{Qwen3-*-Base} student (raw pretrained) with a post-trained
\texttt{Olmo-3-7B-Instruct-SFT}. All three retain high compute-efficiency and full teacher
recovery.

\begin{wrapfigure}{r}{0.45\linewidth}
\vspace{-\baselineskip}
\setlength{\abovecaptionskip}{-5pt}
\setlength{\belowcaptionskip}{-30pt}
\centering
\includegraphics[width=\linewidth]{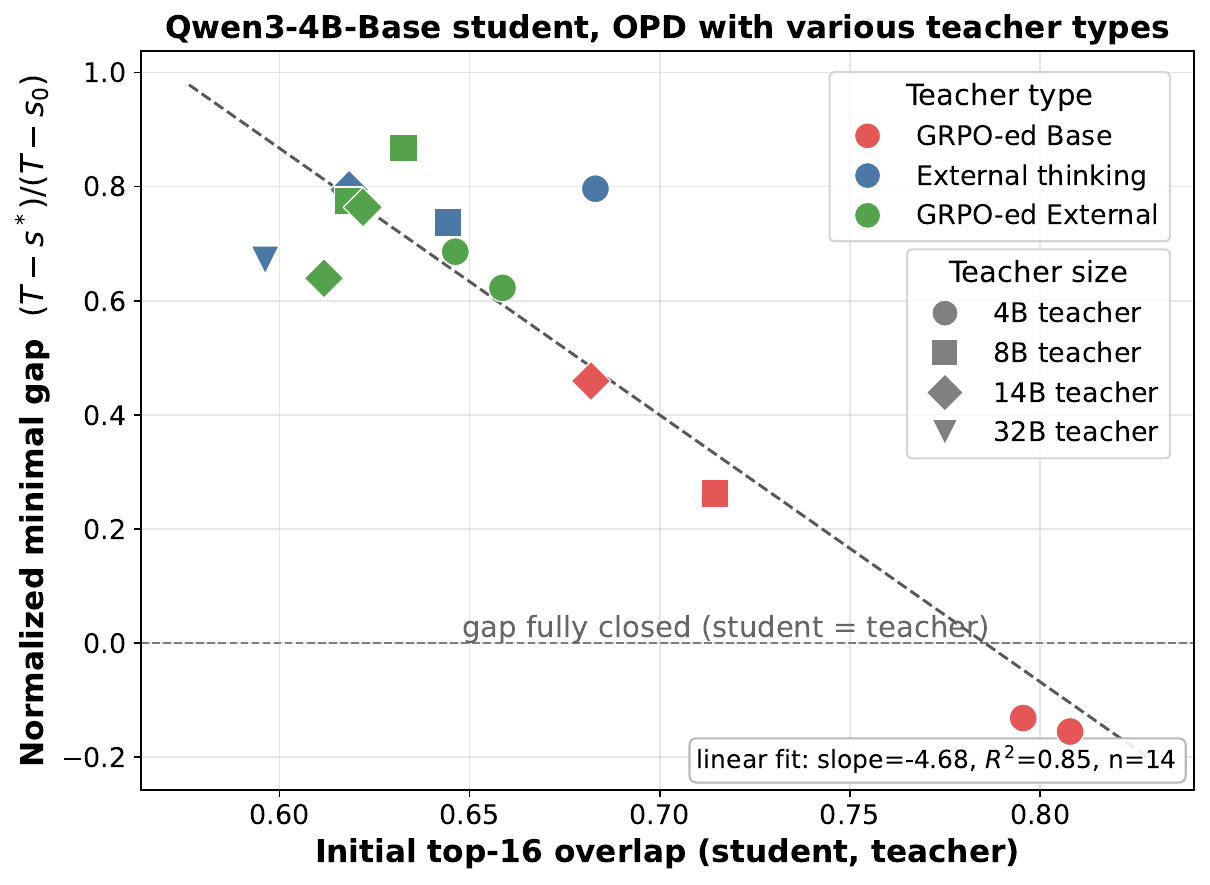}
\vspace{0.5pt}
\caption{\textbf{Initial top-\num{16} overlap correlates with the minimal
teacher--student gap}.}
\label{fig:phase-transition}
\end{wrapfigure}

\paragraph{An intriguing side observation in self-GRPO \opd{}}
Across our self-GRPO experiments on math and code, the student's best accuracy often can exceed the teacher's: in
Figure~\ref{fig:teacher-by-nature} the student reaches \pct{35.8} versus the teacher's
\pct{33.9}. The most striking case is \texttt{Olmo-3-7B-Instruct-SFT} on math, where the
self-GRPO teacher scores 40.0\% but the student reaches 51\% at \SI{1.25}{\exa\flop}
(Figure~\ref{fig:olmo3-selfgrpo}). The cause for such substantial gap is
unknown and left to future work.

\subsection{Can full teacher recovery be predicted?}
\label{sec:overlap-diagnostic}

Section~\ref{sec:teacher-nature} identifies setups that are compute-efficient --- using student's GRPO checkpoint as teacher, and setups that are not --- external thinking and further-GRPO-trained thinking teachers. We ask whether this outcome can be predicted \emph{before} the distillation run itself, using the initial top-$k$ token \emph{overlap} between student and teacher proposed by \citet{li2026a260413016} as a diagnostic, and scoring each run's outcome by its \emph{normalized minimal teacher--student gap}.

\paragraph{Overlap.}
On the first training batch, we score each position with the fraction of
the student's top-\num{16} token IDs that also appear in the teacher's
top-\num{16}, and average over all valid positions.

\paragraph{Normalized minimal teacher--student gap.}
To summarize how close each \opd{} run brings the student to its teacher,
we look at the gap between teacher accuracy $T$ and the best student accuracy $s^*$
reached at any point during training, and divide it by the initial
teacher--student accuracy gap at step zero $T - s_0$. 

\paragraph{Overlap correlates with the teacher--student gap.}
Across all \opd{} runs in \S\ref{sec:teacher-nature} and relevant runs in Figure~\ref{fig:tce-off-shelf},
initial top-\num{16} overlap correlates with a lower normalized minimal
teacher--student gap (Figure~\ref{fig:phase-transition}). However, we want to note that there's a large gap between GRPO-trained \texttt{Qwen3-8B-Base} teacher (red square) and GRPO-trained \texttt{Qwen3-4B-Base} (red dot). With data point missing in between, our observation serves as a side analysis in a limited setting (i.e. student is fixed to be \texttt{Qwen3-4B-Base}) rather than a generally applicable rule. 

\begin{guidance}
    \paragraph{Summary.}
Across nine ablations of reverse-KL \opd{}, no external
teacher transfers compute-efficiently and the student broadly loses to GRPO (\S\ref{sec:main}, \S\ref{sec:teacher-nature}). Only the
\emph{self-GRPO teacher} --- the student's own GRPO checkpoint --- consistently closes the gap with high compute-efficiency.
\end{guidance}

\section{Multi-teacher \opd{}: a natural use case of strong-to-weak \opd{}}
\label{sec:merging}

\begin{figure}[!t]
\centering
\includegraphics[width=\linewidth]{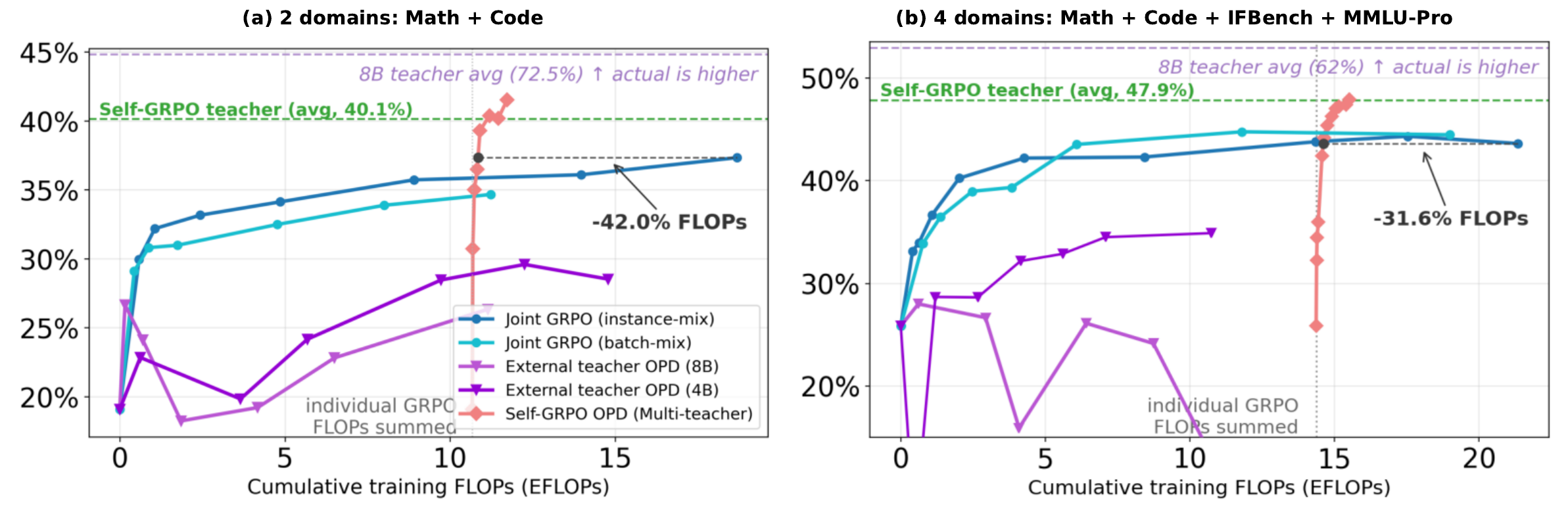}
\caption{\textbf{Multi-teacher \opd{} beats joint GRPO on domain-averaged
performance.} \texttt{Qwen3-4B-Base} student. \textbf{Left:} \num{2}-domain
average (math + code). \textbf{Right:} \num{4}-domain average (math, code,
instruction following, knowledge). Joint GRPO variants train on all domains simultaneously;
multi-teacher \opd{} merges per-domain GRPO experts into a single student. In two plots, the cost of Self-GRPO \opd{} with multiple teachers starts from the total FLOPs used to obtain all self-GRPO teachers. A single external
\num{8}B-thinking teacher (purple) can improve over initial student on 2-domain setting, but collapse on 4-domain setting.  Per-domain breakdowns are in Appendix~\ref{app:multi-teacher-detail}. }
\vspace{-15pt}
\label{fig:multi-teacher-merge}
\end{figure}

The finding of \S\ref{sec:crossover} remains a paradox in a single-task setting: \textit{why distill when you already have the GRPO-trained model?} Across multiple tasks, however, the paradox can become an opportunity.  Consider math and code: separate GRPO runs produce two specialists, but
neither checkpoint can serve both domains (Appendix~\ref{app:extrapolation}). This is where multi-teacher \opd{} has a distinct compute-efficiency opportunity: it can consolidate the GRPO specialists into a single multi-domain model that neither specialist provides, with only a small additional distillation cost. This opportunity only exists when the answer to the following question is true: \textit{Would the cost of training those experts and consolidating them via distillation beat that of joint GRPO on the same data mix?} Recent open-source models~\citep{yang2026a260319220, coreteam2026mimov2flashtechnicalreport, deepseek2026v4,kimi2026a260724653} use this recipe;
we set out to understand the rationale behind it.

\paragraph{Setup.}
We consider 4 domains in total: math on DAPO-Math-17K, code on LiveCodeBench-v1--v5, instruction following
on IFBench, and knowledge on MMLU-Pro. We experiment with 2-domain setting and 4-domain setting separately. 
From \texttt{Qwen3-4B-Base} we train one GRPO expert on each single domain, each with its own hyperparameters (Table~\ref{tab:indiv-grpo-hypers}).
We build a data mix whose per-domain training instance count matches each individual
expert, and we compare four methods on it: (i)~\emph{Joint GRPO (instance-mix)}, which fills every batch with instances drawn proportionally from all domains at one shared batch size; (ii)~\emph{Joint GRPO (batch-mix)}, which draws each batch from each domain (each domain has its own batch size, notably a smaller batch on code domain) per step and interleaves domains across steps; (iii)~\emph{Multi-teacher self-GRPO \opd{}}, which routes each instance to the self-GRPO teacher of the corresponding domain and computes the reverse KL; (iv)~\emph{External teacher \opd{}} with a single \num{8}B-thinking teacher, the best-performing external teacher based on results from \S\ref{sec:main} and \S\ref{sec:crossover}. All \opd{} runs use rollout group size $G{=}1$ unless otherwise noted. We run each method on \num{2}-domain (math + code) and \num{4}-domain mixes to show some scaling behavior; see Appendix~\ref{app:multi-teacher-detail} for more details.

% \leo{discuss results more thoroughly}
% (i)~\emph{joint GRPO (instance-mix)}, which fills every batch with instances drawn proportionally from all domains at one shared batch size;
% (ii)~\emph{joint GRPO (batch-mix)}, which draws each batch from a single domain and interleaves domains across steps, preserving every specialist's own batch size (notably a smaller batch on code domain);
% (iii)~\emph{Self-GRPO \opd{} (multi-teachers)}, which routes each instance to the self-GRPO teacher of the corresponding domain and  compute the reverse KL; and
% (iv)~\emph{external teacher \opd{}} with a single \num{8}B-thinking teacher. 

\paragraph{Results.}
Figure~\ref{fig:multi-teacher-merge} shows that the single-domain success of self-GRPO
\opd{} transfers to the multi-domain setting. On the \num{2}-domain mix, the two specialists
cost \SI{10.7}{\exa\flop}, and multi-teacher self-GRPO \opd{} reaches their ceiling in an
additional \SI{0.55}{\exa\flop} (${\sim}\pct{5.2}$); on the \num{4}-domain mix, the four
specialists cost \SI{14.4}{\exa\flop}, performance recovered in an additional \SI{1.15}{\exa\flop}
(${\sim}\pct{8.0}$). Single-domain and multi-teacher self-GRPO \opd{} are on par in
compute-efficiency (Figures~\ref{fig:mopd-n1-joint4},~\ref{fig:mopd-n1-mathcode}). Even after
accounting for the \opd{} and self-GRPO-teacher costs, both joint-GRPO variants (instance-mix
and batch-mix) underperform on both multi-domain settings; specifically, multi-teacher self-GRPO \opd{} matches Joint GRPO(instance-mix) with \pct{42.0} fewer FLOPs on 2 domains and \pct{31.6} fewer FLOPs on 4 domains. \emph{External-teacher \opd{} is worst}: on 2 domains with 8B-thinking teacher, it reaches
\pct{26.7}, drops, then finishes at a similar level; scaling to 4 domains, the student collapses on most domain
(\pct{0} math, \pct{0} code, \pct{7} MMLU-Pro, \pct{18} IFBench) despite the
\num{8}B-thinking teacher being the strongest model overall, with severe response-length
inflation at inference. The group-size ablation (Appendix~\ref{app:multi-teacher-detail}) shows
$G{=}1$ reaches the specialist ceiling at substantially lower compute than $G{=}8$
(Section~\ref{sec:teacher-nature}). 

\begin{guidance}
    \paragraph{Summary.}
Multi-teacher \opd{} (route each instance to its domain's specialist to
compute the reverse KL) turns the single-task paradox of
\S\ref{sec:crossover} into a recipe: merge per-domain GRPO experts into one
student, reach the specialists' ceiling for an extra
$\pct{5.2}$--$\pct{8.0}$ of the FLOPs spent training them, and beat both
joint GRPO and external-teacher \opd{} at matched compute.

% Multi-teacher \opd{} --- route data to each domain specialists to calculate reverse KL --- turns the single-task paradox of
% \S\ref{sec:crossover} into a recipe: merge per-domain GRPO experts into
% one student, reach the specialists' ceiling for a
% ${\sim}\pct{5.2}$--$\pct{8.0}$ FLOPs of getting the specialists, and beat joint GRPO and external \opd{} at matched compute.
\end{guidance}

\paragraph{Benefit of multi-teacher \opd{} with self-GRPO teachers.}
Mulit-teacher pipeline enjoys three benefits at once.
\emph{Better overall performance:} as we show in Figure~\ref{fig:multi-teacher-merge}, using multi-teacher \opd{} with self-GRPO teachers leads to better performance than joint GRPO.
\emph{Modular development:} different domains naturally require
different RL infrastructure (code sandboxes, tool-use environments, symbolic
verifiers) and different optimal hyperparameters (learning rate, rollout
length, response-length maximum); training each expert independently lets each
team use its own stack and tune at its own schedule to get the best expert on each domain.
\emph{Parallelizable specialist training:} per-domain runs can naturally happen
concurrently rather than sharing a single joint schedule.

\paragraph{Relation to frontier practice.}
Several recent frontier-model reports use multi-teacher \opd{} to combine
domain-expert teachers into a shared student, including
Nemotron-Cascade~2~\citep{yang2026a260319220},
DeepSeek-V4~\citep{deepseek2026v4}, and Kimi
K3~\citep{kimi2026a260724653}. These reports describe the recipe as an
engineering choice without matched-compute measurement. Our results provide
the rationale behind their decision:
Section~\ref{sec:crossover} shows that \opd{} succeeds when the teacher is a
GRPO checkpoint of the student, and
Section~\ref{sec:merging} shows that merging per-domain experts trained under
this condition via \opd{} outperforms training all domains jointly with GRPO or using an external teacher.

\section{Conclusion}

We give the first systematic, FLOPs-based characterization of strong-to-weak
\opd{} against GRPO, and identify when \opd{} is a promising post-training tool.
First, we find external teachers under vanilla reverse-KL broadly
lose to GRPO across our nine-axis ablation (\S\ref{sec:main}); and we widely observe pathological behavior from the trained student with external teachers. 
Second, through carefully studying the impact of GRPO on teachers, we realized the \emph{self-GRPO teacher} (the student's own GRPO checkpoint) is the only setup where the student consistently learns compute-efficiently and fully recovers teacher performance (\S\ref{sec:crossover}). It matches GRPO's performance at \num{50.0}$\times$ fewer FLOPs, replicated across domains, student sizes, and student families; the student's best performance can even substantially exceed the teacher's, calling for further study. 
Third, multi-teacher \opd{} with self-GRPO teachers is a cheap
consolidation step (\S\ref{sec:merging}).  It recovers every specialist for an extra \pct{5.2}
(2 domains) or \pct{8.0} (4 domains) of specialist training FLOPs --- cheaper
than joint GRPO on the same data; and substantially cheaper than
external-teacher \opd{}, which underperforms at 2 domains and collapses at 4. 
Together, these results suggest \opd{} is not a straightforward replacement for GRPO, but it is an efficient way to fold many same-origin GRPO-trained specialists into a single generalist student.

\subsection*{AI use statement}

We used generative AI tools (large language models) to assist with drafting and editing portions of
the manuscript text. All AI-assisted text was reviewed, revised, and verified by the authors. We did
not use generative AI tools (e.g., image generation) to generate experimental results, design the method or produce experimental figures.
LLM-generated code snippets used during development were reviewed and tested for correctness. We take responsibility for the final content of this work, including all text, claims, and artifacts.

\subsection*{Limitation and future work.}
Our study focuses on the vanilla reverse-KL \opd{} objective with a
closed-form top-$k$ estimator (\S\ref{sec:setup}). Recent work extends
this baseline along axes we do not sweep at matched FLOPs: hybrid
objectives that blend \opd{} with an outcome-reward
term~\citep{ko2026a260311137,tan2026a260627814,liu2026a260513230},
alternative divergences~\citep{luo2026a260408527,chen2026a260517862},
trajectory reshaping and per-token
reweighting~\citep{li2026a260413016,rang2026a260505940}, and async or
speculative rollout pipelines~\citep{feng2026a260705394,ren2026a260716872}.
Any of these may narrow the FLOPs gap we identify. We encourage future
work claiming an \opd{} advantage over GRPO to report training FLOPs as
an implementation-invariant measure, so that improvements in objective,
rollout scheduling, and infrastructure can be disentangled and directly
comparable across papers.

\subsection*{Ethics statement}

This work trains and evaluates open-weight language models on public
mathematics and competitive-programming datasets; it involves no human subjects,
no personal data, and no new data collection. We do not foresee direct
negative societal impacts from the distillation methodology itself.

\subsection*{Reproducibility statement}
The training and evaluation recipe for every experiment in this paper is documented in
Section~\ref{sec:setup} and expanded in Appendix~\ref{app:training-hypers}.
Shared hyperparameters common to all runs are listed in
Table~\ref{tab:shared-hypers}; the GRPO- and \opd{}-specific settings are in
Tables~\ref{tab:grpo-hypers}~and~\ref{tab:opd-hypers}, and the per-ablation
overrides for the sweeps in Section~\ref{sec:ablations} are in
Table~\ref{tab:ablation-deviations}. All students, teachers, training corpora, and
evaluation benchmarks are publicly released models and
open datasets identified by version and citation in
Section~\ref{sec:setup}~and~Appendices~\ref{app:codeforces}--\ref{app:ifbench}.
The framework versions and rollout engine are named in
Section~\ref{sec:setup} (verl~\texttt{0.7.0.dev} and \texttt{0.8.0.dev},
SGLang~\texttt{0.5.2}), and the framework replication of
Appendix~\ref{app:framework-detail} additionally names the tinker library and its
LoRA rank, learning rate, batch size, rollout count, and response maximum for the
matched \num{9}B replication. We also release the prompts used for formatting training instances in Appendix~\ref{app:prompt-templates}.
All models, datasets, and framework releases used in this work are public.

% We thank Zhepeng Cen for helpful discussions in the early stages of this project, and Guanghao Ye for personally offering the \texttt{tinker} credits used in our experiments.

\bibliographystyle{iclr2027_conference}
\bibliography{refs}

\newpage
\appendix
% Appendix roadmap:
% - Related work: literature placement.
% - Evaluation hypers / SFT / CodeForces: support \S\ref{sec:setup}.
% - Per-ablation detail through Compute-axis ablations: support \S\ref{sec:main}--\S\ref{sec:ablations}.
% - High-overlap sweep / ISO: support \S\ref{sec:crossover}.
% - Threats to validity: paper-level.
% - Supporting tables: cross-cutting reference at the end.

\section{Literature survey: how we filter to zero FLOPs-based wins}
\label{app:audit}
\begin{wrapfigure}{r}{0.45\linewidth}
\vspace{-\baselineskip}
\setlength{\abovecaptionskip}{-10pt}
\setlength{\belowcaptionskip}{-60pt}
\centering
\includegraphics[width=\linewidth]{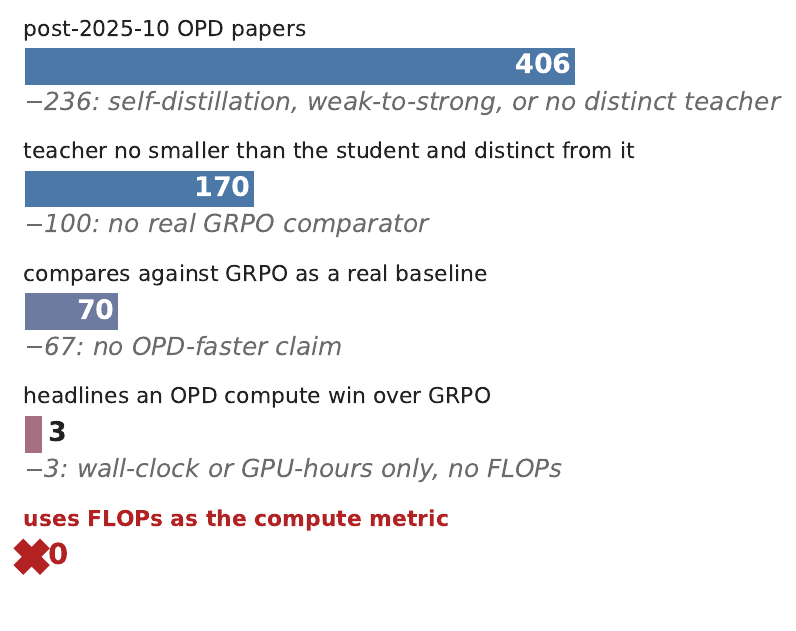}
\vspace{0.5pt}
\caption{\textbf{No paper in our audit reports a FLOPs-based \opd{} compute win over GRPO.} We filter \num{406} \opd{} papers posted after October 2025. Each bar is the number of papers that pass the criterion above it, and the italic text below each bar names what the next filter removes. \num{170} papers use a teacher that is distinct from the student and at least as large; \num{70} of these compare against GRPO as a baseline; \num{3} of those claim that \opd{} is cheaper than GRPO; and none of the three measures compute in FLOPs.}

\label{fig:audit-funnel}
\end{wrapfigure}

\paragraph{Corpus.}
We take the deduplicated union of three public \opd{} bibliographies:
\texttt{chrisliu298/\allowbreak awesome-on-policy-distillation},
\texttt{thinkwee/\allowbreak awesomeopd}, and
\texttt{lingyunyang-opd}. Identity is resolved by arXiv ID, DOI, and
normalized-title merge. Restricted to papers whose first arXiv
version was posted between 2025-10-06 and 2026-08-06, the
corpus is \num{406} unique preprints.

\paragraph{Reading.}
Every paper is read end-to-end by \texttt{Claude Opus 4.8}, which
returns a schema-enforced structured record; every classification-bearing
field is required to be backed by a verbatim quotation with a page
locator copied from the PDF. Papers clearing the coarse rule (a
distinct teacher plus any GRPO mention) are then re-read by a second,
independent Claude Opus 4.8 invocation that does not see the first
pass's verdict. No keyword-only filters make classification calls;
every field is grounded in PDF-verified quotation.

\paragraph{Filter criteria.}
Starting from \num{406} papers, we apply four criteria in strict
precedence. Each row is the surviving count after the criterion is
applied.

\begin{itemize}
\item Corpus of post-2025-10 \opd{} preprints: \textbf{\num{406}}.
\item Teacher no smaller than the student and distinct from the
  student: the teacher is a distinct model that is larger than or the
  same size as the student. Excludes self-distillation and weak-to-strong
  (smaller teacher) setups. \textbf{\num{170}} (drops \num{236}: 146
  self-distillation, 2 smaller (weak-to-strong), 8 unclear or
  not-applicable, plus 80 papers with no distinct teacher or no
  resolvable teacher role).
\item Compares against GRPO as a real baseline: canonical GRPO
  or a named variant (DAPO, GRPO++, REINFORCE++, or a paper-specific
  modification) is run with reported numbers, verified from the PDF; a
  text mention without a GRPO experiment does not count.
  \textbf{\num{70}} (drops \num{100}: papers with an eligible teacher
  but no real GRPO baseline).
\item Reports an \opd{} compute advantage over GRPO: the
  paper's abstract, main table, or discussion explicitly headlines an
  \opd{} wall-clock, GPU-hour, or FLOPs win over the GRPO baseline.
  \textbf{\num{3}} (drops \num{67}: papers that report a compute metric
  but make no \opd{}-faster claim, or where GRPO wins). The three
  surviving papers are REOPOLD~\citep{ko2026a260311137}, Near-Policy
  Distillation~\citep{rang2026a260505940}, and
  $f$-OPD~\citep{chen2026a260517862}.
\item Uses FLOPs as the compute metric. Wall-clock and
  GPU-hours entangle the training objective with implementation and
  hardware; FLOPs is the only implementation-invariant compute metric.
  \textbf{\num{0}} (drops \num{3}). None of the three surviving
  headline claims reports a FLOPs comparison:
  \begin{itemize}
  \item REOPOLD~\citep{ko2026a260311137} reports GPU-hours,
    wall-clock, optimizer steps, and rollouts at a matched 300-step
    budget for RKL \opd{} vs canonical GRPO. Per-step compute differs
    substantially between the two methods (\opd{} adds one teacher
    forward pass per training token), so the reported wall-clock
    speedup does not translate into a matched-FLOPs claim.
  \item Near-Policy Distillation~\citep{rang2026a260505940}
    reports wall-clock and optimizer steps for its async-KD
    acceleration versus synchronous \opd{}; GRPO enters only as a
    downstream RL stage compared against SFT and NPD initializations,
    not as a matched-FLOPs \opd{}-vs-GRPO comparison. No FLOPs are
    reported.
  \item $f$-OPD~\citep{chen2026a260517862}
    reports throughput, resolve rate, and GPU-hours on SWE-bench for
    vanilla sync/async \opd{} versus canonical GRPO. Throughput and
    resolve rate do not capture total compute, and GPU-hours are
    implementation-dependent; no FLOPs comparison appears.
  \end{itemize}
\end{itemize}

\paragraph{Conclusion.}
No paper in the surveyed corpus establishes a FLOPs-based compute
advantage for \opd{} with a same-size-or-larger teacher over GRPO. This
is the gap our work fills.

\section{Experiment setup details}
\label{app:experiment-setup}

\subsection{Compute FLOPs Accounting.}
\label{app:flops}

We account for training compute with a single per-token forward/backward model,
applied identically to both methods. Let $N_s$ be the number of student
parameters that enter a matrix multiplication per token, and $N_t$ the
corresponding count for the teacher. Charging a forward pass at $2$ FLOPs per
parameter per token and a combined forward and backward pass at $6$, the
student is used three times per token: generation ($2N_s$), log-probability
recompute ($2N_s$), and the policy update ($6N_s$), for $10\,N_s$ in total.
\opd{} additionally runs one frozen-teacher forward pass at $2\,N_t$, and GRPO
runs none. The per-token cost is therefore
\begin{equation}
\label{eq:flops-per-token}
f_{\mathrm{GRPO}} = 10\,N_s,
\qquad
f_{\mathrm{OPD}} = 10\,N_s + 2\,N_t,
\end{equation}
We count only generated
tokens and not the shared prompt context for each run.

\subsection{Evaluation: generation hyperparameters and length caps}
\label{app:eval-hypers}

Table~\ref{tab:eval-hypers} lists the generation hyperparameters used for evaluation. All evaluations sample $n{=}\num{16}$ responses per prompt. Different domains use different response-length caps. The decoding configuration with thinking mode on follows the configurations in \citet{qwen3}.

\begin{table}[h]
\centering
\caption{Evaluation generation hyperparameters by thinking mode and domain.}
\label{tab:eval-hypers}
\small
\begin{tabular}{@{}llcccc@{}}
\toprule
Thinking mode & Domain & Temperature & top-$p$ & top-$k$ & Response maximum (tokens) \\
\midrule
Off & Math & \num{0.6} & \num{0.95} & -- & \num{7168} \\
\midrule
On  & Math         & \num{0.6} & \num{0.95} & \num{20} & \num{30720} \\
On  & Code (LCBv6) & \num{0.6} & \num{0.95} & \num{20} & \num{31744} \\
On  & MMLU-Pro     & \num{0.6} & \num{0.95} & \num{20} & \num{7168} \\
On  & IFBench      & \num{0.6} & \num{0.95} & \num{20} & \num{30720} \\
\bottomrule
\end{tabular}
\end{table}

Every table and figure states its generation budget (response-token maximum),
because \opd{} rollouts saturate length caps and the budget therefore
changes conclusions (\S\ref{sec:why}). Where a budget must be matched post
hoc we truncate responses to the token prefix and re-grade with the same
verifier the run used. This is exact because generation is causal.

\subsection{Training hyperparameters}
\label{app:training-hypers}

Tables~\ref{tab:shared-hypers}--\ref{tab:opd-hypers} list the
training-recipe values widely used in our experiments: shared hyperparameters, GRPO-specific configuration, and
\opd{}-specific configuration. Table~\ref{tab:ablation-deviations}
records the appendix sections that document deviations along each ablation axis.

\begin{table}[h]
\centering
\caption{Shared training hyperparameters. Both GRPO and
\opd{} use identical values for every setting in this table.}
\label{tab:shared-hypers}
\small
\begin{tabular}{@{}lp{0.72\linewidth}@{}}
\toprule
Setting & Value \\
\midrule
Framework               & \texttt{verl 0.7.0.dev} (custom fork; see \S\ref{sec:framework} for the native replication) \\
Rollout inference       & SGLang \texttt{0.5.2} \\
Rollout group size $G$  & \num{8} rollouts per prompt \\
Batch size (prompts)    & \num{64} \\
Response-length maximum     & \num{7168} tokens (unless noted otherwise) \\
Rollout sampling        & temperature~$= \num{1.0}$, top-$p = \num{1.0}$, top-$k$ disabled \\
Data shuffling          & disabled \\
Gradient clipping       & max norm $= \num{1.0}$ \\
Optimizer               & Adam ($\beta_1{=}\num{0.9}$, $\beta_2{=}\num{0.95}$, $\epsilon{=}\num{1e-8}$) \\
Weight decay            & \num{0} \\
Learning rate           & \num{2e-6} (constant, no warmup) \\
Optimizer steps         & \num{200} \\
Precision               & bf16 rollout inference; optimizer in fp32 \\
Hardware                & \num{8}$\times$H200, single node \\
Sharding                & FSDP \\
Thinking mode           & on \\
KL-to-reference term    & disabled (neither method pays for a reference-policy forward pass) \\
\bottomrule
\end{tabular}
\end{table}

\begin{table}[h]
\centering
\caption{GRPO-specific configuration.}
\label{tab:grpo-hypers}
\small
\begin{tabular}{@{}lp{0.72\linewidth}@{}}
\toprule
Setting & Value \\
\midrule
Reward               & binary outcome (\num{0}/\num{1}); math: exact match; code: pass/fail \\
KL-to-reference      & disabled ($\beta_{\mathrm{KL}}{=}\num{0}$; reference-policy KL term ablated, \S\ref{sec:setup}) \\
Policy-update epochs  & \num{1} (single pass over each rollout batch) \\
Clip ratio           & asymmetric (DAPO-style): $\epsilon_{\mathrm{low}}{=}\num{0.2}$, $\epsilon_{\mathrm{high}}{=}\num{0.28}$ $\Rightarrow$ $[\num{0.8},\,\num{1.28}]$ \\
Loss aggregation     & token-mean \\
Teacher forward pass & none (GRPO needs no teacher) \\
\bottomrule
\end{tabular}
\end{table}

\begin{table}[h]
\centering
\caption{\opd{}-specific configuration for the headline matched pair.}
\label{tab:opd-hypers}
\small
\begin{tabular}{@{}lp{0.72\linewidth}@{}}
\toprule
Setting & Value \\
\midrule
Teacher temperature     & \num{1.0} \\
Support set             & student top-$k$ tokens, $k{=}\num{16}$ (closed-form reverse KL over student's top-$k$ support) \\
Divergence              & reverse KL, $\mathrm{KL}(\tilde{\pi}_\theta \,\|\, \tilde{\pi}_T)$ on renormalized top-$k$ \\
Outcome reward          & not used (vanilla reverse-KL \opd{}; $r_i$ does not enter $\mathcal{L}_{\text{OPD}}$) \\
Policy-update epochs    & \num{1} (single pass over each rollout batch) \\
Loss aggregation        & token-mean \\
\bottomrule
\end{tabular}
\end{table}

\begin{table}[h]
\centering
\caption{Deviations from the shared recipe by ablation axis. Each row names the
appendix section documenting the change.}
\label{tab:ablation-deviations}
\small
\begin{tabular}{@{}lll@{}}
\toprule
Ablation axis & Key deviation & Section \\
\midrule
Dataset difficulty         & no-think mode & \S\ref{sec:difficulty} \\
Learning rate              & swept $\{\num{2e-6}$--$\num{5e-5}\}$, no-think mode & \S\ref{sec:lr}  \\
Batch size (12800-prompt budget) & $\{64, 128, 256\}$ at $\{200, 100, 50\}$ steps, no-think mode & \S\ref{app:batch-detail} \\
Reverse-KL estimator       & closed-form top-$k$ vs sampled-token, no-think mode  & \S\ref{app:rkl-estimator} \\
Thinking mode              & OpenThoughts-3-SFT init, no-think and think teachers & \S\ref{sec:teacher} \\
Student initialization     & \texttt{Qwen3-4B-Base}, OpenThoughts-3-SFT variants & \S\ref{sec:init} \\
Model sizes                & \texttt{Qwen3-4B-Base} and \texttt{Qwen3-8B-Base} students & \S\ref{app:model-sizes} \\
Training domain            & LCBv1--v5, CodeForces, response maximum \num{16384} & \S\ref{app:training-domain} \\
Framework                  & \texttt{verl 0.8.0.dev} with native \opd{}, \texttt{tinker} & \S\ref{sec:framework} \\
% Framework (tinker)         & Qwen3.5-9B-Base, rank-\num{128} LoRA, lr \num{1e-4} & \S\ref{sec:framework} \\
Rollout size                  & 1  & \S\ref{app:n1-defense} \\
\bottomrule
\end{tabular}
\end{table}

\subsection{Infrastructure.} 
\label{app:infra}
Training uses verl~\citep{sheng2025verl} with
SGLang~\citep{zheng2024sglang}~\texttt{0.5.2} rollout on a single \num{8}-GPU
H200 node; the two framework generations we compare are labelled
\emph{old/custom} (verl~\texttt{0.7.0.dev} fork with our own \opd{}
implementation) and \emph{latest/native} (verl~\texttt{0.8.0.dev} with
native \opd{} implementation).

\subsection{SFT cold-start recipe and learning-rate sweep}
\label{app:sft}

Our cold-started students are full fine-tunes of \texttt{Qwen3-}\{\texttt{1.7B},
\texttt{4B}\}\texttt{-Base} on OpenThoughts-3~\citep{guha2025openthoughts}. We follow the
SFT recipe from step~1 of the Thinking Machines Lab \opd{} blog
post~\citep{tml2025opd} as closely as \texttt{verl} allows.
Table~\ref{tab:sft-hypers} lists the hyperparameters we use.
Tables~\ref{tab:sft-lr-1p7b} and~\ref{tab:sft-lr-4b} report the SFT
learning-rate sweep we ran to pick the initialization: for both students
we sweep lr $\in \{\num{1e-6}, \num{1e-5}, \num{1e-4}, \num{1e-3}\}$ and
evaluate at step \num{3000} under the paper's default decoding, then use
the best learning rate for downstream RL and \opd{} runs.

\begin{table}[h]
\centering
\caption{SFT hyperparameters. We follow the SFT recipe from the
Thinking Machines Lab \opd{} blog post~\citep{tml2025opd} as closely as
\texttt{verl} allows.}
\label{tab:sft-hypers}
\small
\begin{tabular}{@{}ll@{}}
\toprule
Hyperparameter & Value \\
\midrule
Dataset & OpenThoughts-3, 1.2M-example single-turn subset \\
Prompts trained on & \num{128}$\times$\num{3000} = \num{384000} \\
Loss support & response tokens only (all assistant messages) \\
Global batch size & \num{128} \\
Max sequence length & \num{16384} \\
Total epochs & 1 (single pass) \\
Learning-rate schedule & linear decay to \num{0}, no warmup \\
Learning rate & swept in $\{\num{1e-6},\num{1e-5},\num{1e-4},\num{1e-3}\}$ (see Table~\ref{tab:sft-lr-sweep}) \\
Adam ($\beta_1,\beta_2,\epsilon$) & $(0.9,\,0.95,\,\num{1e-8})$ \\
Weight decay & 0 \\
Fine-tuning mode & full (no LoRA) \\
Precision / infra & FSDP, \num{8}$\times$H200 (single node) \\
\bottomrule
\end{tabular}
\end{table}

The SFT learning-rate sweep for the cold-start initialization is reported
in Tables~\ref{tab:sft-lr-1p7b} and~\ref{tab:sft-lr-4b}: \texttt{Qwen3-1.7B-Base}
and \texttt{Qwen3-4B-Base}, full fine-tuned on OpenThoughts-3 for a single pass
(${\approx}\,\num{3000}$ optimizer steps at global batch \num{128}),
evaluated on the paper's OOD macro (AIME24 + AIME25 + AMC23,
\texttt{mean@16} at temperature \num{0.6}, top-$p$ \num{0.95},
top-$k$ \num{20}, thinking enabled, \num{30720}-token maximum). The
\emph{Base ref} column is the external Base model before SFT under the same
decoding. 

\begin{table}[h]
\centering
\caption{SFT learning-rate sweep for the cold-start initialization.
\texttt{mean@16} (\%) is reported. Best per row in bold.}
\label{tab:sft-lr-sweep}
\small
\begin{subtable}{\linewidth}
\centering
\caption{\texttt{Qwen3-1.7B-Base}}
\label{tab:sft-lr-1p7b}
\begin{tabular}{@{}lrrrrr@{}}
\toprule
Benchmark & \num{1e-6} & \num{1e-5} & \num{1e-4} & \num{1e-3} & Base ref \\
\midrule
AIME24    &  4.8 &  1.3 & \best{11.3} & 0.0 &  3.5 \\
AIME25    &  3.1 &  2.7 & \best{13.8} & 0.0 &  2.7 \\
AMC23     & 26.4 & 13.6 & \best{37.5} & 0.3 & 27.0 \\
\midrule
OOD macro & 11.4 &  5.9 & \best{20.8} & 0.1 & 11.1 \\
\bottomrule
\end{tabular}
\end{subtable}
\\[1em]
\begin{subtable}{\linewidth}
\centering
\caption{\texttt{Qwen3-4B-Base}}
\label{tab:sft-lr-4b}
\begin{tabular}{@{}lrrrrr@{}}
\toprule
Benchmark & \num{1e-6} & \num{1e-5} & \num{1e-4} & \num{1e-3} & Base ref \\
\midrule
AIME24    &  7.9 & 20.8 & \best{47.1} & 0.0 & 10.0 \\
AIME25    &  5.0 & 16.9 & \best{40.0} & 0.0 &  7.7 \\
AMC23     & 31.6 & 49.3 & \best{73.0} & 2.5 & 39.0 \\
\midrule
OOD macro & 14.8 & 29.0 & \best{53.4} & 0.8 & 18.9 \\
\bottomrule
\end{tabular}
\end{subtable}
\end{table}

\subsection{High-quality CodeForces training set}
\label{app:codeforces}

We carve a high-quality CodeForces training set from
\texttt{open-r1/codeforces} (Hugging Face). Starting from
the \num{6594}-problem bank, we keep only the \num{1213} problems whose
\texttt{official\_tests\_complete} flag is true (the release carries the
full official test set rather than a truncated prefix, so a wrong solution
cannot pass by chance), and drop \num{6} problems whose graded tests are
already shown as worked examples in the statement.
The remaining \num{1207} problems split \num{1107} train / \num{100}
val by a random shuffle.

\subsection{MMLU-Pro: 5-shot knowledge prompts}
\label{app:mmlu-pro}

We use the MMLU-Pro test set~\citep{wang2024mmlu} (\num{11904} multiple-choice questions across \num{14} disciplines) as our knowledge-domain training source. Each prompt prepends a fixed block of \num{5} randomly sampled in-context examples from the MMLU-Pro validation split, drawn once and shared across all prompts. We hold out \num{512} prompts (stratified by discipline) for evaluation and train on the remaining. The reward is binary: \num{1} if the model's extracted answer matches the ground-truth option, \num{0} otherwise.

We use the MMLU-Pro test set rather than a separate training corpus because the multi-domain experiments (\S\ref{sec:merging}) are designed to test whether \opd{} can recover multiple specialist teachers simultaneously, not to produce a competitive MMLU-Pro score.

\subsection{IFBench-RLVR: instruction following with rule-based reward}
\label{app:ifbench}

For instruction following we use IFBench-RLVR~\citep{pyatkin2025generalizingverifiableinstructionfollowing}, which pairs natural-language instructions with programmatically verifiable constraints (e.g.\ word count, formatting, keyword inclusion). The training set contains \num{94873} prompts. The reward is the fraction of constraints satisfied per response, thresholded to binary (\num{1} if all constraints pass, \num{0} otherwise). Evaluation uses the IFBench test split.

\section{Prompt templates}
\label{app:prompt-templates}

We list the prompt templates used for each training and evaluation
dataset in this paper. Placeholders (\texttt{\{...\}}) denote per-row
substitutions.

\begin{tcolorbox}[breakable,enhanced, left=\promptleft, right=\promptright, top=2pt, bottom=2pt, enlarge top by=0.1cm, enlarge bottom by=0.1cm, title={\hspace{\ifarxiv 2mm\else 1cm\fi} DAPO-Math-17K (math training) and AIME24/25, AMC23 (math eval)}, fonttitle=\bfseries\small]
\begin{quote}
\begin{lstlisting}[basicstyle=\ttfamily\scriptsize]
<System Prompt>
Please reason step by step, and put your final answer within \boxed{}.

<User Prompt>
{problem statement}
\end{lstlisting}
\end{quote}
\end{tcolorbox}
Reward: rule-based extractor over the value inside \verb|\boxed{}|; a
response is scored $1$ if the extracted string matches the
ground-truth answer verifier and $0$
otherwise.

\begin{tcolorbox}[breakable,enhanced, left=\promptleft, right=\promptright, top=2pt, bottom=2pt, enlarge top by=0.1cm, enlarge bottom by=0.1cm, title={\hspace{\ifarxiv 2mm\else 1cm\fi} LiveCodeBench v1--v5 and CodeForces (code training), LCB v6 (code eval)}, fonttitle=\bfseries\small]
\begin{quote}
\begin{lstlisting}[basicstyle=\ttfamily\scriptsize]
<User Prompt>
You are a coding expert. You will be given a coding problem, and you need
to write a correct Python program that matches the specification and
passes all tests. The time limit is 1 second. You may start by outlining
your thought process. In the end, please provide the complete code in a
code block enclosed with ``` ```.

{problem statement}
\end{lstlisting}
\end{quote}
\end{tcolorbox}
Reward: sandboxed unit-test execution against the row's
\texttt{ground\_truth} \texttt{inputs}/\texttt{outputs}; a response is
scored $1$ if every test case passes within the time limit and $0$
otherwise.

\begin{tcolorbox}[breakable,enhanced, left=\promptleft, right=\promptright, top=2pt, bottom=2pt, enlarge top by=0.1cm, enlarge bottom by=0.1cm, title={\hspace{\ifarxiv 2mm\else 1cm\fi} MMLU-Pro 5-shot (knowledge training and eval)}, fonttitle=\bfseries\small]
\begin{quote}
\begin{lstlisting}[basicstyle=\ttfamily\scriptsize]
<System Prompt>
Given a question and up to ten options, please select the right answer.
Respond in the following format:
<reasoning>
...
</reasoning>
<answer>
...
</answer>

For the answer, only output the letter corresponding to the correct
option (one of A, B, C, D, E, F, G, H, I, or J), and nothing else. Do
not restate the answer text. For example, if the answer is "A", just
output:
<answer>
A
</answer>

<User Prompt>
Here are 5 worked examples showing the required format.

=== Example 1 ===
{example 1 question}
{options 1 (A ... J)}

<reasoning>
{worked reasoning 1}
</reasoning>
<answer>
{letter 1}
</answer>

... (five worked examples in total, drawn once from the MMLU-Pro
validation split and shared across all training prompts) ...

=== Now answer this ===
{target question}
{target options}
\end{lstlisting}
\end{quote}
\end{tcolorbox}
Reward: exact letter match; a response is scored $1$ if the string
inside \verb|<answer>...</answer>| equals the ground-truth letter and
$0$ otherwise.

\begin{tcolorbox}[breakable,enhanced, left=\promptleft, right=\promptright, top=2pt, bottom=2pt, enlarge top by=0.1cm, enlarge bottom by=0.1cm, title={\hspace{\ifarxiv 2mm\else 1cm\fi} IFBench-RLVR (instruction-following training)}, fonttitle=\bfseries\small]
\begin{quote}
\begin{lstlisting}[basicstyle=\ttfamily\scriptsize]
<User Prompt>
{base prompt} {constraint sentence}

Example row (instruction_id = paragraphs:paragraphs):
John goes to the store to buy Slurpees and gives them $20. Slurpees cost
$2 each and he got $8 in change. How many Slurpees did he buy? There
should be 2 paragraphs. Paragraphs are separated with the markdown
divider: ***
\end{lstlisting}
\end{quote}
\end{tcolorbox}
Reward: rule-based; a response is scored $1$ if it satisfies all the
constraints, and $0$ otherwise.

\section{External robustness: per-ablation detail}
\label{app:ablation-detail}

This section gives the detailed description per ablation of \S\ref{sec:ablations}.

\subsection{Training-data difficulty}
\label{sec:difficulty}

This subsection documents the data construction and \texttt{Qwen3-1.7B-Base} results behind the difficulty grid in \S\ref{sec:stratification}. The main in-domain and OOD findings are discussed there; here we provide construction details, the full per-cell OOD breakdown, and the 1.7B-specific observations that inform the broader study.

\paragraph{Data construction.}
We score every prompt in DAPO-Math-17K with \texttt{mean@32} under both the \texttt{Qwen3-1.7B-Base} student and the \texttt{Qwen3-4B} thinking teacher, then bin along both axes ($=\!0$/$>\!0$ for the student; $=\!0$, $(0,\num{0.3}]$, $(\num{0.3},\num{0.6}]$, $>\!\num{0.6}$ for the teacher), yielding eight cells with matched train/in-domain-validation splits; for every subset, we keep the training set size to be 442 and validation set to be 128.  We use a \texttt{Qwen3-1.7B-Base} student in no-think mode for the full eight-cell sweep; the \opd{} recipe in \citet{qwen3} distills in either \texttt{/think} or \texttt{/no\_think} mode, so no-think is a supported regime, and the \num{1.7}B pattern is if anything more severe than the \num{4}B headline.

\begin{figure}[h]
\centering
\includegraphics[width=\linewidth]{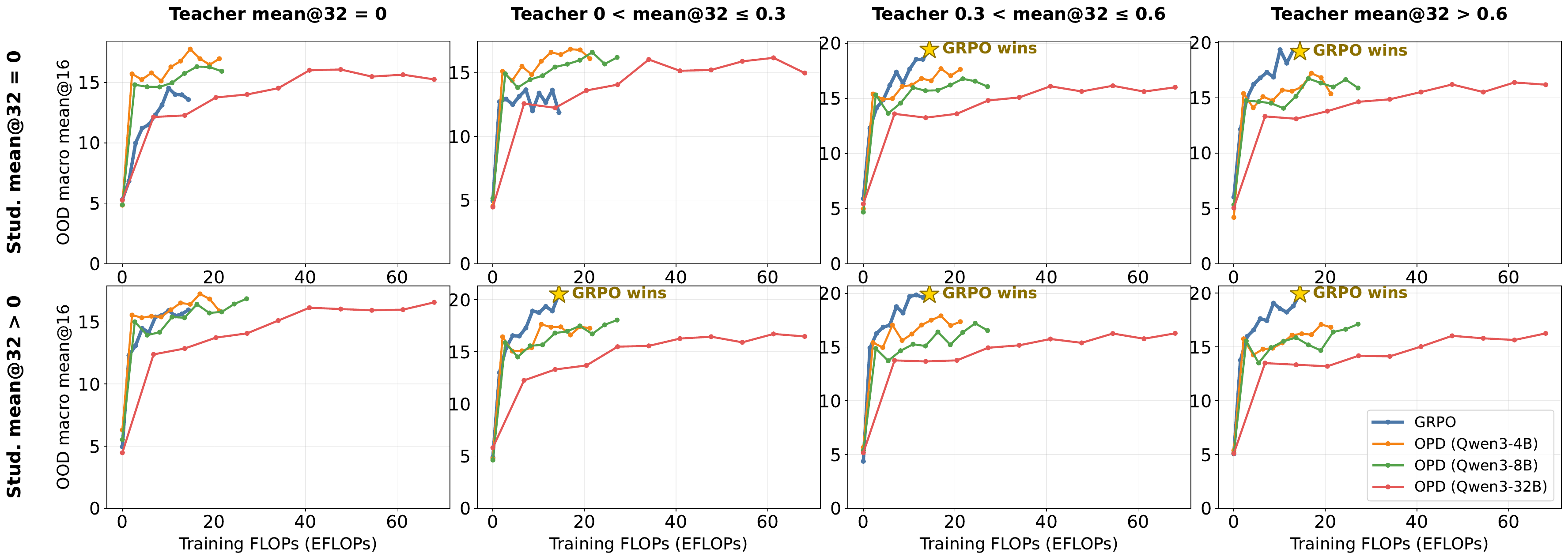}
\caption{\textbf{On OOD math, \opd{} overtakes GRPO only on the hardest training subsets.} Companion to Figure~\ref{fig:difficulty-grid}, with OOD macro \texttt{mean@16} (AIME24, AIME25, AMC23) in place of in-domain accuracy. Rows bin the training problems by the student's \texttt{mean@32} ($=0$, $>0$), and columns bin them by the teacher's \texttt{mean@32} ($=0$, $(0,\num{0.3}]$, $(\num{0.3},\num{0.6}]$, $>\num{0.6}$). \texttt{Qwen3-1.7B-Base} student; \texttt{Qwen3-4B}, \texttt{Qwen3-8B}, and \texttt{Qwen3-32B} teachers; no-think, learning rate \num{2e-6}, batch size \num{64}, \num{8} rollouts per prompt, \num{200} steps. GRPO wins the five cells marked ``GRPO wins'', and \opd{} with the \num{32}B teacher has the lowest peak among the \opd{} runs in each of them. In the other three cells (the teacher $=0$ column and the cell with student $=0$ and teacher $\in(0,\num{0.3}]$), every \opd{} run finishes above GRPO on OOD math, even though GRPO wins the same cells in-domain.}

\label{fig:difficulty-grid-ood}
\end{figure}

\paragraph{OOD performance of \texttt{Qwen3-1.7B-Base}.}
The most striking feature of the \num{1.7}B difficulty grid is the behaviour on the OOD macro (AIME24/25 + AMC23; Figure~\ref{fig:difficulty-grid-ood}). While GRPO's OOD accuracy varies substantially across the eight cells, tracking which training subset it was given, \opd{}'s OOD macro is nearly constant at \pct{15}--\pct{18} regardless of the training cell. This constancy holds across all three teacher sizes (\num{4}B, \num{8}B, \num{32}B) and persists even in the degenerate cell where both student and teacher score zero on the training prompts. In-domain, GRPO wins every cell over the best \opd{} results by a noticeable margin. On OOD, GRPO wins five of the eight cells, but \opd{}-4B and \opd{}-8B finish above GRPO in the three cells where the teacher itself performs badly (the entire teacher~$= 0$ column plus the student-zero/low-teacher cell). This reversal is the opposite of what the dense-signal motivation predicts: \opd{} gains its OOD advantage precisely where the teacher has the least task-solving information to contribute, suggesting that \opd{}'s OOD gain comes from the teacher's general behavior rather than from the task-specific signal on the training prompts (see Appendix~\ref{app:extrapolation}).

\begin{figure}[h]
\centering
\includegraphics[width=\linewidth]{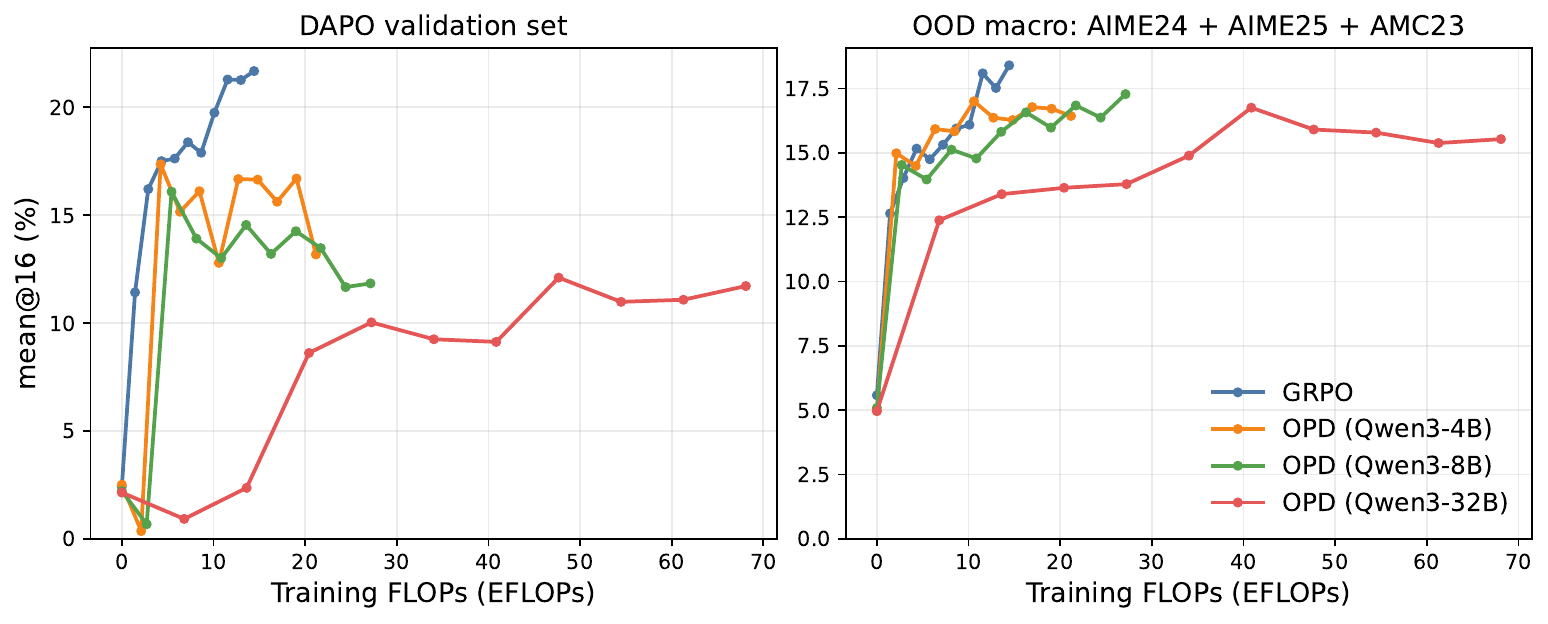}
\caption{\textbf{On the full DAPO-Math-17k training set, GRPO reaches higher accuracy than \opd{} with less compute.} \texttt{Qwen3-1.7B-Base} student, no-think, learning rate \num{2e-6}, batch size \num{64}, \num{8} rollouts per prompt, \num{200} steps; \opd{} uses \texttt{Qwen3-4B}, \texttt{Qwen3-8B}, or \texttt{Qwen3-32B} as the teacher. Points are evaluations every \num{20} steps, plotted against cumulative training FLOPs. \textbf{Left:} \texttt{mean@16} on a held-out DAPO validation split. \textbf{Right:} OOD macro \texttt{mean@16} over AIME24, AIME25, and AMC23. GRPO finishes at about \SI{15}{\exa\flop}. Every \opd{} run stays below GRPO while spending \num{1.4}--\num{4.7}$\times$ as much compute.}

\label{fig:full-dapo}
\end{figure}

\begin{figure}[h]
\centering
\includegraphics[width=\linewidth]{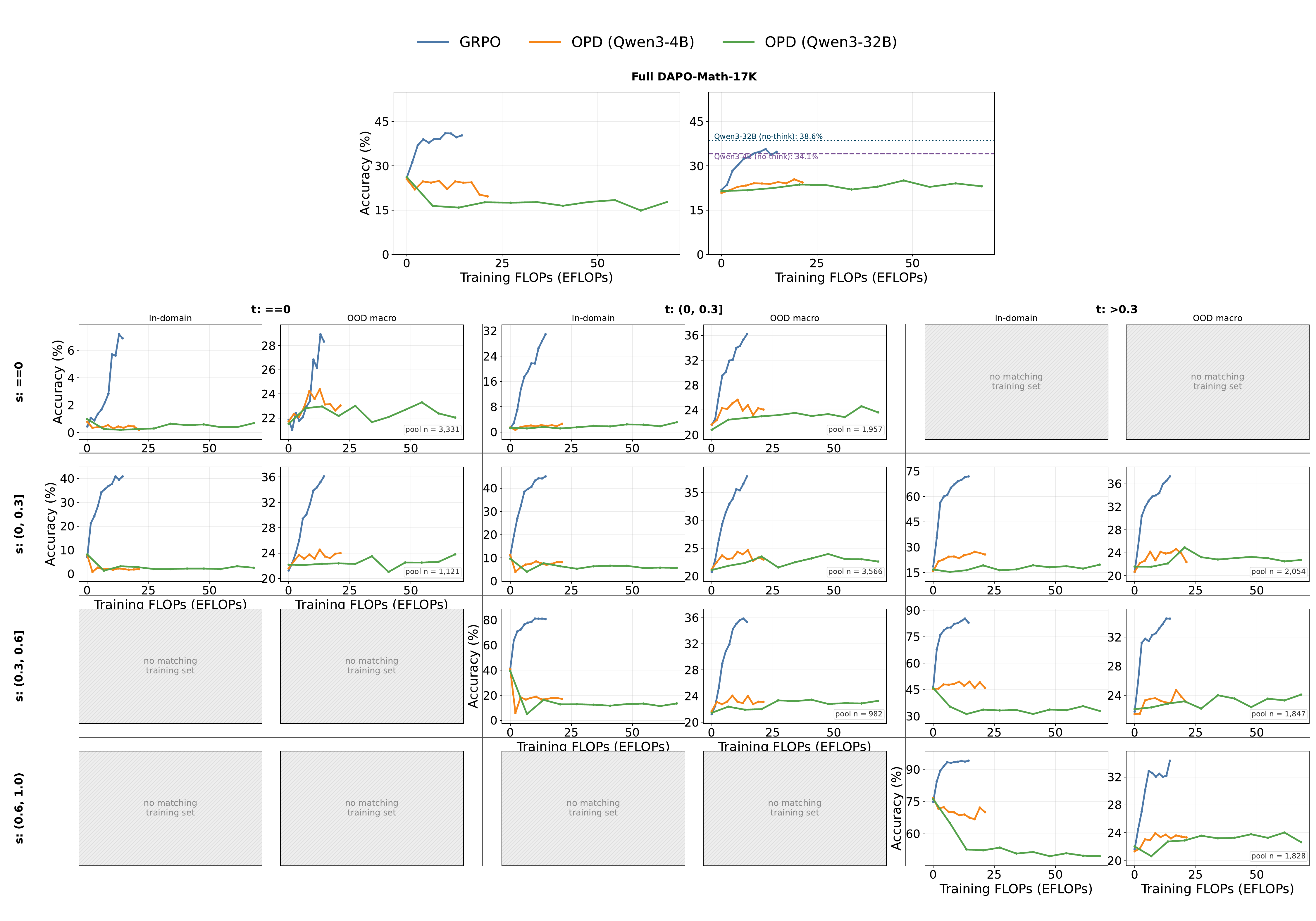}
\caption{\textbf{GRPO also outperforms \opd{} across the difficulty grid with the post-trained \texttt{Qwen3-1.7B} student.} Same stratification as Figure~\ref{fig:difficulty-grid}, with \texttt{Qwen3-1.7B} (no-think) as the student and \texttt{Qwen3-4B} and \texttt{Qwen3-32B} (no-think) as teachers. Rows bin the training problems by the student's \texttt{mean@32} ($=0$, $(0,\num{0.3}]$, $(\num{0.3},\num{0.6}]$, $(\num{0.6},1)$), and column groups bin them by the teacher's \texttt{mean@32} ($=0$, $(0,\num{0.3}]$, $>\num{0.3}$). Each cell shows in-domain validation \texttt{mean@16} (left) and OOD macro \texttt{mean@16} (right) against cumulative training FLOPs; grey cells have too few training problems. The top row trains on the full DAPO-Math-17k, and horizontal lines mark the teachers' OOD accuracy (\texttt{Qwen3-32B}: \pct{38.6}; \texttt{Qwen3-4B}: \pct{34.1}). Each cell trains on \num{442} problems from its subset, and ``pool $n$'' gives the size of the full subset. GRPO outperforms both \opd{} runs in every cell, both in-domain and OOD.}

\label{fig:difficulty-grid-1p7b}
\end{figure}

\paragraph{Difficulty study with \texttt{Qwen3-1.7B}.}
Figure~\ref{fig:difficulty-grid-1p7b} repeats the difficulty study with \texttt{Qwen3-1.7B} (in no-think mode) instead of \texttt{Qwen3-1.7B-Base}, using a finer student-side stratification (\num{4} student bins $\times$ \num{3} teacher bins). We keep the amount of training/validation sets the same as what's used for \texttt{Qwen3-1.7B-Base}. The performance gap between GRPO and \opd{} is even larger. In all training sets, GRPO performance increases on both in-domain and out-of-domain. In contrast, \opd{} with external teachers makes student performance even worse. 

\subsection{Learning rate}
\label{app:lr-detail}\label{sec:lr}

\begin{figure}[h]
\centering
\includegraphics[width=\linewidth]{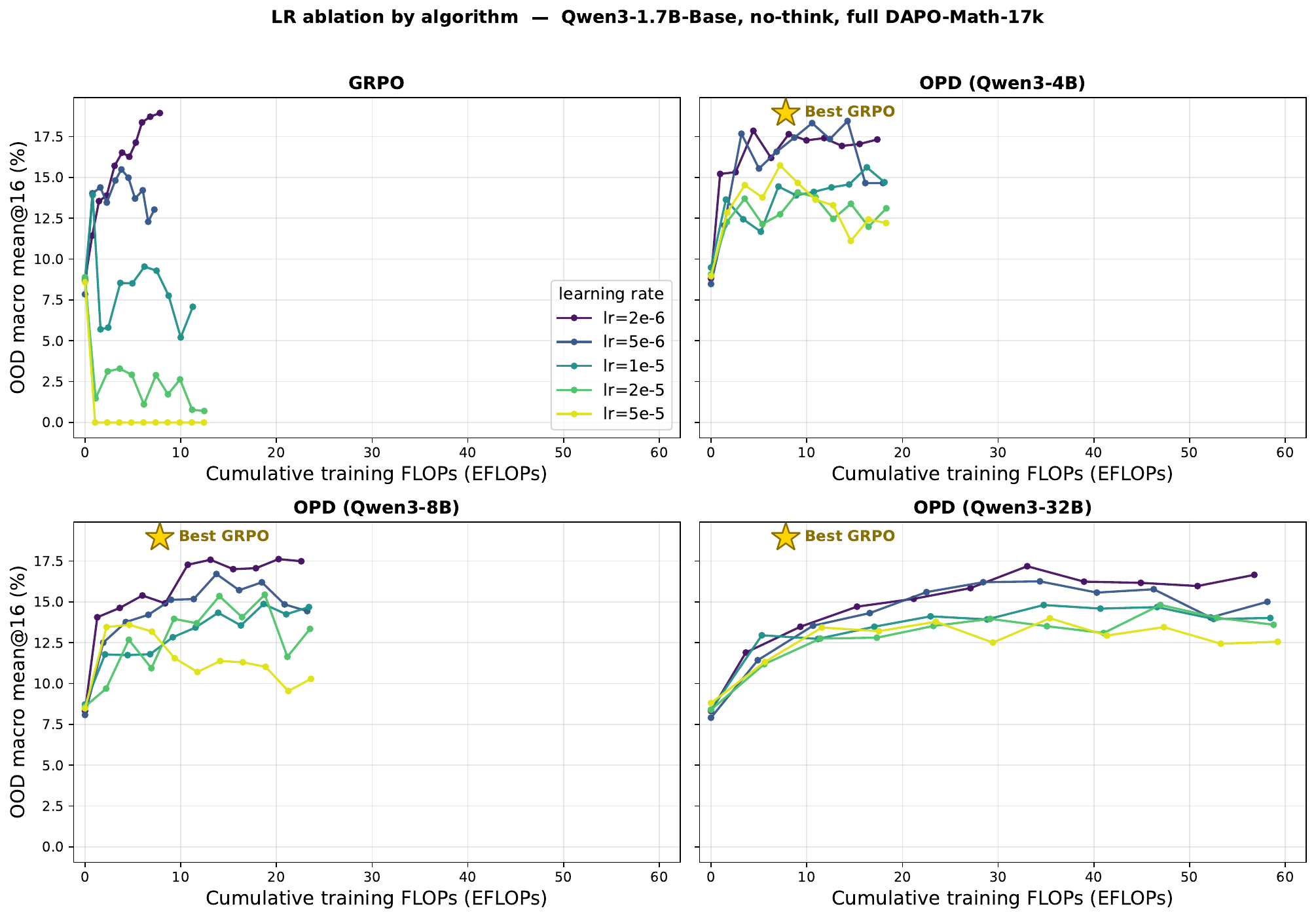}
\caption{\textbf{No learning rate lets \opd{} reach the best GRPO accuracy.} \texttt{Qwen3-1.7B-Base} student, no-think, full DAPO-Math-17k; OOD macro \texttt{mean@16} against cumulative training FLOPs for learning rates \num{2e-6}, \num{5e-6}, \num{1e-5}, \num{2e-5}, and \num{5e-5}. One panel per method: GRPO (top left) and \opd{} with the \texttt{Qwen3-4B}, \texttt{Qwen3-8B}, and \texttt{Qwen3-32B} teachers. The star in each \opd{} panel marks the best GRPO accuracy (learning rate \num{2e-6}, about \pct{19}). GRPO degrades as the learning rate grows and collapses at \num{5e-5}. \opd{} is more stable across learning rates, but its best configuration (\texttt{Qwen3-4B} teacher at \num{5e-6}, about \pct{18.4}) still ends below the star.}

\label{fig:lr-sweep}
\end{figure}

We swept the learning rate over five values (\num{2e-6}, \num{5e-6}, \num{1e-5}, \num{2e-5}, \num{5e-5}) for GRPO and for all three \opd{} teacher sizes on DAPO-Math-17K, no-think (Figure~\ref{fig:lr-sweep}). \opd{} is more lenient: \opd{}-4B's best-over-training OOD macro spans \num{4.3} points across rates (\num{14.1}--\num{18.4}), whereas GRPO's spans \num{10.3} points (\num{8.6}--\num{18.9}). GRPO is more sensitive to the learning rate, but the best GRPO configuration outperforms the best \opd{} in both accuracy and compute efficiency: GRPO at lr \num{2e-6} reaches \pct{18.9} OOD macro at \SI{7.8}{\exa\flop}, while \opd{}'s best (\opd{}-4B at lr \num{5e-6}) reaches \pct{18.4} at \SI{14.3}{\exa\flop}, requiring \num{1.8}$\times$ the compute for a lower score.

\subsection{Batch size}
\label{app:batch-detail}

\begin{figure}[h]
\centering
\includegraphics[width=\linewidth]{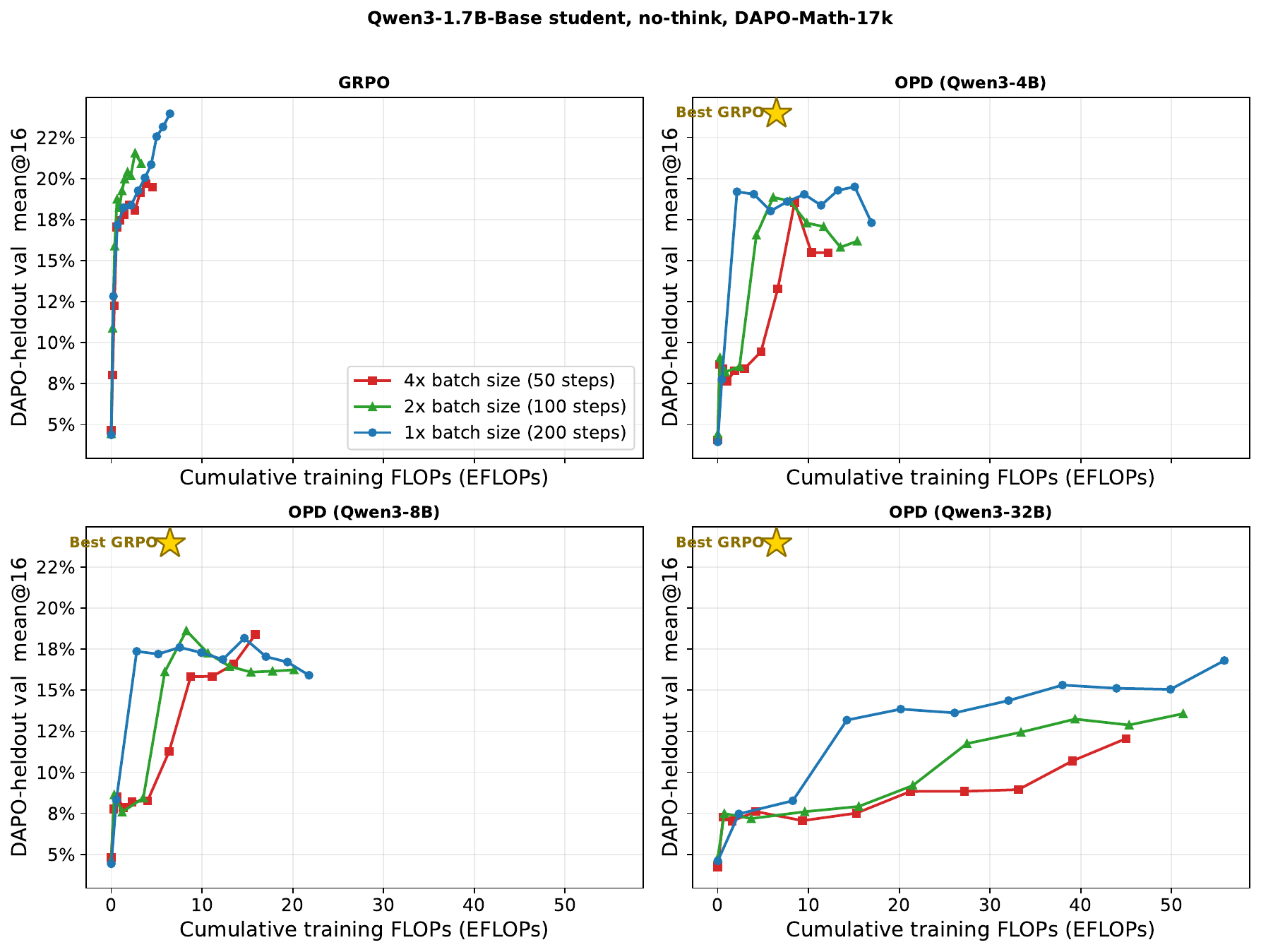}
\caption{\textbf{At a fixed prompt budget, no batch size lets \opd{} catch up with GRPO.} Every configuration trains on \num{12800} prompts: batch size \num{64} for \num{200} steps (1$\times$), \num{128} for \num{100} steps (2$\times$), or \num{256} for \num{50} steps (4$\times$). \texttt{Qwen3-1.7B-Base} student, no-think, DAPO-Math-17k, learning rate \num{2e-6}; DAPO validation \texttt{mean@16} against cumulative training FLOPs, one panel per method. The smallest batch is the best or tied-best setting for GRPO and for \opd{} with the \num{4}B and \num{32}B teachers, while the \num{8}B teacher shows no consistent ordering. No \opd{} run reaches the best GRPO accuracy (star) at any batch size.}

\label{fig:bsz}
\end{figure}

Batch size is another key hyperparameter for model tuning: it controls the noise level of gradient estimates. To isolate its effect while keeping the total training data fixed, we ran three batch-size
configurations -- \num{64} for $\num{200}$ steps, \num{128} for $\num{100}$ steps,
and \num{256} at $\num{50}$ steps -- for GRPO and for \opd{} under each of the
three teacher sizes, all on DAPO-Math-17K with a \texttt{Qwen3-1.7B-Base} student in
no-think mode at learning rate \num{2e-6} (Figure~\ref{fig:bsz}). Smaller
batches help both methods: GRPO climbs from \num{19.5} to \num{20.9}
to \num{24.0} (in-domain validation mean@16, \%) as the batch shrinks from
\num{256} to \num{128} to \num{64}; \opd{}-4B follows the same staircase from
\num{16.0} to \num{16.2} to \num{17.3}. The gain is roughly parallel, so the
relative ordering GRPO $>$ \opd{} holds at every batch size. For the two larger
teachers (\num{8}B, \num{32}B) the pattern is noisier but still favours small
batch, and at no batch size does any \opd{} run overtake GRPO.

\subsection{Reverse-KL estimator variant}
\label{app:rkl-estimator}

The default \opd{} loss in our experiments uses the \emph{closed-form top-$k$} reverse KL: at each token position, the student's top-$k{=}\num{16}$ tokens define a support set, both the student and teacher distributions are renormalized onto that set, and the KL is computed exactly over the $k$ tokens (\S\ref{sec:setup}). An alternative is the \emph{sampled-token} estimator, which evaluates the reverse KL at only the single token the student actually produced. The $k_1$ estimator\footnote{\url{http://joschu.net/blog/kl-approx.html}} replaces the inner sum over the top-$k$ support with the importance-weighted log-ratio at the single token $o_{i,t}$ sampled from the rollout policy $\pi_{\text{old}}$:
\begin{equation*}
\mathcal{L}_{\text{OPD}}^{{k_1}}(\theta) = \mathbb{E}_{q,\{o_i\}}\!\left[\frac{1}{G}\sum_{i=1}^{G} \sum_{t=1}^{|o_i|} \frac{\pi_\theta(o_{i,t}\mid q, o_{i,<t})}{\pi_{\text{old}}(o_{i,t}\mid q, o_{i,<t})} \Big[\log\pi_{\text{old}}(o_{i,t}\mid q, o_{i,<t}) - \log\pi_T(o_{i,t}\mid q, o_{i,<t})\Big]\right].
\end{equation*}
The ratio $\pi_\theta/\pi_{\text{old}}$ corrects for off-policy sampling, and the per-token signal $\log\pi_{\text{old}} - \log\pi_T$ is negative where the teacher assigns higher probability than the rollout policy. This estimator needs only the student, rollout, and teacher log-probabilities at the sampled token, avoiding the top-$k$ renormalization entirely.

\begin{figure}[h]
\centering
\includegraphics[width=\linewidth]{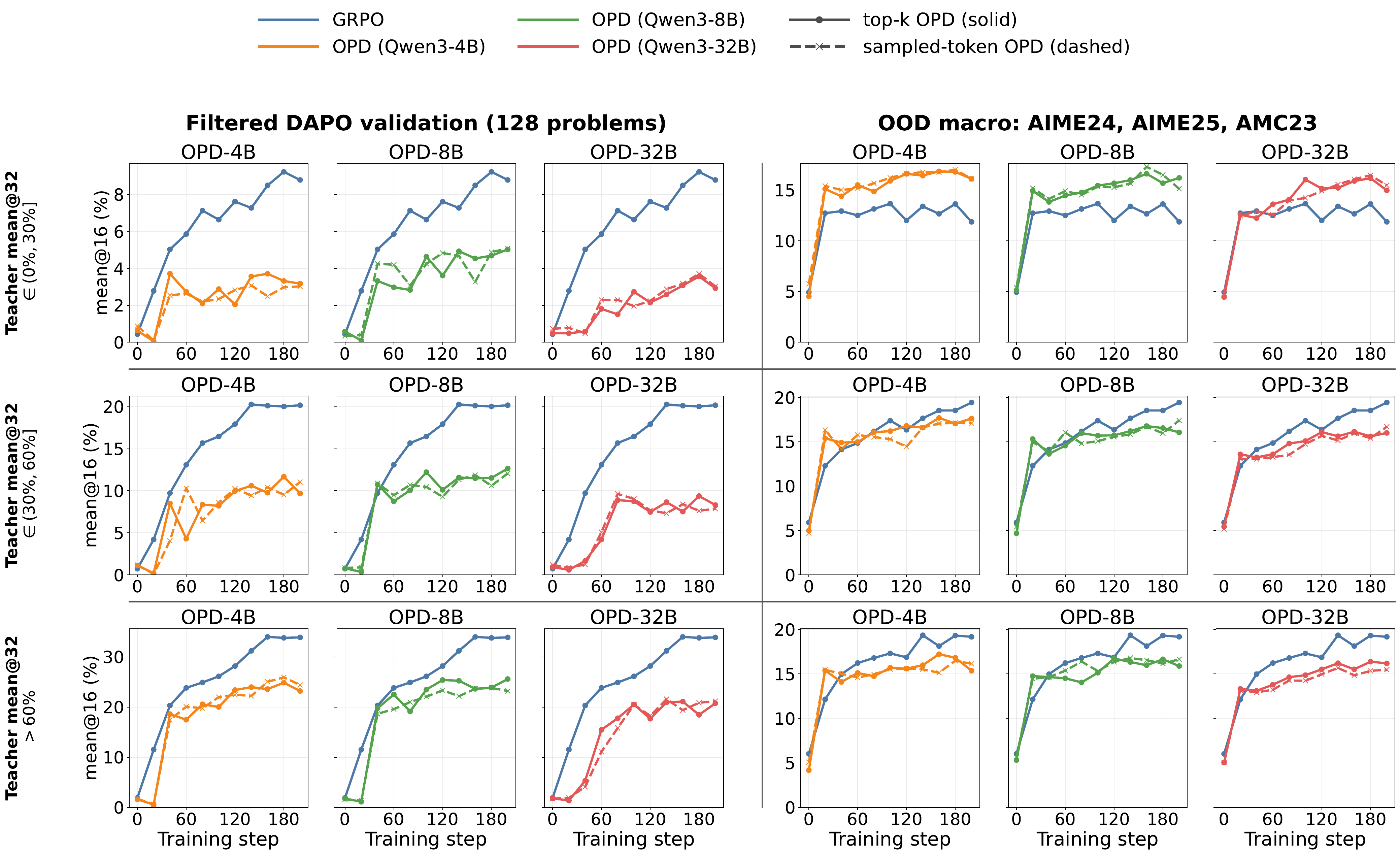}
\caption{\textbf{The reverse-KL estimator barely changes \opd{} accuracy.} The two \opd{} losses differ only in the reverse-KL estimator. Solid lines use the closed-form reverse KL over the student's top-$k{=}16$ tokens; dashed lines use the sampled-token $k_1$ estimator at the token the student generated. \texttt{Qwen3-1.7B-Base} student, no-think, DAPO-Math-17k, learning rate \num{2e-6}, with GRPO shown for reference. Rows are training subsets that the student never solves, binned by the teacher's \texttt{mean@32} ($(0,\pct{30}]$, $(\pct{30},\pct{60}]$, $>\pct{60}$). Columns are the \texttt{Qwen3-4B}, \texttt{Qwen3-8B}, and \texttt{Qwen3-32B} teachers. \textbf{Left block:} \texttt{mean@16} on a filtered DAPO validation set of \num{128} problems. \textbf{Right block:} OOD macro \texttt{mean@16}. The $x$-axis is the training step. The two estimators trace nearly the same curves in every cell.}

\label{fig:topk-supsamp}
\end{figure}

Figure~\ref{fig:topk-supsamp} compares the two estimators under the same recipe (\texttt{Qwen3-1.7B-Base}, no-think, DAPO-Math-17K, lr $\num{2e-6}$, batch \num{64}, \num{200} steps). The two variants trace essentially the same accuracy curves across all three teacher sizes and across difficulty bands, so the choice of reverse-KL support set is not what determines \opd{}'s loss to GRPO. This is consistent with the observation by \citet{kimi2026a260724653} that the estimator choice has minimal impact in practice.

\subsection{Thinking mode}
\label{app:teacher-detail}\label{sec:teacher}\label{app:ratio-detail}\label{sec:ratio}

\begin{figure}[h]
\centering
\includegraphics[width=\linewidth]{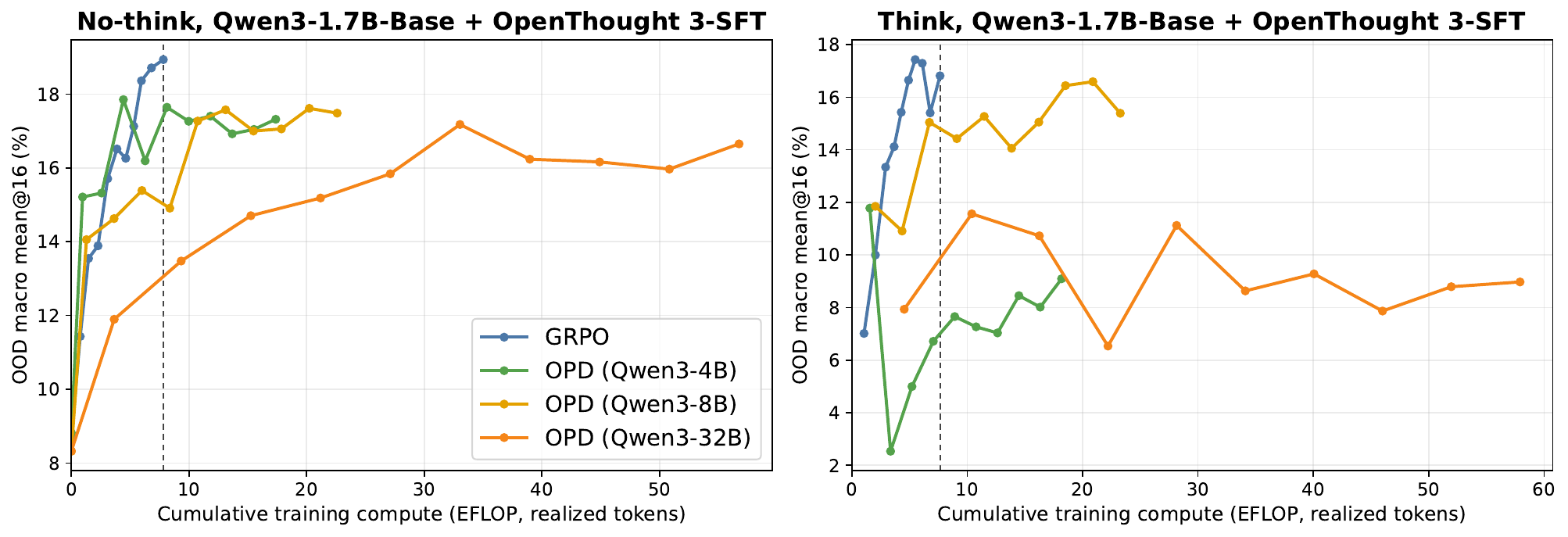}
\caption{\textbf{Thinking mode hurts \opd{} much more than GRPO.} \texttt{Qwen3-1.7B-Base} student initialized from OpenThoughts-3 SFT, DAPO-Math-17k; OOD macro \texttt{mean@16} against cumulative training FLOPs. \textbf{Left:} student and teachers in no-think mode. \textbf{Right:} student and teachers in thinking mode. The vertical dashed line marks the compute at which GRPO finishes. Switching from no-think to thinking lowers the final accuracy of GRPO by \num{2.1} points (\pct{18.9} to \pct{16.8}) and of \opd{} with the \num{8}B teacher by \num{2.1} points (\pct{17.5} to \pct{15.4}). \opd{} with the \num{4}B and \num{32}B teachers loses \num{8.2} and \num{7.7} points, ending at \pct{9.1} and \pct{9.0}.}

\label{fig:think-vs-nothink}
\end{figure}

Figure~\ref{fig:think-vs-nothink} isolates the effect of the teacher's thinking mode. All runs use the same student (\texttt{Qwen3-1.7B-Base} with OpenThoughts-3-SFT initialization) and the same data (DAPO-Math-17K); only the teacher decoding mode changes. In no-think mode (left panel), GRPO reaches roughly \pct{19} OOD macro and the three \opd{} variants (\num{4}B, \num{8}B, and \num{32}B teachers) cluster in a narrow \pct{16}--\pct{17.5} band, consistent with the main-text pattern. Switching to think mode (right panel) costs GRPO about \num{2.1} points, but the effect on \opd{} varies sharply with teacher size: \opd{}-8B loses a comparable \num{2.1} points and stays within \num{1.5} points of GRPO, whereas \opd{}-4B drops \num{8.2} points and \opd{}-32B drops \num{7.7} points, leaving both at roughly half the GRPO level.

Two observations follow. First, thinking mode hurts \opd{} more than GRPO across most teacher sizes, consistent with the hypothesis that long chains of thought amplify the distribution mismatch between student rollouts and teacher token-level supervision. Second, teacher capacity does not rescue the gap: the \num{32}B teacher fares worse than the \num{8}B teacher in think mode, not better, so the fragility is not a matter of teacher scale. We note that no-think and think runs differ in decoding temperature and prompt format, so the two panels are not a controlled comparison with each other; the GRPO-vs-\opd{} contrast within each panel is controlled.

\subsection{Student initialization}
\label{sec:init}

A natural hypothesis is that the base model must first learn to produce rollouts in the right format before \opd{} can distill from them effectively: if the student's on-policy outputs are too far from the teacher's distribution, token-level supervision may have little to work with. SFT cold-starting on high-quality traces (e.g.\ OpenThoughts-3) could therefore be important for \opd{} to succeed. We test this hypothesis at two student scales.

\begin{figure}[t!]
\centering
\includegraphics[width=\linewidth]{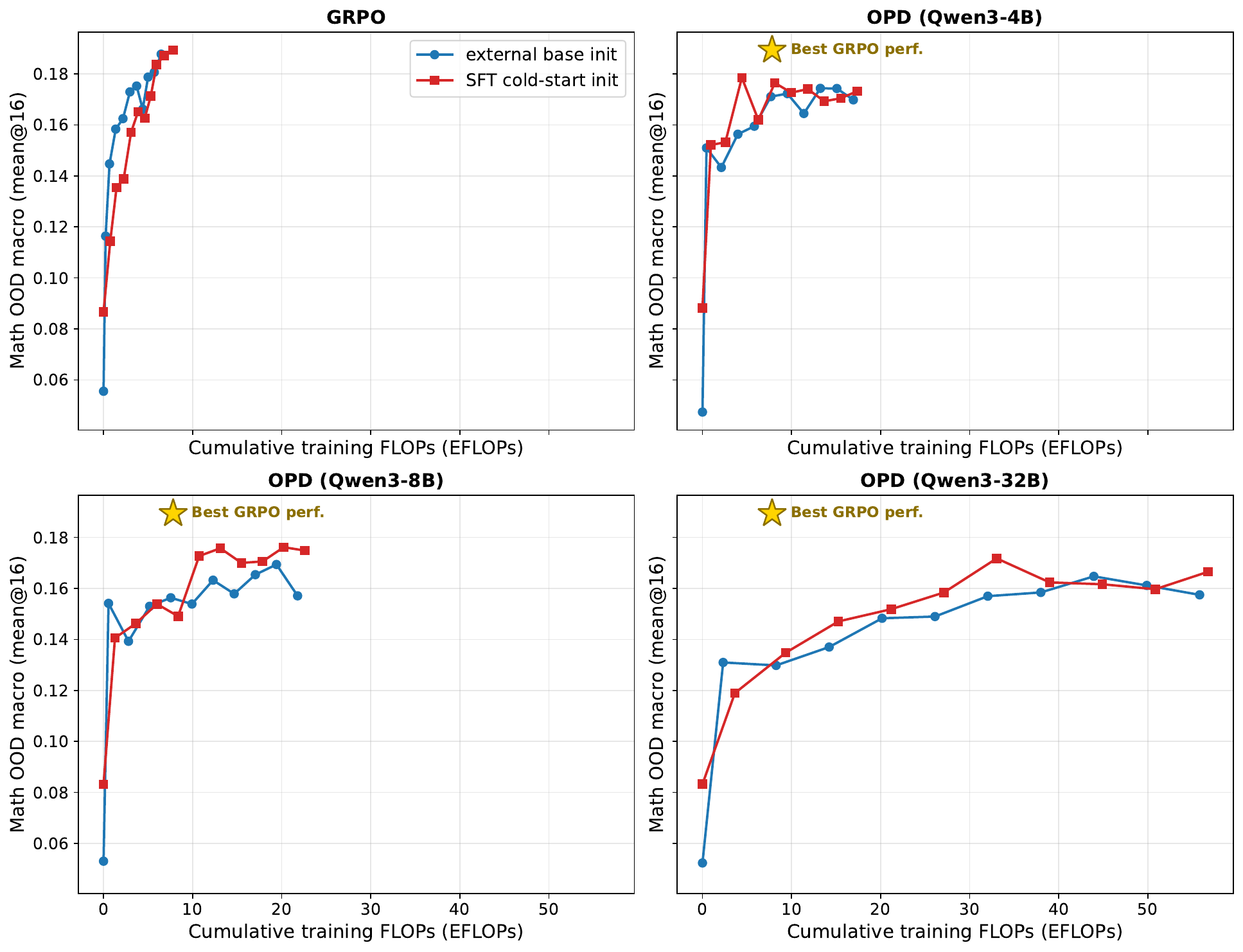}
\caption{\textbf{An SFT cold start does not change the ordering of GRPO and \opd{} (\texttt{Qwen3-1.7B-Base}, no-think).} OOD macro \texttt{mean@16} (AIME24, AIME25, AMC23) against the training step, one panel per method. Each panel compares the base initialization (blue) with an OpenThoughts-3 SFT initialization (red; SFT learning rate \num{1e-5}, step \num{3000}). All runs use no-think mode, learning rate \num{2e-6}, DAPO-Math-17k, batch size \num{64}, \num{200} steps, and a \num{7168}-token response limit. The SFT initialization starts higher (about \pct{8.5} versus \pct{5.5}), but the difference disappears by step \num{20}. GRPO reaches about \pct{19} under both initializations, and every \opd{} run stays below the best GRPO accuracy (star).}

\label{fig:sft-init-1p7b-nothink}
\end{figure}

\paragraph{1.7B-Base, no-think mode.}
Figure~\ref{fig:sft-init-1p7b-nothink} compares \texttt{Qwen3-1.7B-Base} initialized from the external pretrained checkpoint against the same model initialized from an OpenThoughts-3-SFT cold start (lr \num{1e-5}, step \num{3000}). Both runs use no-think mode with matched hyperparameters (lr \num{2e-6}, $k{=}\num{32}$ rollouts, batch \num{64}, \num{200} steps). The SFT initialization provides a head start at step~\num{0}, but GRPO overtakes all \opd{} variants under both initializations. The SFT cold start does not change the conclusion.

\begin{figure}[!t]
\centering
\includegraphics[width=\linewidth]{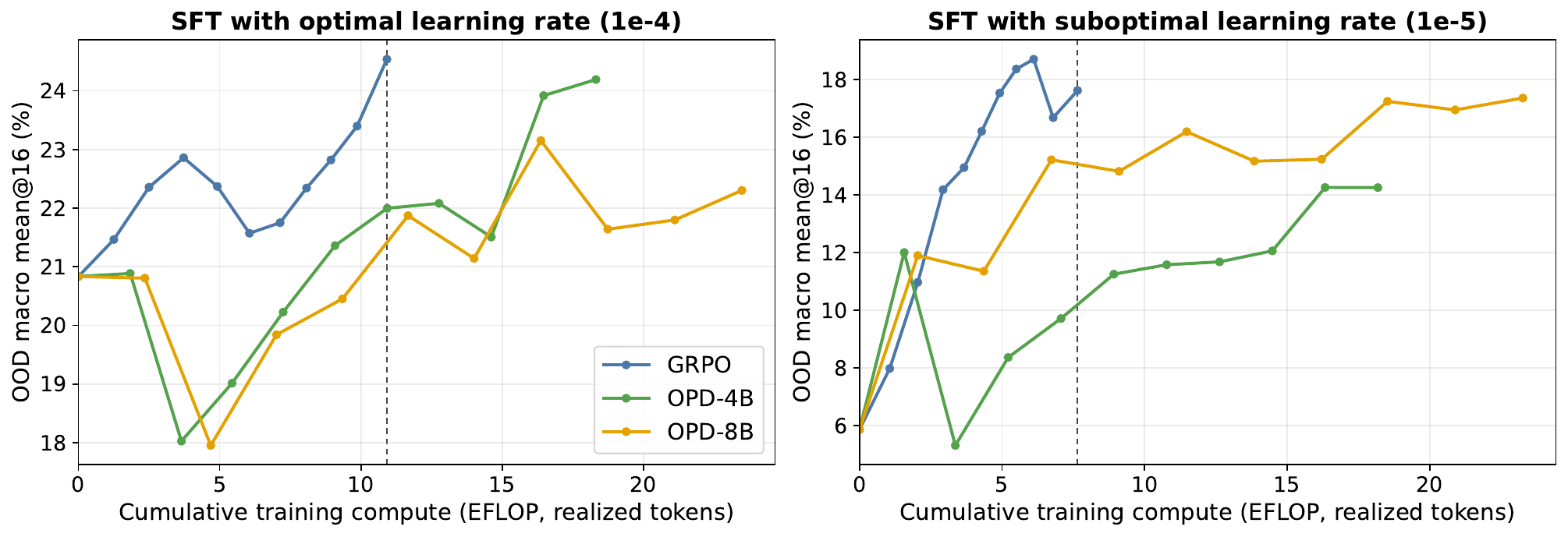}
\caption{\textbf{With thinking enabled, GRPO still leads \opd{} after an SFT cold start (\texttt{Qwen3-1.7B-Base}).} OOD macro \texttt{mean@16} (AIME24, AIME25, AMC23) against cumulative training FLOPs, with \opd{} using \texttt{Qwen3-4B} or \texttt{Qwen3-8B} as the teacher. Both panels use thinking mode, DAPO-Math-17k, batch size \num{64}, \num{8} rollouts per prompt, a \num{7168}-token training limit, and a \num{30720}-token evaluation limit. The vertical dashed line marks the compute at which GRPO finishes. \textbf{Left:} SFT initialization at the best SFT learning rate (\num{1e-4}). \textbf{Right:} SFT initialization at a lower SFT learning rate (\num{1e-5}). GRPO reaches \pct{24.5} (left) and \pct{18.7} (right). On the left, \opd{} with the \num{4}B teacher comes within \num{0.3} points of GRPO after about \num{1.7}$\times$ the compute. On the right, no \opd{} run reaches GRPO.}

\label{fig:sft-init-1p7b}
\end{figure}

\paragraph{1.7B-Base, think mode.}
We next test two SFT cold starts with thinking enabled: lr \num{1e-5} (the same checkpoint as above) and lr \num{1e-4} (the optimal SFT learning rate). Figure~\ref{fig:sft-init-1p7b} plots OOD macro against training FLOPs for both. Under both cold starts the within-panel ordering holds: GRPO sits above every \opd{} run per FLOP. Switching to a stronger SFT initialization and enabling thinking does not change the ranking.

\begin{figure}[!t]
\centering
\includegraphics[width=\linewidth]{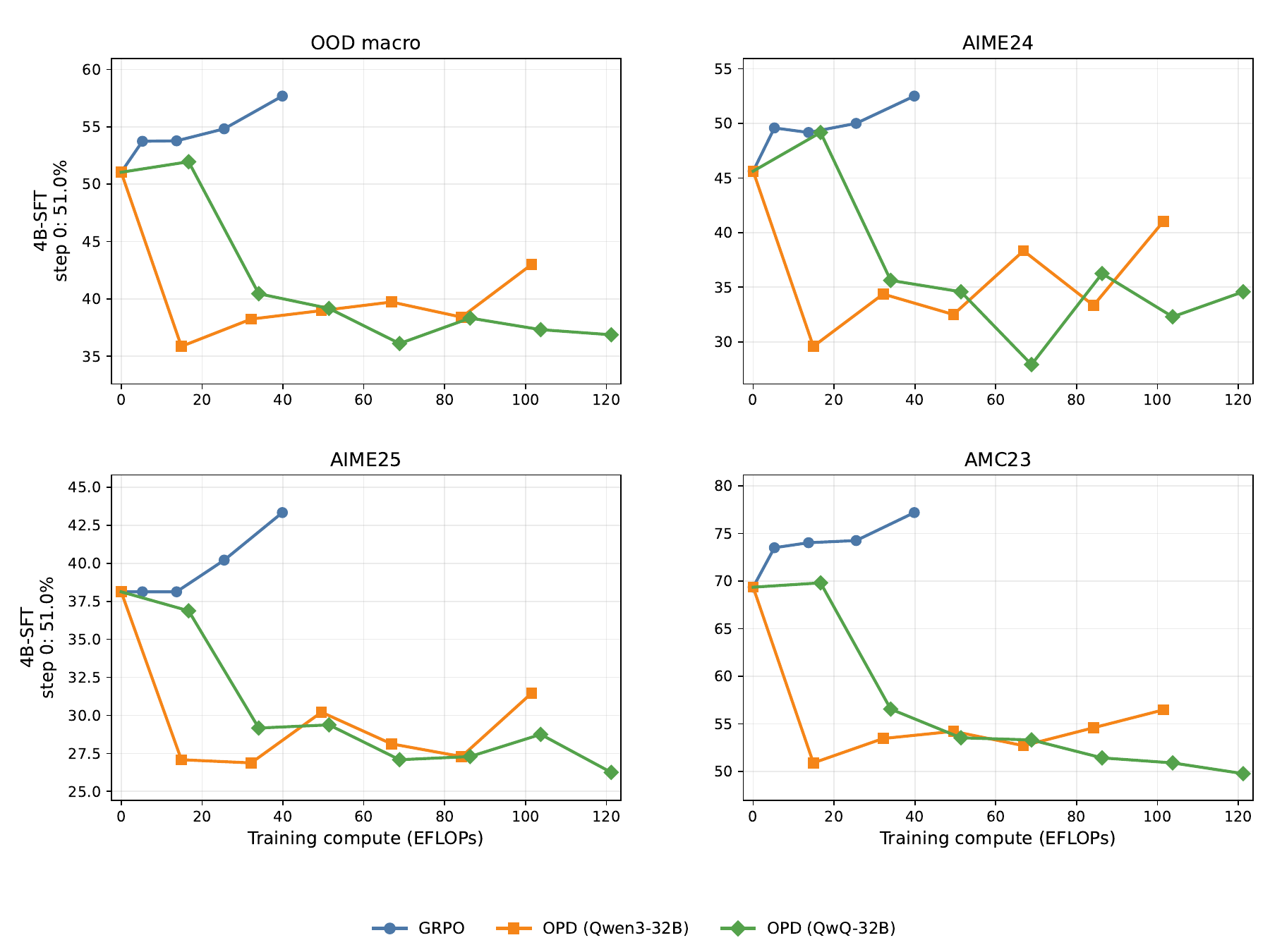}
\caption{\textbf{From a strong SFT initialization, \opd{} falls below the starting student while GRPO improves it (\texttt{Qwen3-4B-Base}).} The student is \texttt{Qwen3-4B-Base} after OpenThoughts-3 SFT (SFT learning rate \num{1e-4}), which starts at \pct{51.0} OOD macro. \opd{} uses \texttt{Qwen3-32B} or \texttt{QwQ-32B} as the teacher; \texttt{QwQ-32B} generated the SFT corpus, so its outputs match the student's starting distribution. Panels show the OOD macro \texttt{mean@16} and its components (AIME24, AIME25, AMC23) against cumulative training FLOPs. GRPO improves to \pct{57.7} within about \SI{40}{\exa\flop}. Both \opd{} runs drop to about \pct{36} OOD macro and remain \num{8}--\num{15} points below the starting student through \SI{120}{\exa\flop}. The \texttt{QwQ-32B} teacher only delays the drop.}

\label{fig:sft-init-4b}
\end{figure}

\paragraph{4B-Base with SFT cold start.}
We scale the test to \texttt{Qwen3-4B-Base} with thinking enabled, initialized from OpenThoughts-3-SFT (lr \num{1e-4}, the optimal learning rate for this model), and increase the response maximum from \num{7168} to \num{16384} tokens. In addition to the standard \opd{}-32B teacher, we include \texttt{QwQ-32B}, the model whose rollouts generated the OpenThoughts corpus used for the SFT cold start, to accommodate teacher similarity with the SFT-initialized student. Figure~\ref{fig:sft-init-4b} shows GRPO outperforming both \opd{} teachers across all four evaluation benchmarks. The \texttt{QwQ-32B} teacher does not recover the gap, and both \opd{} variants decline over training while GRPO continues to improve.

\subsection{Student model sizes}
\label{app:model-sizes}

\begin{figure}[!t]
\centering
\includegraphics[width=\linewidth]{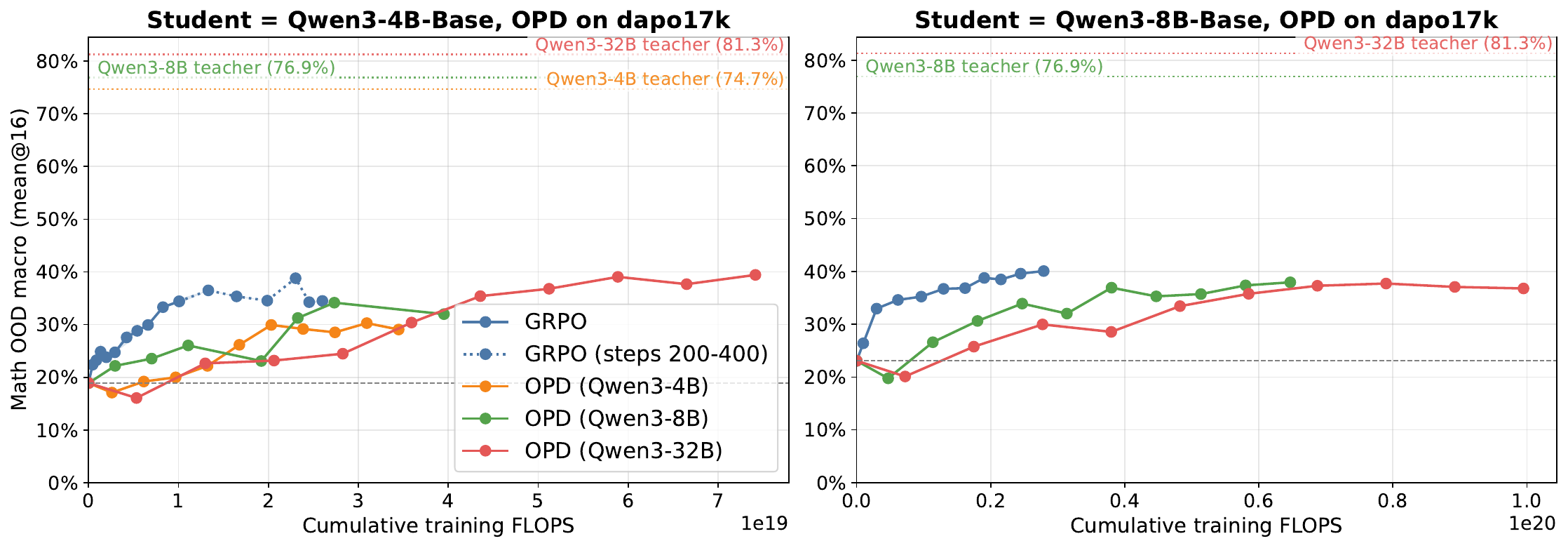}
\caption{\textbf{External thinking teachers, 4B and 8B students.}
Two panels, one per student size. Curves are training trajectories on
DAPO-Math-17K for GRPO and for \opd{} against external
\texttt{Qwen3-}\{\texttt{4B}, \texttt{8B}, \texttt{32B}\} thinking teachers; $x$ is cumulative training
FLOPs, $y$ is OOD macro \texttt{mean@16} (AIME24 + AIME25 + AMC23).
GRPO curves:
solid up to optimizer step \num{200}, dotted for steps \num{200}--\num{400}
(the \opd{} runs trained for at most \num{200} steps, so the solid segments
are the matched-step comparison). No \opd{} run reaches its teacher's
accuracy within the compute budget, and a larger external teacher
does not close the gap. }
\label{fig:tce-off-shelf}
\end{figure}

The negative result in \S\ref{sec:main} is established on
\texttt{Qwen3-1.7B-Base}. A natural concern is that the 1.7B student is simply
too small to benefit from a strong teacher and that the gap would close
at larger student scale. We test this by running the same external
comparison at \texttt{Qwen3-4B-Base} and \texttt{Qwen3-8B-Base}
(Figure~\ref{fig:tce-off-shelf}). At both scales, no external \opd{}
run reaches its teacher's accuracy within the compute budget, and a
larger teacher does not improve compute efficiency: the 32B teacher
curves sit close to (and sometimes below) the same-size teacher curves.
The negative result is not a small-student artifact.

\subsection{Training domain}
\label{app:training-domain}

\begin{figure}[!t]
\centering
\includegraphics[width=\linewidth]{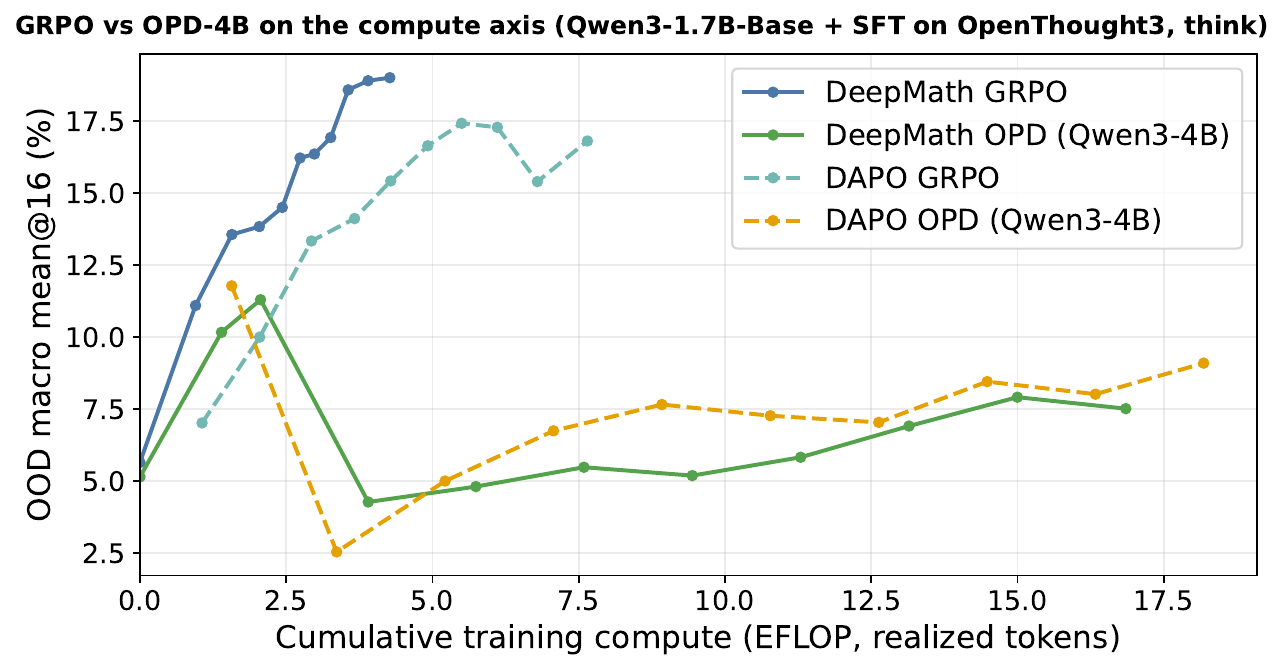}
\caption{\textbf{Switching the training set from DAPO-Math-17k to DeepMath-103k does not change the ordering.} \texttt{Qwen3-1.7B-Base} student after OpenThoughts-3 SFT, thinking mode; \opd{} uses the \texttt{Qwen3-4B} teacher. OOD macro \texttt{mean@16} against cumulative training FLOPs; solid lines train on DeepMath-103k and dashed lines on DAPO-Math-17k. On both training sets, \opd{} first rises, then drops to \pct{2.5}--\pct{4.3}, and recovers only to \pct{7.5}--\pct{9.1} by about \SI{17}{\exa\flop}. GRPO reaches \pct{19.0} (DeepMath) and \pct{17.4} (DAPO) within \SI{8}{\exa\flop}.}

\label{fig:flops-domain}
\end{figure}

\paragraph{Math.}
Our default math training set is DAPO-Math-17K. We also run on
DeepMath-103k, the dataset used by Thinking Machines
Lab~\citep{tml2025opd} for their \opd{} experiments. Swapping
DAPO-Math for DeepMath does not change the ordering; it widens it, with
GRPO ahead by \num{11.5} points of \texttt{mean@16} on DeepMath versus
\num{7.7} on DAPO (Figure~\ref{fig:flops-domain}).

\begin{figure}[!t]
\centering
\includegraphics[width=\linewidth]{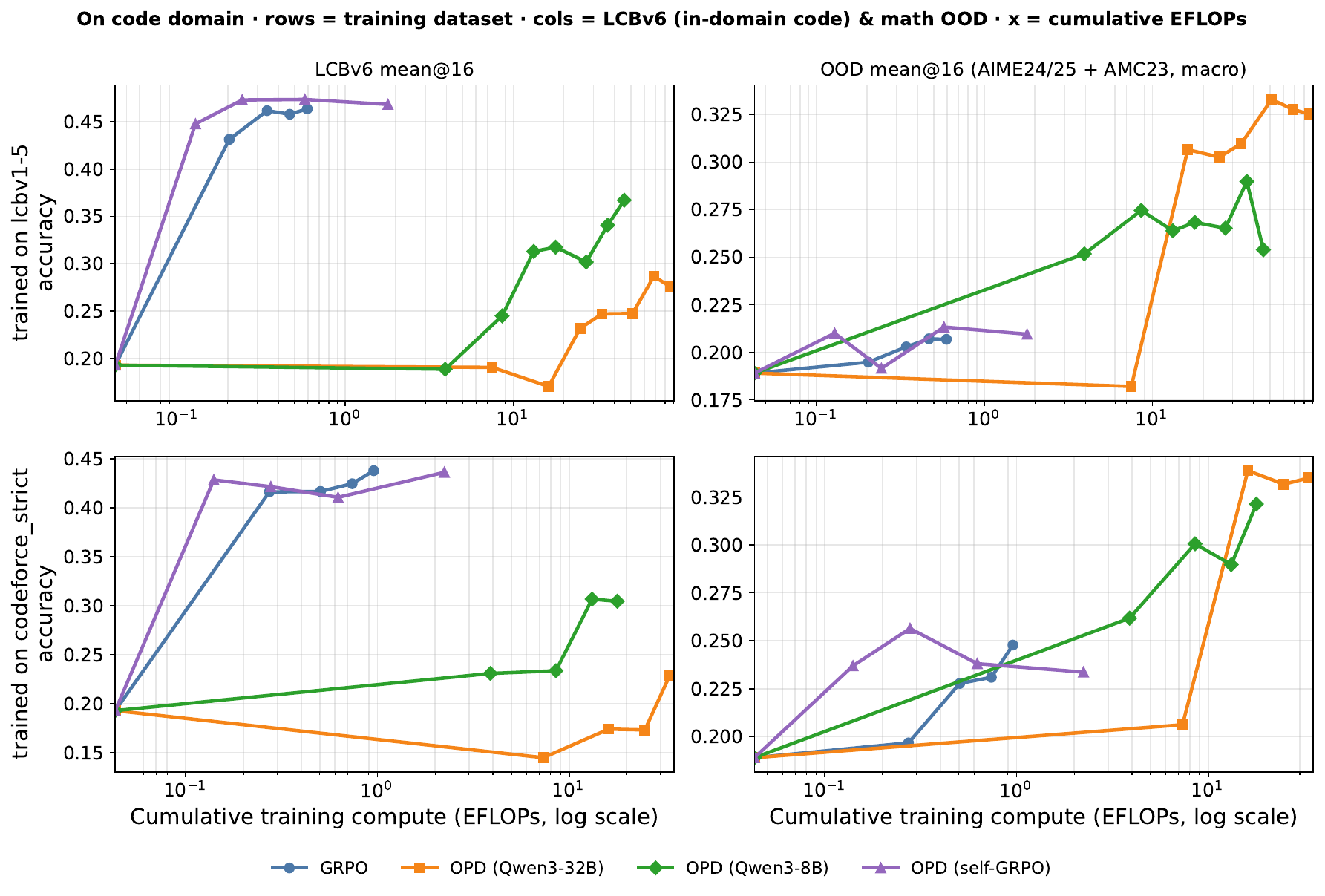}
\caption{\textbf{Code domain training: in-domain collapse, out-of-domain math
transfer.} \texttt{Qwen3-4B-Base} student trained on LiveCodeBench-v1--v5
(top row) and strict CodeForces (bottom row) with GRPO or external
\opd{}-\num{32}B, lr \num{2e-6}, \num{16384}-token response maximum; $x$-axis is
cumulative training compute (EFLOPs). \textbf{Left column:} LCBv6
\texttt{mean@16} (in-domain code). \opd{} (\texttt{Qwen3-32B}) loses to GRPO at every
evaluated compute budget on both training sets.
\textbf{Right column:} math OOD macro \texttt{mean@16} on AIME24 + AIME25 +
AMC23. Despite training only on code, \opd{}-\num{32}B reaches
${\approx}\pct{31}$--\pct{34} math OOD, beating the code-trained GRPO by
roughly \num{10} points on both training sets --- an additional evidence for
\opd{}'s cross-domain transfer (Appendix~\ref{app:extrapolation}).}
\label{fig:code-math-extrap}
\end{figure}

\paragraph{Code.}
We train the same \texttt{Qwen3-4B-Base} student on two code datasets
(LiveCodeBench v1--v5 and a high-quality CodeForces set we carve from
\texttt{open-r1/codeforces}; Appendix~\ref{app:codeforces}) with GRPO
and \opd{} under external \num{8}B and \num{32}B teachers. The
ordering observed on math holds on code: GRPO outperforms both \opd{}
variants on LCBv6 at every evaluated compute budget on both training
sets (Figure~\ref{fig:code-math-extrap}, left column).

\subsection{Replication across RL frameworks}
\label{app:framework-detail}\label{sec:framework}

% We replicated the comparison across three RL stacks. The first two are
% two generations of our \texttt{verl}-based training pipeline. Our headline
% infrastructure uses a slightly older \texttt{verl} generation in which we
% implement the reverse-KL \opd{} objective ourselves. We also re-ran
% the comparison in the latest \texttt{verl} release, which natively supports
% reverse-KL \opd{} with a sampled-token estimator. Hyperparameters are
% controlled between the two stacks; the only difference is the
% reverse-KL estimator (closed-form top-$k$ in our custom implementation
% against sampled-token in the native one).
% Figure~\ref{fig:flops-framework} shows both \texttt{verl} generations: GRPO
% stays above every \opd{} run at every evaluated step under either
% implementation.

\begin{figure}[!t]
\centering
\includegraphics[width=\linewidth]{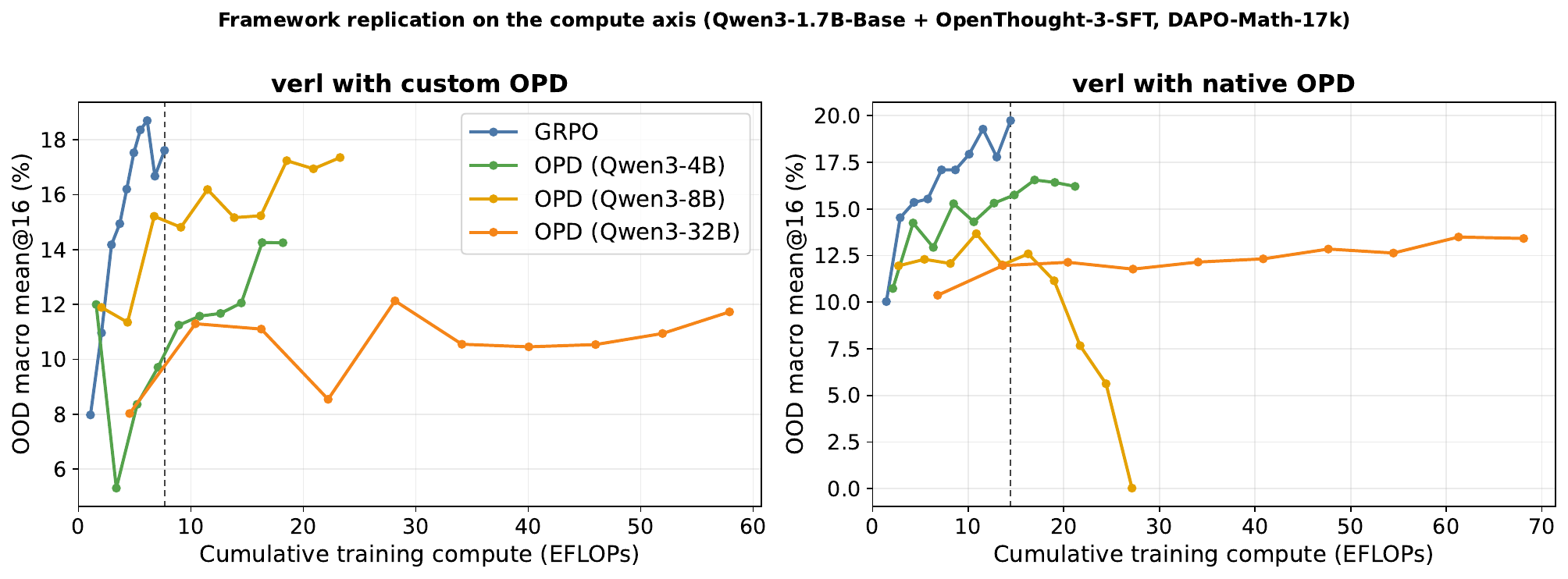}
\caption{\textbf{Replicate GRPO $>$ \opd{} ordering on different frameworks.} The compute ordering survives the change of framework: GRPO is above every
\opd{} run per FLOP in both the old/custom and latest/native stacks. On latest/native stack, we follow their default choice to use sampled-token reverse KL estimator. Other than the difference we specified, the hyperparamters and configurations match between left and right plots.}
\label{fig:flops-framework}
\end{figure}

\paragraph{Replication across different verl versions.} Framework choice bundles the reverse-KL estimator, the rollout engine,
and a long tail of implementation details, any of which could in principle
flip the GRPO-versus-\opd{} ordering. We replicate the comparison on
three training stacks to isolate this. The first two stacks are two
version of our \texttt{verl}-based pipeline, run on a
external-teacher setup of \S\ref{sec:main}: \texttt{Qwen3-1.7B-Base} student,
external \texttt{Qwen3-}\{\texttt{4B}, \texttt{8B}, \texttt{32B}\} thinking teachers, DAPO-Math-17K, and the
shared hyperparameters of Table~\ref{tab:shared-hypers} (batch \num{64},
learning rate \num{2e-6}, \num{200} optimizer steps, group size $G{=}8$,
\num{7168}-token response maximum). One set of results use an older
\texttt{verl} generation (\texttt{0.7.0.dev}) with our own reverse-KL
\opd{} implementation and a closed-form top-\num{16} estimator; the
second stack is the current \texttt{verl} release
(\texttt{0.8.0.dev}), whose native reverse-KL \opd{} implementation uses
a sampled-token estimator. All other hyperparameters match between the
two stacks; the only axis of variation is the estimator and the
surrounding implementation. Figure~\ref{fig:flops-framework} shows both
results from both verl versions side by side: GRPO sits above every \opd{} run at every
evaluated FLOP count under either implementation, so the estimator
variant does not flip the ordering. 

\paragraph{tinker replication at 9B scale.}
The third replication uses the \textbf{\texttt{tinker}} training library from Thinking
Machines Lab~\citep{tml2025opd}, at a different scale and student family. From
a shared rank-\num{128} SFT checkpoint of \texttt{Qwen3.5-9B-Base} (as student/initialization) and a \texttt{Qwen3.5-9B}
teacher (\opd{} run) we ran GRPO and \opd{} for \num{40} optimizer steps at
matched settings --- learning rate \num{1e-4}, \num{64} prompts per batch,
\num{4} rollouts per prompt, \num{16384}-token response maximum, DeepMath-103k training set; the choice of hyperparamters are detailed in TML's tutorial page \footnote{\url{https://tinker-docs.thinkingmachines.ai/cookbook/recipes/distillation/}} or default values in the \texttt{tinker} codebase.
At step \num{40}, GRPO reaches AIME'24 \texttt{mean@8} $=\num{0.542}$
while \opd{} collapses to \texttt{mean@8} $=\num{0.013}$; \pct{98.75} of \opd{}'s evaluation rollouts saturate the
\num{16384}-token maximum against \pct{26.3} for GRPO (Figure~\ref{fig:tinker-9b}). This is the same length-inflation pathology documented on \texttt{verl} in Appendix~\ref{sec:why}, now reproduced on a different framework, student family (Qwen3.5), base scale (9B), and cold start (rank-128 LoRA).

\begin{figure}[!t]
\centering
\includegraphics[width=0.8\linewidth]{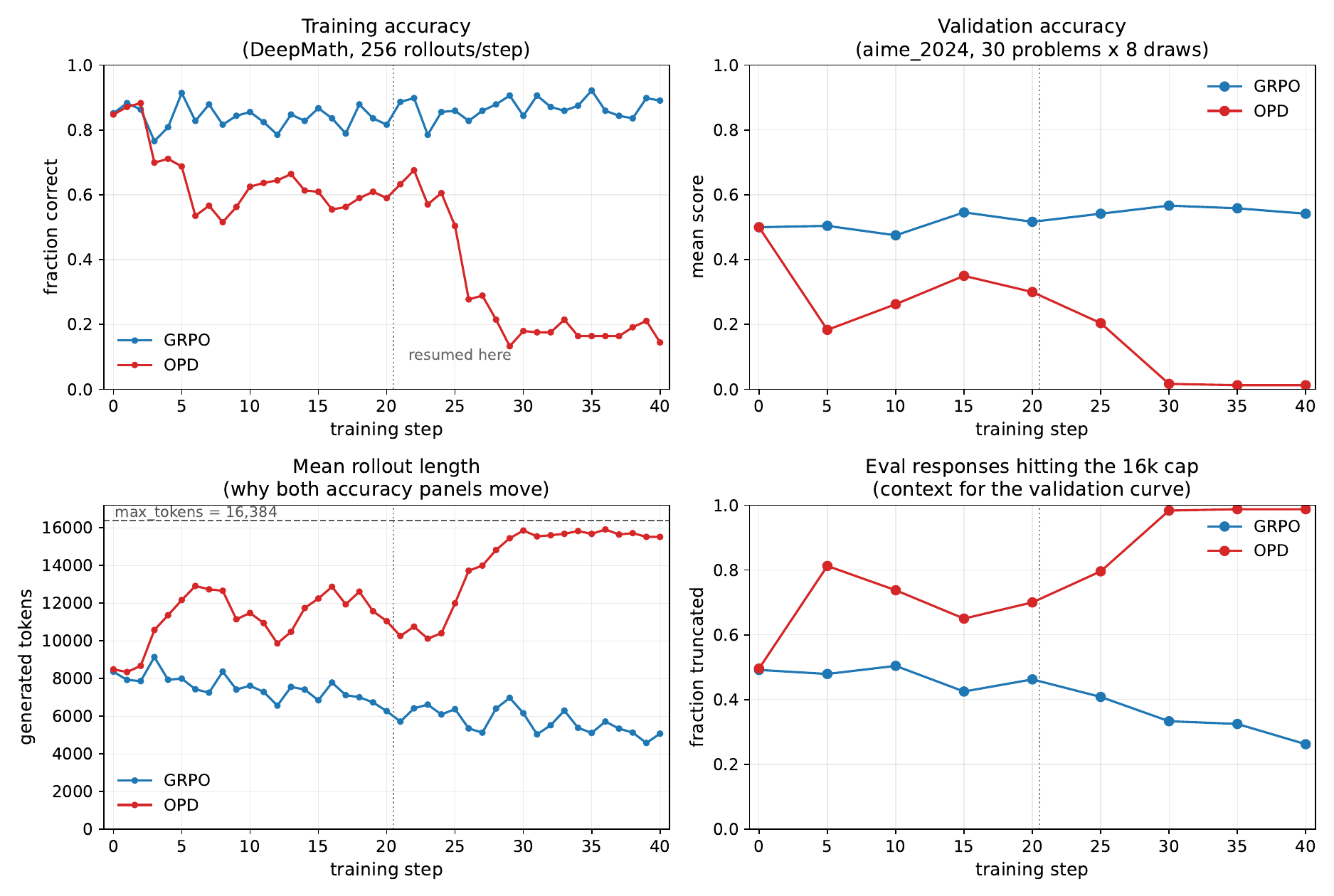}
\caption{\textbf{The same failure appears in a different framework (tinker, \texttt{Qwen3.5-9B-Base}).} GRPO and \opd{} on DeepMath from a shared rank-\num{128} SFT checkpoint, with \texttt{Qwen3.5-9B} as the \opd{} teacher, for \num{40} optimizer steps. \textbf{Top left:} training accuracy (\num{256} rollouts per step). \textbf{Top right:} AIME24 validation accuracy (\num{30} problems, \num{8} samples each). \textbf{Bottom left:} mean rollout length; the dashed line is the \num{16384}-token limit. \textbf{Bottom right:} fraction of evaluation responses truncated at the limit. The dotted vertical line marks where training resumed from the step-\num{20} checkpoint. GRPO holds AIME24 accuracy near \num{0.54}. \opd{} falls to \num{0.013} as its rollouts grow to the limit, and \pct{98.75} of its evaluation responses are truncated. The ordering and the length saturation match the \texttt{verl} results in Figures~\ref{fig:flops-framework} and~\ref{fig:length}.}

\label{fig:tinker-9b}
\end{figure}

Using a controlled suite of GRPO and \opd{} within each framework, we
show that GRPO is more compute-efficient than \opd{} with an
external teacher across all three stacks.

\section{Length as a lens to compare model behaviors}
\label{app:why-detail}
\label{sec:why}

\subsection{Length inflation in \opd{} except self-GRPO teacher}

\begin{figure}[!t]
\centering
\includegraphics[width=0.9\linewidth]{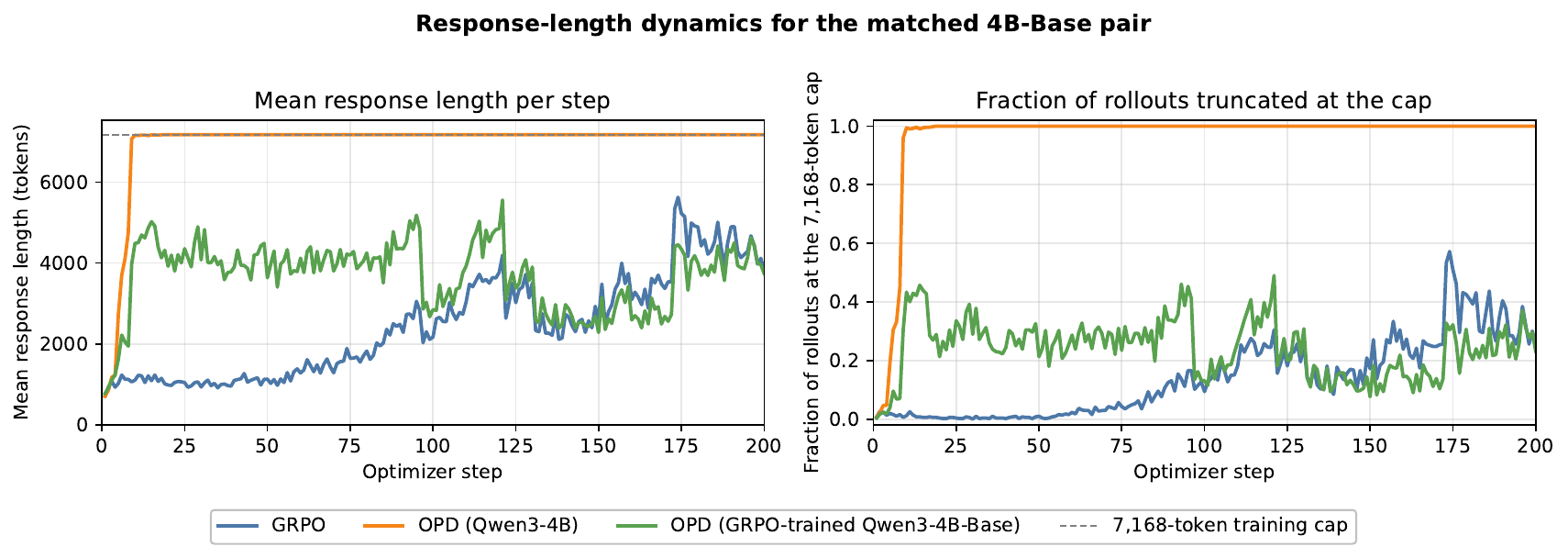}
\caption{\textbf{With an external thinking teacher, \opd{} rollouts fill the length limit within ten steps.} \texttt{Qwen3-4B-Base} student on DAPO-Math-17k, trained with GRPO, with \opd{} from the external \texttt{Qwen3-4B} thinking model (the same size as the student), and with \opd{} from the self-GRPO teacher. \textbf{Left:} mean response length per optimizer step; the dashed line is the \num{7168}-token training limit. \textbf{Right:} fraction of rollouts truncated at the limit. \opd{} with the external teacher reaches the limit by step \num{9} and truncates every rollout for \num{182} of \num{200} steps. GRPO and \opd{} with the self-GRPO teacher end at similar lengths (\num{3813} and \num{3741} tokens) and never saturate the limit.}

\label{fig:length}
\end{figure}

With an external thinking teacher, \opd{} response lengths inflate rapidly until
every rollout saturates the response maximum. Here the teacher is an external thinking
 model \texttt{Qwen3-4B} ---the \emph{same size} as the \texttt{Qwen3-4B-Base} student---so
the collapse cannot be attributed to the student lacking the capacity to represent
a much larger teacher. \opd{} nonetheless drives the mean response length past
\num{7000} tokens by step \num{9} (with \pct{96} of rollouts already truncated)
and then holds every sequence in the batch at the \num{7168}-token maximum (clip ratio
$\ge \num{0.999}$) from step \num{19} onward (Figure~\ref{fig:length}). Trained with GRPO, the same student stays near \num{1000}--\num{1500} tokens for the first ${\approx}\num{55}$ steps and ends at \num{3813} tokens (\pct{53} of the limit).

However, this behavior is not inherent to dense per-token supervision. With the self-GRPO teacher (\S\ref{sec:crossover}), \opd{} does not inflate length: the student ends at \num{3741} tokens, close to GRPO's \num{3813}. 
This inflation encodes pathological behavior of the model (Appendix~\ref{app:length-inflation:pathology}). We believe, following the dense-signal analogy of \opd{}, the problem is not the
dense signal itself but the wrong dense signal: an external teacher
whose token distribution is far from the student's. However, it is
difficult to characterize a priori what makes a given model a good
teacher. Given an arbitrary LLM, there is currently no way to predict
whether it will function as a stable \opd{} teacher without running
the distillation. Developing such a diagnostic remains an open problem.

\begin{figure}[h]
\centering
\includegraphics[width=\linewidth]{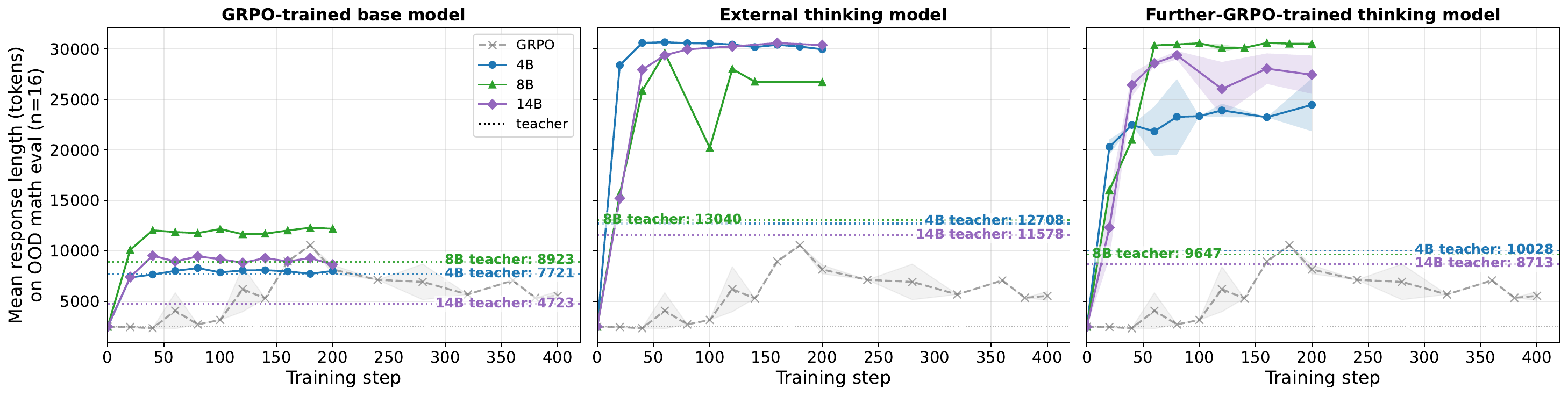}
\caption{\textbf{External thinking teachers make \opd{} students two to three times as long as the teacher.} \texttt{Qwen3-4B-Base} student; mean response length on the OOD math evaluation (\num{16} samples per problem) over training steps. Teachers at \num{4}B, \num{8}B, and \num{14}B are of three types: GRPO-trained base models (left), external thinking models (centre), and external thinking models further trained with GRPO (right). Dotted horizontal lines mark each teacher's own response length on the same prompts, and the grey dashed curve is GRPO. With both kinds of thinking teacher, the student's responses grow to \num{2}--\num{3}$\times$ the teacher's length, close to the \num{30720}-token evaluation limit. With GRPO-trained base teachers the gap is much smaller: the student matches the \num{4}B teacher's length and is \num{1.4}$\times$ and \num{1.9}$\times$ as long as the \num{8}B and \num{14}B teachers.}

\label{fig:resp-len-by-teacher}
\end{figure}

\paragraph{Additional results with other teacher types}
Figure~\ref{fig:resp-len-by-teacher} plots mean response length
(tokens) on the OOD math evaluation across training steps for three
teacher families (\S\ref{sec:crossover}) --- GRPO-trained base models, external thinking models, and further-GRPO-trained thinking models. With an external thinking teacher (centre
panel), the \opd{}-trained student's responses balloon to
${\approx}\num{28}$K--$\num{30}$K tokens by step \num{40} and stay
there, while the teachers themselves respond with only
${\approx}\num{12}$K--$\num{13}$K tokens on the same prompts
(4B teacher: \num{12708}; 8B teacher: \num{13040}; 14B teacher:
\num{11578}). The student is therefore emitting
${\approx}\num{2.3}$--$\num{2.6}\times$ longer responses than the
teacher whose token distribution it is trying to match, an outright
divergence from teacher behavior rather than mere length matching. The
same pattern holds with a further-GRPO-trained thinking teacher (right panel:
student ${\approx}\num{22}$K--$\num{30}$K vs teacher
${\approx}\num{9}$K--$\num{10}$K). With GRPO-trained base teachers (left panel), the gap is much smaller. The student is about $2\times$ as long as the \num{14}B teacher (9K vs.\ 4.7K tokens) and $1.4\times$ as long as the \num{8}B teacher (12K vs.\ 8.9K). Only with the student's own GRPO checkpoint as teacher do the two lengths match.

\subsection{Pathological modes co-occur with length inflation}
\label{app:length-inflation:pathology}

External-teacher \opd{} does not merely inflate response length. It also
produces short-period repetition, drops the required output format, and,
on the multi-domain mix, switches language. On DAPO-math-17K, we measure this on 3 runs in Figure~\ref{fig:length}: GRPO on the target reward, \opd{} distilled from a self-GRPO teacher
(the same 4B model, trained via GRPO on this data), and \opd{} distilled
from an external thinking teacher \texttt{Qwen3-4B}. Those runs share the training and evaluation configurations (e.g., learning rate, batch size); only the training objective and teacher differ. For each run we take
first $500$ rollouts from the final (i.e., step-$200$) evaluation on AIME24/25 + AMC23
and measure response length with the Qwen3 tokenizer.

\begin{table}[h]
\centering
\small
\begin{tabular}{lccc}
\toprule
metric & GRPO & \opd{}, self-GRPO teacher & \opd{}, external teacher\\
\midrule
repetition loop (naive detection) & $16.6\%$ & $15.2\%$ & \textcolor{red}{\textbf{$63.6\%$}} \\
median response length (tokens) & $6{,}504$ & $6{,}542$ & \textcolor{red}{\textbf{$30{,}720$~(=~maximum)}} \\
\quad \emph{$\hookrightarrow$ if detected loops removed} & $4{,}573$ & $4{,}937$ & \textcolor{red}{\textbf{$21{,}031$}} \\
mean response length (tokens) & $11{,}233$ & $11{,}605$ & \textcolor{red}{\textbf{$30{,}229$}} \\
\quad \emph{$\hookrightarrow$ if detected loops removed} & $7{,}763$ & $8{,}353$ & \textcolor{red}{\textbf{$19{,}627$}} \\
$\geq 95\%$ of eval maximum & $19.0\%$ & $21.0\%$ & \textcolor{red}{\textbf{$97.8\%$}} \\
CJK language collapse & $0.0\%$ & $0.0\%$ & $0.2\%$ \\
missing $\backslash$boxed$\{\cdot\}$ & $21.2\%$ & $23.2\%$ & \textcolor{red}{\textbf{$49.0\%$}} \\
\bottomrule
\end{tabular}
\caption{Rollout pathologies at step $200$ on math-only training ($500$
rollouts per run, Qwen3 tokenizer). Values in \textcolor{red}{\textbf{bold
red}} are $\geq 2\times$ the GRPO baseline on the same metric.
``Repetition loop (naive detection)'' flags a rollout when the $100$-char
window \texttt{response[-200:-100]} appears $\geq 3$ times elsewhere in
the response --- a conservative probe that catches short-period token-level
loops but misses longer-period, paraphrased, and semantic repetition.
``$\hookrightarrow$ if detected loops removed'' cuts each detected-loop
response at the first occurrence of its probe pattern and re-measures
length; because the probe is conservative, this is an \emph{upper bound}
on unique-content length, and still contains an unknown amount of
undetected higher-order repetition. Even against this upper bound the
external-teacher response is $\sim 4\times$ the GRPO de-looped
median.}
\label{tab:app:math-only-teacher-choice}
\end{table}

GRPO and self-GRPO-teacher \opd{} are essentially indistinguishable: within
$\sim 40$ tokens of median response length and $\sim 2$ percentage points
on every metric. The self-GRPO teacher transfers its stopping behavior
almost perfectly to the student. In contrast, the external-4B-teacher \opd{} is
qualitatively different: median response saturates at the eval maximum,
$97.8\%$ of rollouts hit the maximum, and $63.6\%$ contain a detected
short-period repetition loop --- the student has lost its ability to stop.
Removing detected loops brings the median only from $30{,}720$ to
$21{,}031$ tokens, still $\sim 4\times$ GRPO's de-looped median of
$4{,}573$.

The same picture holds on the 2-domain math + code mix.
Joint GRPO and multi-teacher self-GRPO \opd{} ($G{=}1$, one GRPO specialist
teacher per domain) both terminate cleanly: maximum-hit stays under $21\%$
on math and under $4\%$ on code for both, and multi-teacher self-GRPO
\opd{} is in fact \emph{tighter} than GRPO on code ($0.2\%$ maximum-hit and
$0.2\%$ detected repetition, versus $3.6\%$ / $3.6\%$ for GRPO). The
external teacher (\texttt{Qwen3-8B}) collapses to the opposite regime on
\emph{both} domains: $100\%$ maximum-hit, $\sim85\%$ detected repetition
loops, and median response saturated at the eval maximum ($30.7$K math,
$31.7$K code). On top of this shared length-inflation failure, the
\texttt{Qwen3-8B} external teacher demonstrates a language-collapse mode absent
elsewhere --- $100\%$ of its math and code rollouts contain Chinese
characters in the first $2$K characters, versus $0\%$ under GRPO and
multi-teacher self-GRPO.

\subsection{Length behavior in multi-domain settings}
\label{app:multi-domain}

\begin{table}[h]
\centering
\small
\caption{Single-task GRPO reference length per task (\texttt{Qwen3-4B-Base}): mean
evaluation response length, in tokens, at the checkpoint used as the
multi-teacher \opd{} teacher.}
\label{tab:baseline_len}
\begin{tabular}{lr}
\toprule
Task & Length (tokens) \\
\midrule
Math     & $7{,}721$ \\
Code     & $1{,}003$ \\
IFBench  & $829$ \\
MMLU-Pro & $804$ \\
\bottomrule
\end{tabular}
\end{table}

\begin{figure}[t]
\centering
\includegraphics[width=\textwidth]{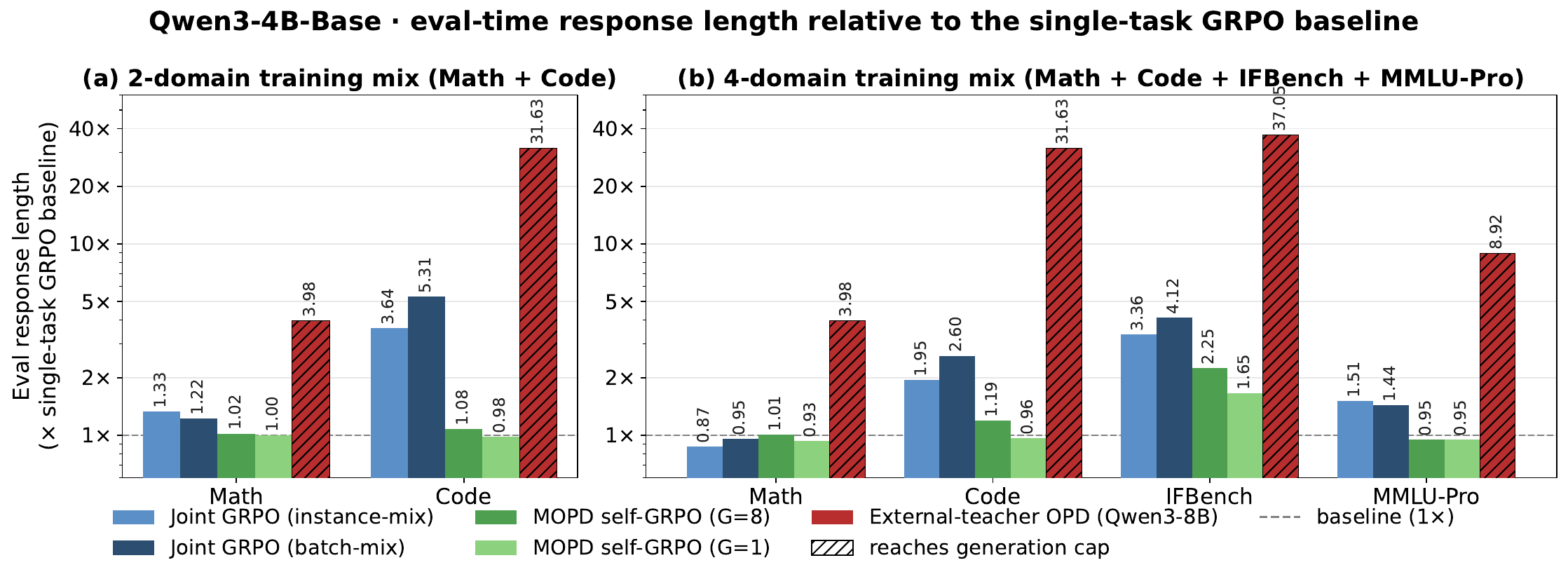}
\caption{\textbf{Multi-teacher \opd{} with self-GRPO teachers keeps the specialists' response length, while an external teacher drives responses to the generation limit.} Evaluation response length of \texttt{Qwen3-4B-Base} students trained on (a) the \num{2}-domain mix (math, code) and (b) the \num{4}-domain mix (math, code, IFBench, MMLU-Pro). Lengths are shown as a ratio to the single-task GRPO specialist for each domain (Table~\ref{tab:baseline_len}). The dashed line marks a ratio of \num{1}, and the $y$-axis is logarithmic. Hatched bars reach the generation limit. Multi-teacher \opd{} stays within \num{0.93}--\num{1.19}$\times$ of the specialists on every domain except IFBench (\num{1.65}--\num{2.25}$\times$). Joint GRPO produces \num{0.87}--\num{5.31}$\times$ the specialist length, and \opd{} with the external \texttt{Qwen3-8B} teacher produces \num{3.98}--\num{37.05}$\times$.}

\label{fig:len_by_method}
\end{figure}

On each tasks, we measure mean evaluation response length (tokens, $n{=}16$ per prompt) for
\texttt{Qwen3-4B-Base}. We use single-task GRPO as the per-task reference; we use the GRPO checkpoint that each multi-teacher \opd{} distills from (math, IFBench, and MMLU-Pro at
200 steps, code at 80; Table~\ref{tab:baseline_len}).
Figure~\ref{fig:len_by_method} reports each regime's length as a ratio to this
reference. Multi-teacher \opd{}, which distills from these single-task GRPO
teachers, recovers the teacher length closely: within roughly $20\%$ on math,
code, and MMLU-Pro ($0.93$--$1.19\times$), and somewhat longer on IFBench
($1.65$--$2.25\times$). \opd{} from an external teacher (\texttt{Qwen3-8B})
instead inflates length on every task like what happens in Appendix~\ref{app:length-inflation:pathology}. Joint GRPO lies between the two, and its length is
setting-dependent.

Under joint GRPO, each task's response length is pulled toward the length of the \emph{other} tasks sharing its batch
(Table~\ref{tab:baseline_len}). In the two-task math + code mix, math's
long single-task baseline drags code well above its own, to $3.64\times$
the code reference, while math itself is only mildly inflated
($1.33\times$). Adding MMLU-Pro and IFBench, the two shortest single-task
baselines, pulls both shared tasks back down: math from $1.33\times$ to
$0.87\times$ the reference and code from $3.64\times$ to $1.95\times$. Response length under joint GRPO is thus governed by the mix of
tasks in the training batch.

\section{Out-of-distribution generalization of \opd{} with an external teacher}
\label{app:extrapolation}

Although external thinking teachers are compute-inefficient for
\opd{} on the training domain (\S\ref{sec:main}), our experiments
surface an interesting side benefit: \opd{} with these teachers
produces surprisingly strong out-of-distribution generalization that
GRPO does not match. We collect two observations below suggesting that
\opd{}'s per-token objective carries information beyond the training
distribution.

\paragraph{Observation 1: \opd{} wins GRPO in OOD at the difficulty cells where
the teacher itself is weak.}
On the difficulty-stratified DAPO-Math-17K grid (\S\ref{sec:stratification}),
GRPO wins in-domain in every one of the eight
cells (Figure~\ref{fig:difficulty-grid}).
On the OOD macro (AIME24 + AIME25 + AMC23), however, \opd{}-4B and
\opd{}-8B finish \emph{above} GRPO in three of the eight cells,
every cell in the teacher-solves-nothing column plus the
(student~$=\!0$, teacher~$\in(0,\num{0.3}]$) cell
(Figure~\ref{fig:difficulty-grid-ood}). In these cells the teacher signal is weak, so it is surprising
that \opd{} outperforms GRPO on OOD. In fact, to our surprise, \opd{}'s OOD macro sits at a
nearly constant \pct{15}--\pct{18} across all eight regimes while
GRPO's varies with the data it was trained on;
\opd{}'s OOD score looks more like a property of the objective than of what it was trained on.

\paragraph{Observation 2: code-trained \opd{} with external teachers improves math OOD
by ${\approx}\num{7}$--\num{13} points over code-trained GRPO.}
On two code training sets (LCBv1--v5 and CodeForces), both external
teachers reach math OOD macro well above the code-trained GRPO despite
never seeing math training data (Figure~\ref{fig:code-math-extrap} right
column). 

\opd{}(\texttt{Qwen3-32B}) peaks at \pct{33.3} on LCBv1--v5 and \pct{33.9} on CodeForces; \opd{}(\texttt{Qwen3-8B}) peaks at \pct{29.0} and \pct{32.1} respectively;
code-trained GRPO peaks at \pct{20.7} and \pct{24.8} (math OOD, for the two code training sets), compared to the math specialist's own math-OOD accuracy of \pct{33.9} (Figure~\ref{fig:teacher-by-nature} dashed ``GRPO teacher" line, \S\ref{sec:teacher-nature}).
% The two external
% teachers are stacked, with \opd{}(Qwen3-32B) leading \opd{}(Qwen3-8B) by \num{4.3} points
% on LiveCodeBenchv1--v5 and \num{1.8} points on CodeForces, so the cross-domain
% transfer is not specific to a single teacher size. Caveat: the GRPO code runs stop at
% step \num{80} (${\approx}\SI{1.4}{\exa\flop}$), below the \opd{}
% compute budgets, so from these plots alone we cannot rule out that GRPO
% would also improve on math OOD given equal compute; on
% CodeForces, GRPO's math-OOD curve is still climbing at the last logged
% step, so the compute-match caveat bites harder there than on LCBv1--v5. 

\section{Additional results of running GRPO on teacher}
\label{app:samebase-replication}

Section~\ref{sec:teacher-nature} shows that \opd{} with a self-GRPO teacher,
the student's own GRPO checkpoint, recovers the teacher's accuracy with far
fewer FLOPs than GRPO. This appendix tests whether that result holds
beyond the main setting of a \texttt{Qwen3-4B-Base} student on math. We vary three
factors one at a time: student size (\texttt{Qwen3-8B-Base}), domain (code), and model
family with a post-trained starting point (\texttt{Olmo-3-7B-Instruct-SFT}). In each
case the self-GRPO teacher reaches the teacher's accuracy faster than GRPO.

\begin{figure}[t]
\centering
\includegraphics[width=\linewidth]{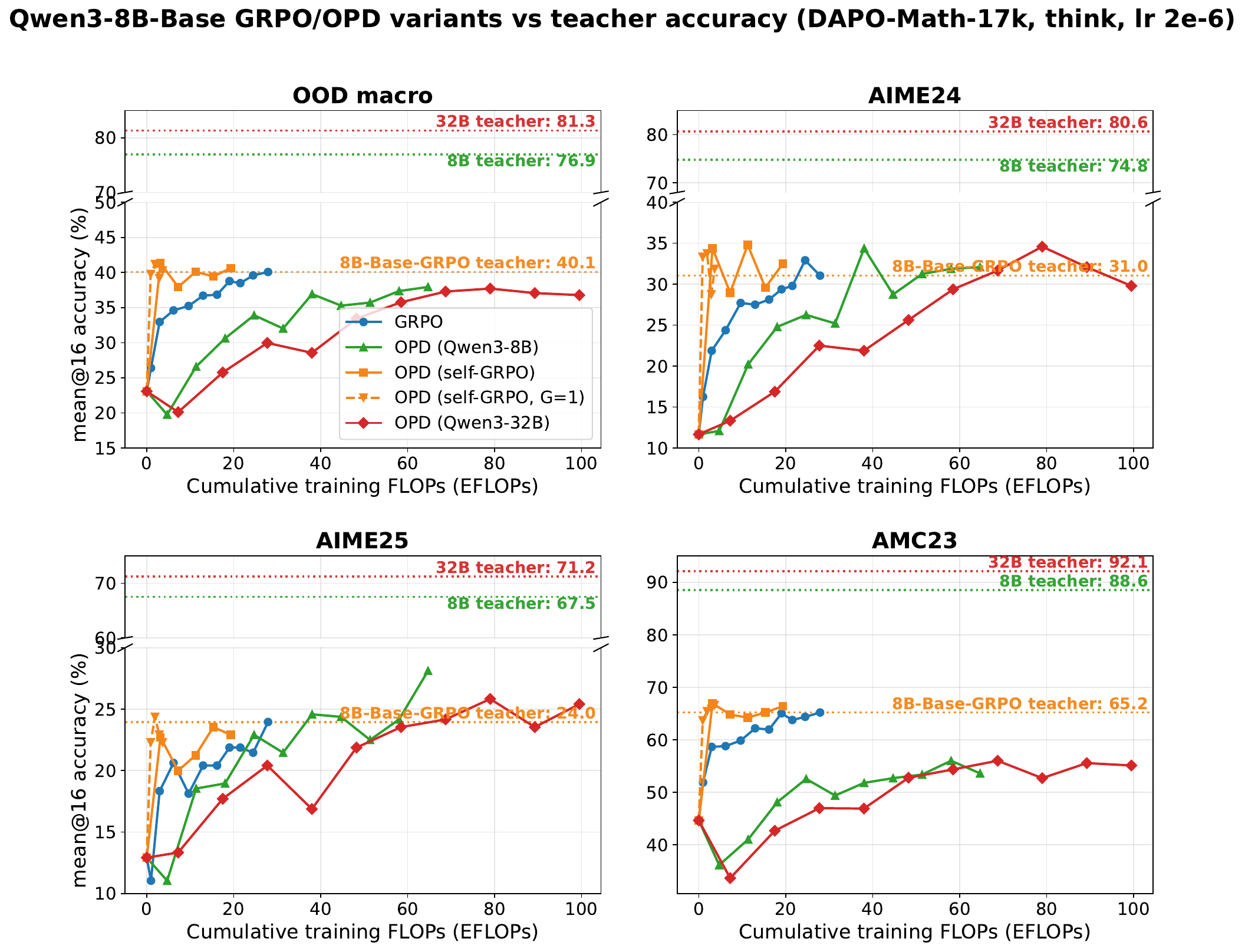}
\caption{\textbf{The self-GRPO teacher result holds for an 8B student (companion to Figure~\ref{fig:teacher-by-nature}).} \texttt{Qwen3-8B-Base} student on DAPO-Math-17k; \texttt{mean@16} against training FLOPs. Dotted lines mark each teacher's accuracy, and the $y$-axis is broken to show the external teachers. \opd{} with the self-GRPO teacher reaches the teacher's \pct{40.1} OOD macro within about \SI{3}{\exa\flop} (group size $G{=}8$) or \SI{1}{\exa\flop} ($G{=}1$), against \SI{27.9}{\exa\flop} for GRPO. With the external \texttt{Qwen3-8B} and \texttt{Qwen3-32B} teachers, \opd{} ends at \pct{37}--\pct{38} after \num{65}--\SI{100}{\exa\flop}, far below those teachers' own \pct{76.9} and \pct{81.3}.}

\label{fig:tce-self}
\end{figure}

\paragraph{Replication with \texttt{Qwen3-8B-Base}.} The fast-saturation pattern of the self-GRPO teacher
(\S\ref{sec:teacher-nature}) is not specific to the 4B student.
Figure~\ref{fig:tce-self} repeats the comparison, on DAPO-Math-17K, with a \texttt{Qwen3-8B-Base}
student against three teachers: external \texttt{Qwen3-8B}, external
\texttt{Qwen3-32B}, and the student's own GRPO-trained \texttt{Qwen3-8B-Base} checkpoint. The
With the self-GRPO teacher, the student reaches the teacher's accuracy within about \SI{3}{\exa\flop} at $G{=}8$ and \SI{1}{\exa\flop} at $G{=}1$, faster than with either external teacher. With the external \texttt{Qwen3-8B} and \texttt{Qwen3-32B} teachers, the student climbs more slowly and stays below the teacher's accuracy throughout the compute budget shown.

\paragraph{Replication on code domain.} On code, the same pattern holds: the GRPO-trained 4B-base teacher reaches
GRPO's LCBv6 accuracy at $2.4\times$ fewer FLOPs on LiveCodeBench-v1--v5
and ties GRPO's quality on strict CodeForces, while the external
$32$B teacher stays far below GRPO
(Figure~\ref{fig:code-math-extrap} left column). GRPO-trained student itself is
therefore the only strong-to-weak \opd{} setup that consistently works
across both math and code domains.

\begin{figure}[h]
\centering
\includegraphics[width=\linewidth]{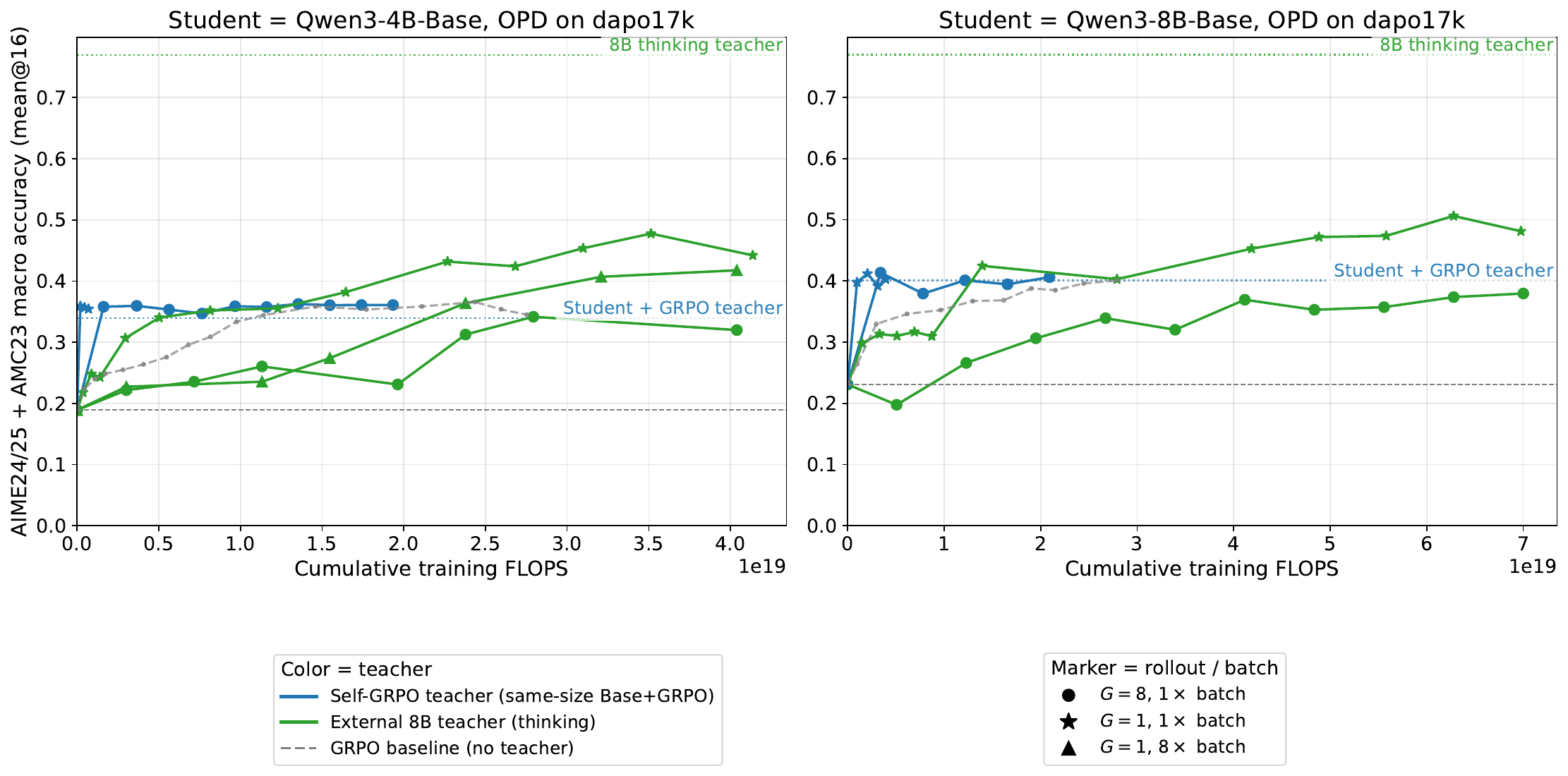}
\caption{\textbf{On math, $G{=}1$ makes \opd{} with an external teacher more compute-efficient, but the self-GRPO teacher is still far cheaper.} OOD macro \texttt{mean@16} on DAPO-Math-17k for \texttt{Qwen3-4B-Base} (left) and \texttt{Qwen3-8B-Base} (right) students. Colour marks the teacher (blue: self-GRPO; green: external \texttt{Qwen3-8B} thinking); marker shape marks the rollout setting. With the external teacher, $G{=}1$ beats $G{=}8$ at every compute budget and reaches \pct{47.7} (4B) and \pct{50.6} (8B), above GRPO. The self-GRPO teacher at $G{=}1$ reaches its teacher's accuracy within about \SI{1}{\exa\flop}, an order of magnitude less compute.}

\label{fig:tce-self-4b-8b}
\end{figure}

\begin{figure}[h]
\centering
\includegraphics[width=0.95\linewidth]{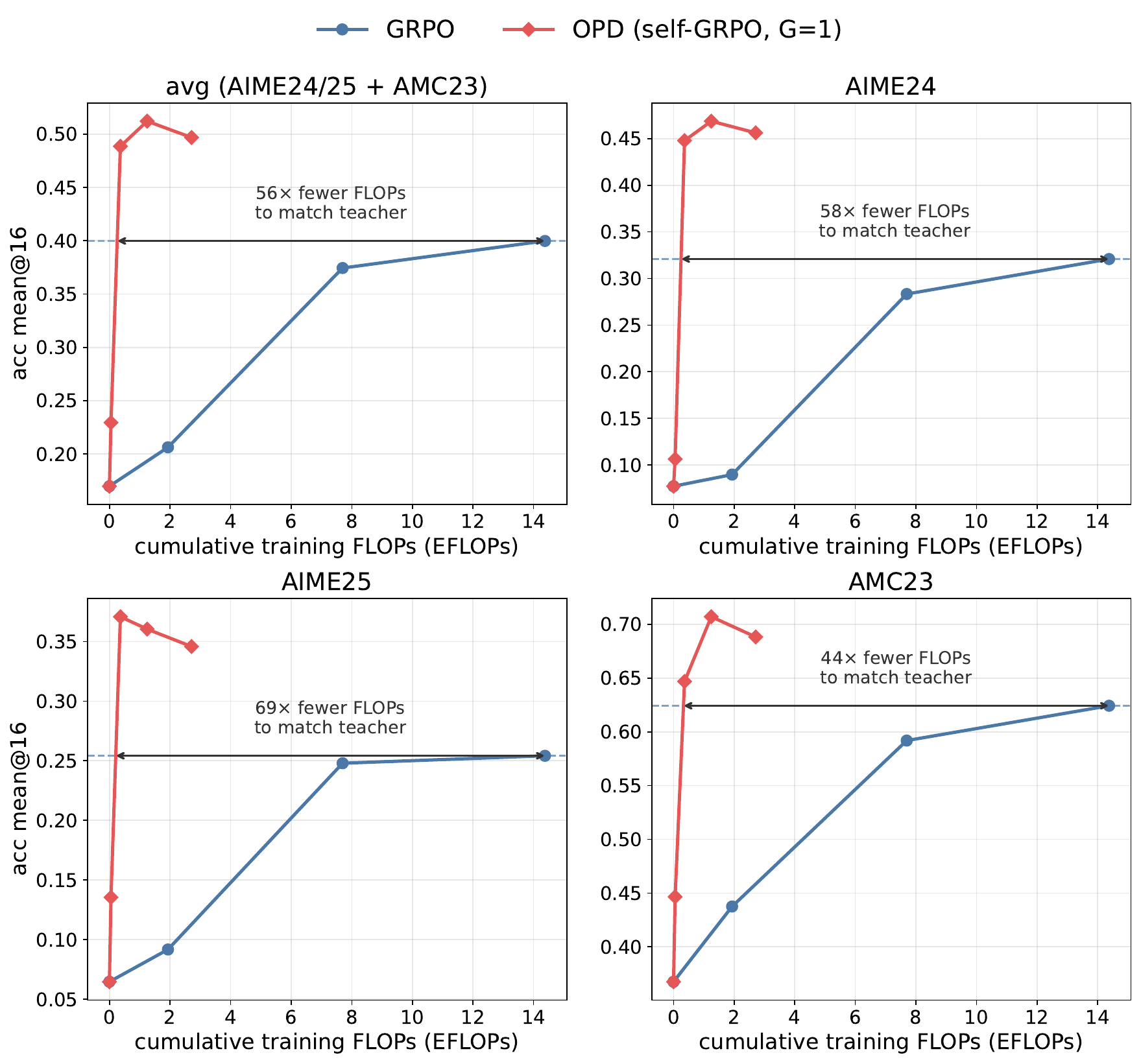}
\caption{\textbf{The self-GRPO teacher also works from an instruction-tuned SFT model (\texttt{Olmo-3-7B-Instruct-SFT}).} OOD macro \texttt{mean@16} and per-benchmark accuracy against training FLOPs on DAPO-Math-17k. GRPO trains for \num{200} steps at $G{=}8$; \opd{} distills the step-\num{200} GRPO checkpoint into the starting model at $G{=}1$. \opd{} matches the teacher's \pct{40.0} OOD macro with \num{56}$\times$ fewer FLOPs (\num{44}--\num{69}$\times$ per benchmark) and peaks at \pct{51.2}.}

\label{fig:olmo3-selfgrpo}
\end{figure}

\paragraph{Non-base cold-start replication (\texttt{Olmo-3-7B-Instruct-SFT}).}
Figure~\ref{fig:olmo3-selfgrpo} runs the same self-GRPO recipe on a
post-trained checkpoint from a different model family. Both curves share the
step-\num{0} \texttt{Olmo-3-7B-Instruct-SFT} student, the DAPO-Math-17K training set,
and an identical evaluation configuration. \emph{GRPO (self)} runs direct
GRPO on the SFT student ($G{=}8$, \num{200} steps) and reaches
\pct{40} OOD macro at \SI{14.4}{\exa\flop}. \emph{\opd{} $\leftarrow$
self-GRPO teacher} distills the step-\num{200} GRPO checkpoint back into the
same step-\num{0} SFT student ($G{=}1$, \num{80} steps) and reaches the
teacher's end-of-training accuracy within ${\sim}\SI{0.36}{\exa\flop}$
(step \num{20}), a ${\sim}\num{56}\times$ compute reduction that lines up with
the same-base results in Section~\ref{sec:teacher-nature}. The \opd{} student then continues past the
teacher and peaks at \pct{51} at \SI{1.25}{\exa\flop}, an
${\approx}\num{11}$-point overshoot over the self-GRPO teacher's own plateau. In the same-base setting, we also observe similar pattern (Figure~\ref{fig:code-math-extrap} and Figure~\ref{fig:teacher-by-nature}) but we are surprised to observe such a big out-performing.

\section{Measure model weight similarity by matrix spectra}
\label{app:iso-mechanism}

Prior work \citep{zhu2026iso} has investigated the change of model weight during post-training through the lens of spectral inheritance: RLVR can reuse the base model’s weight spectra while acquiring new behavior through changes in the associated input and output singular frames. In this section, we will conduct similar analysis through the measure proposed by \citet{zhu2026iso} in attempt to explain why \opd{} with external teacher doesn't lead the student to the teacher's performance.

\subsection{Setup}

For each pretrained matrix $W_B \in \mathbb{R}^{d_{\text{out}} \times d_{\text{in}}}$
of the base student, take its SVD $W_B = U_B\,\mathrm{diag}(\sigma_B)V_B^\top$
with $q = \min(d_{\text{out}}, d_{\text{in}})$, then decompose any candidate
displacement $\Delta W = W_1 - W_B$ in the base frames. Here $\sigma(W)$ denotes the $q$-vector of singular values of $W$ -- $\sigma(W) = (\sigma_1, \ldots, \sigma_q)$ in
non-increasing order.

Three spectrum-based scalars
\citep{zhu2026iso} are used to characterize changes in each matrix. First, $\kappa_{\mathrm{spec}}$:
\begin{align}
\kappa_{\mathrm{spec}}(\Delta W) &= \frac{d_{\text{out}} d_{\text{in}}}{q}
    \cdot \frac{\Vert \mathrm{diag}(U_B^\top \Delta W\,V_B) \Vert^2}
                {\Vert \Delta W \Vert^2}
\end{align}
Larger $\kappa_{\mathrm{spec}}$
indicates more singular-value shift. \citet{zhu2026iso} shows that RLVR updates leave a student's singular-value spectrum nearly unchanged
($\kappa_{\mathrm{spec}}\!\approx\!1$--$3$), while SFT updates
shift it by orders of magnitude
($\kappa_{\mathrm{spec}}\!\gtrsim\!10^{2}$--$10^{3}$). 

$\rho_{\Sigma}$ is the share of the update magnitude that shows up as a
change in singular values rather than a rotation of the base frames:
$\rho_{\Sigma}=0$ marks a pure Stiefel move (frames rotate, spectrum
unchanged), $\rho_{\Sigma}=1$ marks a purely spectral update (same
singular directions, only the values change). 
\begin{align}
\rho_{\Sigma}(\Delta W) &= \frac{\Vert \sigma(W_1) - \sigma(W_B) \Vert}
                                {\Vert \Delta W \Vert}
\end{align}
$\delta_{\Sigma}$ is the
absolute size of that spectral shift relative to the base weight scale:
$\delta_{\Sigma}\!\approx\!0$ means the spectrum barely moves compared
to $W_B$, while $\delta_{\Sigma}\!\approx\!1$ means the singular values
change by an amount comparable to the base norm itself. On the same
runs \citet{zhu2026iso} reports RLVR updates keep $\rho_{\Sigma}$ small
(a few percent) and $\delta_{\Sigma}$ vanishingly small
($\sim\!\num{10}^{-5}$--$\num{10}^{-4}$), while SFT-scale updates push
$\rho_{\Sigma}\!\gtrsim\!\num{0.15}$ and $\delta_{\Sigma}$ by orders
of magnitude.
\begin{align}
\delta_{\Sigma}(\Delta W) &= \frac{\Vert \sigma(W_1) - \sigma(W_B) \Vert}
                                    {\Vert W_B \Vert}
\end{align}

Following the practice in \citet{zhu2026iso}, we calculate each value for each matrix and aggregate by averaging.

\subsection{Spectral distance from the base checkpoint}
\label{app:iso-per-teacher}
Table~\ref{tab:iso-per-teacher} reports the spectrum-based scalars for two kinds of comparison at 4B and 8B scale. The first kind compares a candidate teacher of the same size against the student base. At both scales the external thinking teacher sits far from the base ($\kappa_{\mathrm{spec}} \approx \num{2000}$--$\num{3600}$, $\rho_{\Sigma} \approx \pct{15}$). The student's own GRPO-trained checkpoint is nearly isospectral ($\kappa_{\mathrm{spec}} \approx 1$--$3$, $\rho_{\Sigma} \approx \pct{6}$--$\pct{12}$), which is consistent with the RLVR-versus-SFT dichotomy of \citet{zhu2026iso}. Three orders of magnitude in $\kappa_{\mathrm{spec}}$ separate the two teacher types, a weight-space signature of the overlap phase transition of \S\ref{sec:crossover}. The second kind compares the \opd{} student at different training stage against the same base. At 4B, where the student is distilled from the external \texttt{Qwen3-4B} teacher, the update grows with training ($\Vert\Delta W\Vert$ rises from $2.69$ at mid checkpoint to $3.76$ at last checkpoint d). Its $\kappa_{\mathrm{spec}}$ stays at $3.2$--$3.4$, the same order as the GRPO checkpoints and three orders of magnitude below the teacher it is trained against. At 8B, where the student is distilled from an external \texttt{Qwen3-8B} teacher, the step-100 (mid checkpoint) student has $\kappa_{\mathrm{spec}} = 1.50$, again within the RLVR range.

\paragraph{Discussion.}
Like GRPO, reverse-KL \opd{} updates the student on its own rollouts, and the resulting spectral change stays within the same RLVR $\kappa_{\mathrm{spec}}$ range regardless of the teacher (Table~\ref{tab:iso-per-teacher}). If \opd{} shifts the student's spectrum by only a small amount per unit of training compute, a teacher whose spectrum sits three orders of magnitude away is out of reach at any realistic compute budget: matching it would require an SFT-scale spectral move, which \opd{}'s gradient direction does not produce. A self-GRPO teacher sits in the same RLVR range as anything \opd{} can produce, so a small number of RLVR-magnitude steps is enough to reach it. The three-orders-of-magnitude $\kappa_{\mathrm{spec}}$ gap in Table~\ref{tab:iso-per-teacher} is therefore a plausible mechanism for why external-teacher \opd{} is not compute-efficient on the training domain (\S\ref{sec:main}).

\begin{table}[h]
\centering
\caption{Spectrum-based scalars of \citet{zhu2026iso} between a base checkpoint $W_B$ and a second checkpoint $W_1$, with $\Delta W = W_1 - W_B$. All scalars are computed in the base's own SVD frames and aggregated over all 2-D weight matrices. Within each scale, the first two rows compare a candidate teacher against the base: the external thinking teacher and the student's own GRPO-trained checkpoint. The remaining rows compare the \opd{} student against the same base (4B: external \texttt{Qwen3-4B} teacher, steps 100 and 200; 8B: self-GRPO teacher, step 100). External thinking teachers sit ${\sim}3$ orders of magnitude farther from the base in $\kappa_{\mathrm{spec}}$ than GRPO checkpoints. \opd{} students stay in the GRPO range even when they are trained against an external teacher.}
\label{tab:iso-per-teacher}

\small
\begin{tabular}{@{}llrrrr@{}}
\toprule
student base weight $W_B$ & $W_1$ & $\kappa_{\mathrm{spec}}$ & $\rho_{\Sigma}$ & $\delta_{\Sigma}$ & $\Vert\Delta W\Vert$ \\
\midrule
\texttt{Qwen3-4B-Base} & \texttt{Qwen3-4B}     & \num{3571} & \num{0.155} & $\num{1.6e-2}$ & \num{162}  \\
\texttt{Qwen3-4B-Base} & GRPO-trained \texttt{Qwen3-4B-Base}     & \num{2.81} & \num{0.063} & $\num{5.3e-5}$ & \num{1.33} \\
\texttt{Qwen3-4B-Base} & mid checkpoint of OPD (\texttt{Qwen3-4B})     & \num{3.16} & \num{0.033} & $\num{5.7e-5}$ & \num{2.69} \\
\texttt{Qwen3-4B-Base} & last checkpoint of OPD (\texttt{Qwen3-4B})     & \num{3.42} & \num{0.025} & $\num{6.1e-5}$ & \num{3.76} \\
\midrule
\texttt{Qwen3-8B-Base} & \texttt{Qwen3-8B}     & \num{2077} & \num{0.155} & $\num{1.3e-2}$ & \num{202}  \\
\texttt{Qwen3-8B-Base} & GRPO-trained \texttt{Qwen3-8B-Base}    & \num{1.19} & \num{0.125} & $\num{7.7e-5}$ & \num{1.48} \\
\texttt{Qwen3-8B-Base} & mid checkpoint of OPD (\texttt{Qwen3-8B})     & \num{1.50} & \num{0.051} & $\num{8.14e-5}$ & \num{3.83}  \\
\texttt{Qwen3-8B-Base} & last checkpoint of OPD (\texttt{Qwen3-8B})     & \num{1.61} & \num{0.036} & $\num{8.06e-5}$ & \num{5.38}  \\
\bottomrule
\end{tabular}
\end{table}

\section{External \opd{} under single-rollout regimes}
\label{app:n1-defense}
% The main negative experiments on external teachers (\S\ref{sec:main}) uses the default
% rollout group size $G$ for each experiment. A natural objection is that
% reducing to $G{=}1$ (a single rollout per prompt) changes the compute
% profile: \opd{} pays for fewer rollouts per step, sees more unique prompts
% at matched total compute, and avoids the group-relative baseline that GRPO
% relies on. We test this on math and code, where many current models models are heavily trained on \cite{olmo2025olmo3}.
The main comparisons with external teachers (\S\ref{sec:main}) use a
rollout group size of $G{=}8$. The \opd{} loss is defined per token and
needs no group baseline, so a smaller group could make \opd{} cheaper:
with $G{=}1$, \opd{} spends fewer rollouts per prompt and sees more unique
prompts at matched training FLOPs. We test whether $G{=}1$ makes
external-teacher \opd{} competitive with GRPO on math, on code, and on the
\num{4}-domain mix of Section~\ref{sec:merging}.

\paragraph{Math: $G{=}1$ improves compute efficiency.}
On DAPO-Math-17K with a \texttt{Qwen3-4B-Base} student and an external \texttt{Qwen3-8B}
teacher, we keep the default batch size and train at $G{=}1$ for $8\times$
more optimizer steps, which matches the total number of rollouts in the
$G{=}8$ run. This variant improves the compute efficiency of \opd{} over the
$G{=}8$ run (Figure~\ref{fig:tce-self-4b-8b}). It takes longer in wall-clock
time because it runs more optimizer steps, and the response-length inflation
described in Appendix~\ref{app:length-inflation:pathology} still occurs. We additionally experiment with \texttt{Qwen3-8B-Base} student in Figure~\ref{fig:tce-self-4b-8b} (right panel).
% TODO: add whether G=1 reaches GRPO at matched FLOPs, e.g.
% "but remains below GRPO at matched FLOPs."

\begin{figure}[h]
\centering
\includegraphics[width=0.9\linewidth]{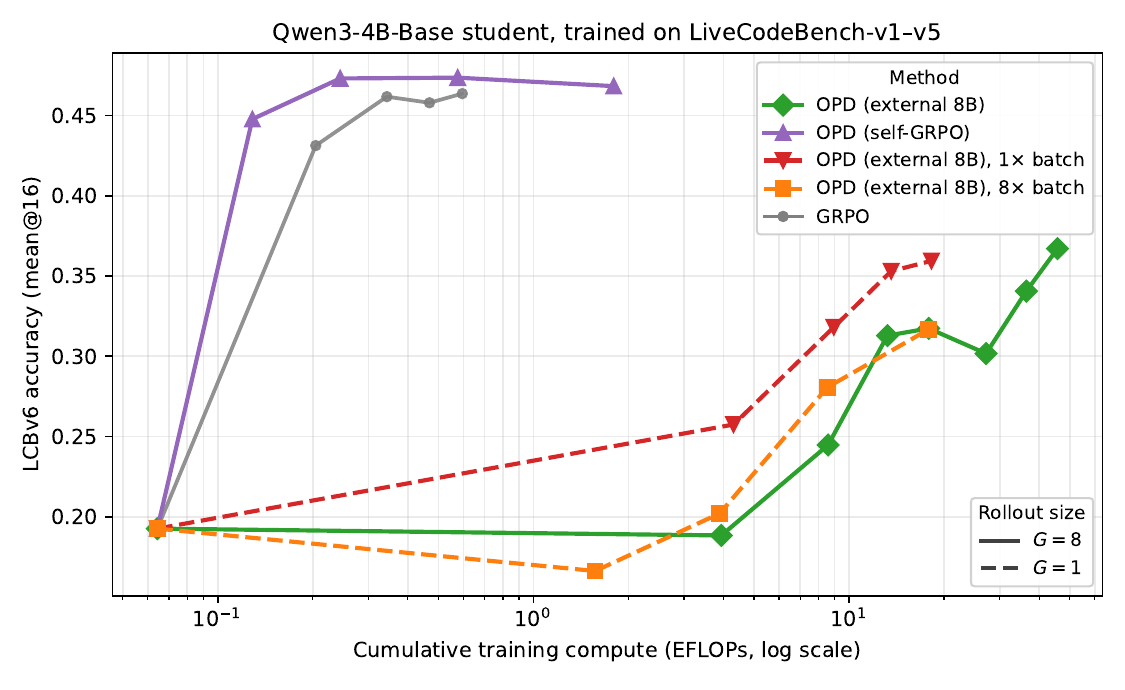}
\caption{\textbf{On code, $G{=}1$ does not rescue \opd{} with an external teacher.} \texttt{Qwen3-4B-Base} student trained on LiveCodeBench v1--v5; LCBv6 \texttt{mean@16} against training FLOPs (log scale). Solid lines use $G{=}8$ and dashed lines $G{=}1$; the external teacher is \texttt{Qwen3-8B} (thinking). The self-GRPO teacher reaches \pct{47.3} within \SI{0.25}{\exa\flop}, above GRPO (\pct{46.4}). With the external teacher, $G{=}1$ improves on $G{=}8$ at the default batch size but not with an $8\times$ batch, and both $G{=}1$ runs stop at \pct{32}--\pct{36}, more than \num{10} points below GRPO.}

\label{fig:n1-bsz-lcbv6}
\end{figure}

\paragraph{Code: $G{=}1$ does not help.}
On LiveCodeBench-v1--v5 with the same student and teacher, we test two
compute-matched $G{=}1$ variants: one at $8\times$ batch size, and one
at the default batch size trained for $8\times$ more optimizer steps.
Neither closes the gap: \opd{} remains substantially below GRPO on
LCBv6 \texttt{mean@16} (Figure~\ref{fig:n1-bsz-lcbv6}). The self-GRPO
teacher run on the same recipe reaches ${\approx}\pct{47}$ LCBv6
accuracy within ${\sim}\SI{0.5}{\exa\flop}$; the external
8B-thinking teacher at either $G{=}8$ or either $G{=}1$
matched-compute variant plateaus far below and pays an order of
magnitude more compute for less accuracy.

\paragraph{Multi-domain: the math advantage disappears.}
On math alone, $G{=}1$ lets external-teacher \opd{} outperform GRPO at
matched FLOPs. This advantage does not survive when math is trained together
with other domains. On the \num{2}-domain math + code mix, external-teacher
\opd{} at $G{=}1$ no longer outperforms GRPO on math
(Figure~\ref{fig:mathcode-per-domain}).
% TODO: add numbers, e.g. "reaching X mean@16 vs GRPO's Y at Z EFLOPs".
On the \num{4}-domain mix, training with the \texttt{Qwen3-8B} teacher becomes unstable: math accuracy falls from \pct{18.9} to \pct{7.2} by step \num{80}, briefly recovers to \pct{37.7} at step \num{160}, and ends at \pct{0.3} at step \num{200}; code falls from \pct{19.3} to \pct{2.5} by step \num{80} and ends at \pct{13.3} (Figure~\ref{fig:joint4-per-domain}).

\paragraph{Summary.}
Reducing the rollout group size to $G{=}1$ lets external-teacher \opd{}
outperform GRPO at matched FLOPs on single-domain math, at the cost of
$8\times$ as many optimizer steps. The same configuration loses on
code, loses its math advantage on the \num{2}-domain mix, and collapses on
the \num{4}-domain mix. Therefore, we believe $G{=}1$ does not conflict with our conclusion and observation in \S\ref{sec:ablations}.

\section{Multi-teacher \opd{}: experiment detail}
\label{app:multi-teacher-detail}

This appendix gives the details of the multi-teacher experiments in
Section~\ref{sec:merging}. The pipeline has two stages. First, we train one
GRPO specialist per domain from \texttt{Qwen3-4B-Base}. Second, we distill all
specialists into a fresh \texttt{Qwen3-4B-Base} student with multi-teacher
\opd{}. We compare this pipeline with two joint GRPO baselines that train a
single model on the same data mix, and with \opd{} from a single external
teacher. We run the comparison on a \num{4}-domain mix (math, code,
knowledge, and instruction following) and on its \num{2}-domain math + code
subset.

\paragraph{Training data.}
The \num{4}-domain mix combines DAPO-Math-17K (math), LiveCodeBench-v1--v5
(code), MMLU-Pro with \num{5}-shot prompts (knowledge), and IFBench-RLVR
(instruction following). Each domain contributes the same number of training
prompts as its individual GRPO specialist (Table~\ref{tab:indiv-grpo-hypers}).
The \num{2}-domain mix uses the same math and code data without the other two
domains.

\paragraph{Individual GRPO specialists.}
Starting from \texttt{Qwen3-4B-Base}, we train one GRPO specialist per domain.
All four share the hyperparameters in Tables~\ref{tab:shared-hypers}
and~\ref{tab:grpo-hypers}. Table~\ref{tab:indiv-grpo-hypers} lists the
per-domain differences in training data, length caps, batch size, and number
of steps. These specialists are the teachers for multi-teacher \opd{}.

\begin{table}[h]
\centering
\caption{Per-domain deviations for the four individual GRPO
specialists. All other hyperparameters follow
Tables~\ref{tab:shared-hypers}~and~\ref{tab:grpo-hypers}.}
\label{tab:indiv-grpo-hypers}
\small
\begin{tabular}{@{}lllrrr@{}}
\toprule
Domain & Training data & Prompt maximum & Response cap & Batch size & Steps \\
\midrule
Math      & DAPO-Math-17K        & \num{1024} & \num{7168}  & \num{64} & \num{200} \\
Code      & LiveCodeBench v1--v5 & \num{1024} & \num{16384} & \num{32} & \num{80}  \\
IFBench   & IFBench-RLVR         & \num{1024} & \num{10240} & \num{64} & \num{200} \\
MMLU-Pro  & MMLU-Pro \num{5}-shot & \num{4096} & \num{3072}  & \num{64} & \num{200} \\
\bottomrule
\end{tabular}
\end{table}

\paragraph{Joint GRPO baselines.}
We train two joint GRPO baselines on each mix, both with the GRPO
hyperparameters in Table~\ref{tab:grpo-hypers}. \emph{Instance-mix} samples
each batch from the full mixture, so one batch contains prompts from several
domains. It uses a \num{16}K-token response cap and a batch size of \num{64},
and trains for \num{800} optimizer steps on the \num{4}-domain mix and
\num{240} on the \num{2}-domain mix. \emph{Batch-mix} draws each batch from a
single domain with that domain's specialist settings, and trains for the sum
of the specialists' steps: \num{680} on the \num{4}-domain mix and \num{280}
on the \num{2}-domain mix. Both variants see each domain's prompts at least as often as
its specialist does.

\paragraph{Multi-teacher \opd{} (MOPD).}
A fresh \texttt{Qwen3-4B-Base} student is trained with multi-teacher \opd{},
routing each prompt to the GRPO specialist that owns its
domain. The hyperparameters are listed in Table~\ref{tab:opd-hypers}. We test three rollout group sizes $\{8, 4, 1\}$.

\begin{figure}[ht]
\centering
\includegraphics[width=\linewidth]{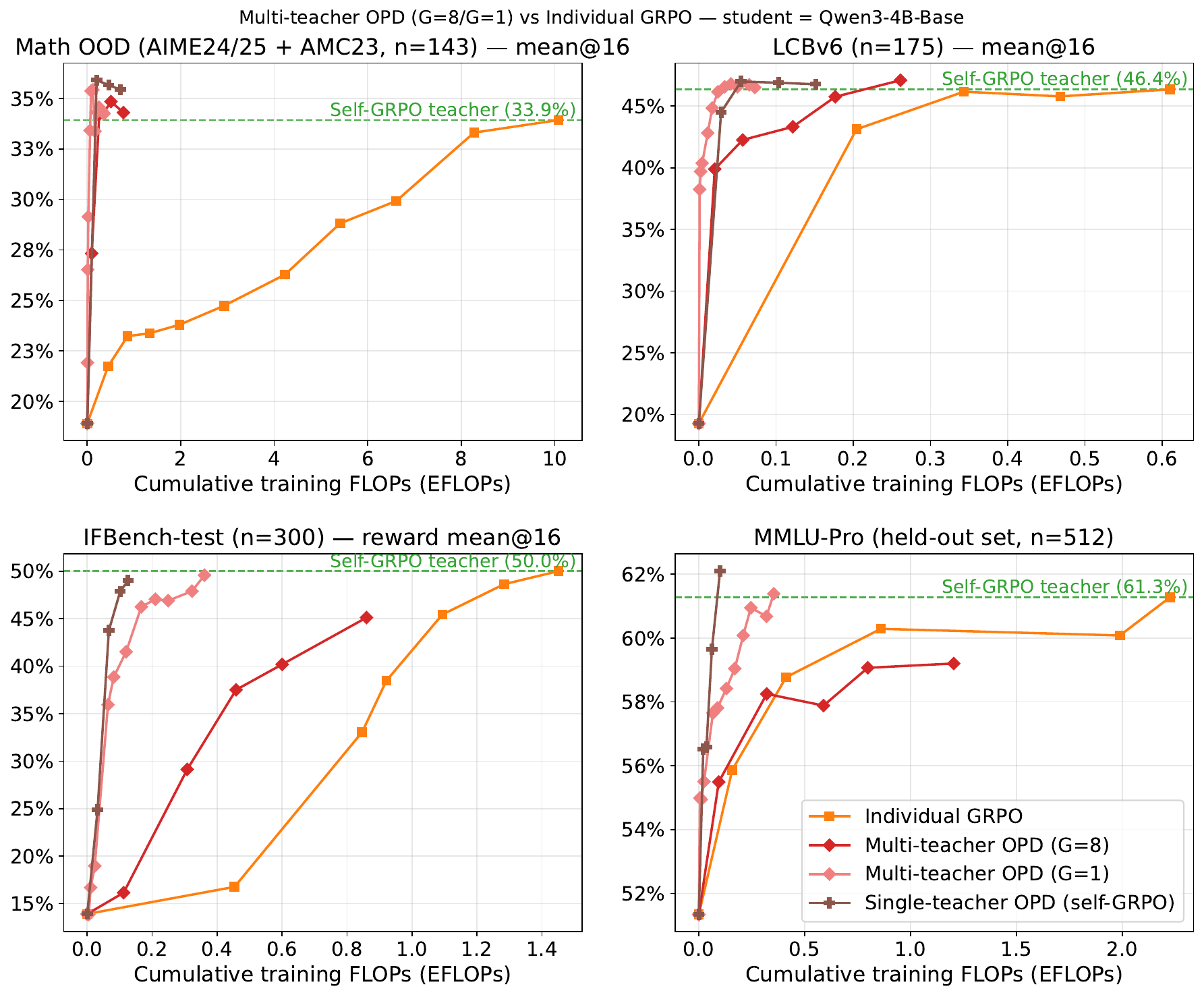}
\caption{\textbf{Group-size ablation for multi-teacher \opd{} on the \num{4}-domain mix.} \texttt{Qwen3-4B-Base} student; the individual GRPO specialists serve as teachers, and their accuracy is the dashed line. FLOPs exclude the cost of training the teachers. At $G{=}1$, multi-teacher \opd{} reaches or comes within \num{0.5} points of the specialists on all four domains within \SI{0.4}{\exa\flop}. At $G{=}8$, it needs more compute and stays below the specialists on IFBench (\pct{45.1} vs.\ \pct{50.0}) and MMLU-Pro (\pct{59.2} vs.\ \pct{61.3}).}

\label{fig:mopd-n1-joint4}
\end{figure}

\begin{figure}[ht]
\centering
\includegraphics[width=\linewidth]{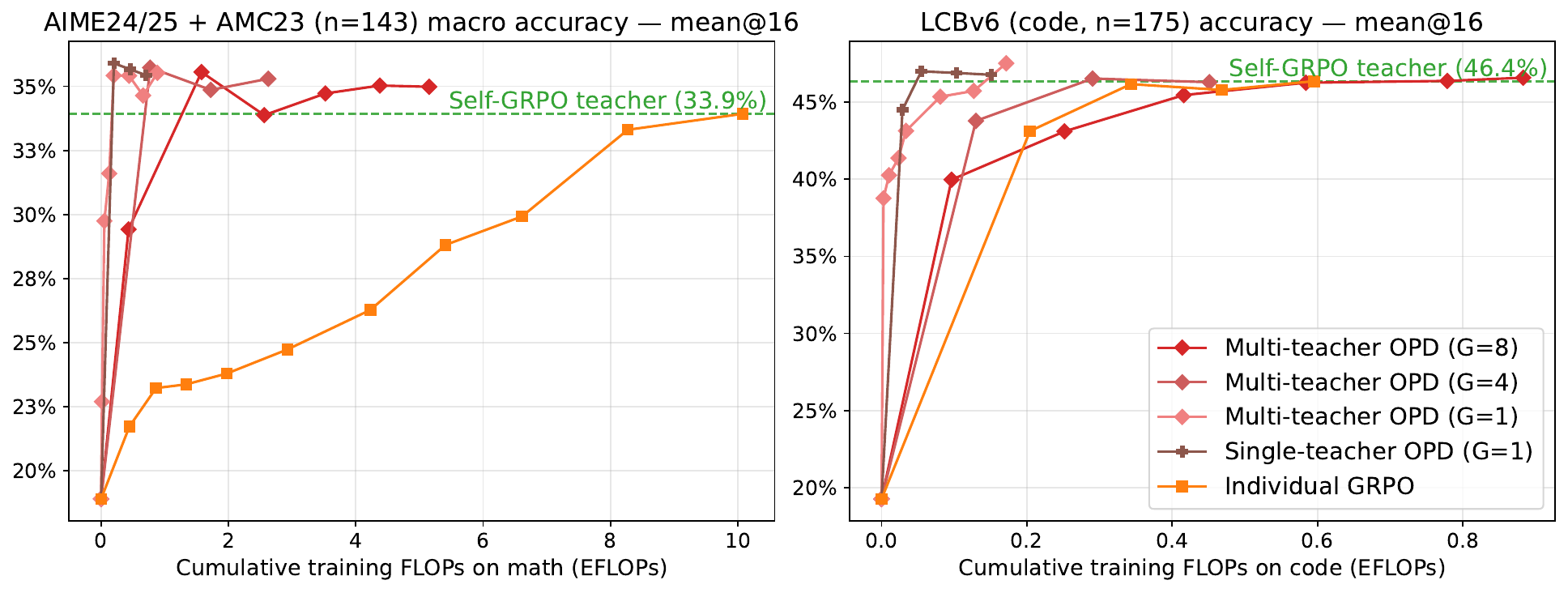}
\caption{\textbf{Group-size ablation for multi-teacher \opd{} on the \num{2}-domain math and code mix.} Same setup as Figure~\ref{fig:mopd-n1-joint4}, with an added $G{=}4$ run. Smaller groups reach the specialists' accuracy with less compute: on math, $G{=}1$, $G{=}4$, and $G{=}8$ need about \num{0.2}, \num{0.8}, and \SI{1.6}{\exa\flop}, against \SI{10}{\exa\flop} for the GRPO specialist; on code, about \num{0.17}, \num{0.3}, and \SI{0.9}{\exa\flop}.}

\label{fig:mopd-n1-mathcode}
\end{figure}

\paragraph{Rollout group size for multi-teacher \opd{}.}
Figures~\ref{fig:mopd-n1-joint4} and~\ref{fig:mopd-n1-mathcode} vary $G$
with the student, teachers, and data mix fixed. On both mixes, $G{=}1$
reaches the specialists' accuracy with substantially less compute than
$G{=}8$. We therefore use
$G{=}1$ in Section~\ref{sec:merging}.

\paragraph{External teacher baseline.}
A single external teacher is applied to the same
\num{2}-domain and \num{4}-domain mix via standard \opd{}. With the \texttt{Qwen3-8B} teacher, training is unstable on math, code, and MMLU-Pro. Same strategy is used for a \texttt{Qwen3-4B} teacher; with the \texttt{Qwen3-4B} teacher, it stays well below the specialists on code and IFBench. The hyperparameters are listed in Table~\ref{tab:opd-hypers} and we keep group size $G=1$. With \texttt{Qwen3-8B} teacher, we keep total training step to be 200. With \texttt{Qwen3-4B}, we keep total training step to be 300.

\begin{figure}[ht]
\centering
\includegraphics[width=\linewidth]{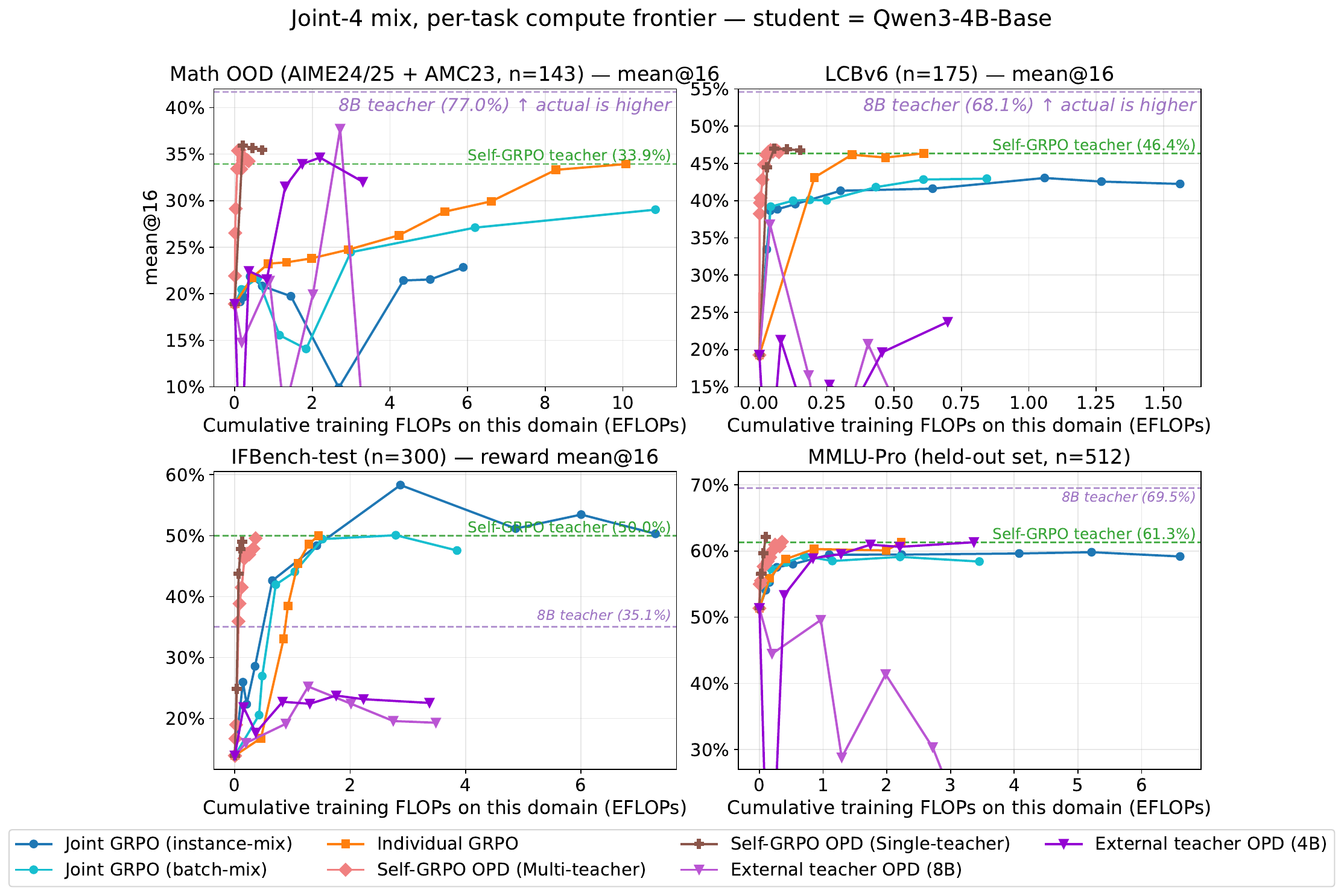}
\caption{\textbf{Per-domain breakdown of the \num{4}-domain mix.} \texttt{Qwen3-4B-Base} student trained on math, code, IFBench, and MMLU-Pro; \texttt{mean@16} on each domain against the training FLOPs spent on that domain. Dashed lines mark the self-GRPO teacher (green) and the external \texttt{Qwen3-8B} teacher (purple; math and code values lie above the plotted range). All \opd{} runs use $G{=}1$. Self-GRPO \opd{}, with one or multiple teachers, reaches or comes within one point of the specialists on every domain within about \SI{1}{\exa\flop}. Joint GRPO trails the specialists on math and code but exceeds them on IFBench. \opd{} with an external teacher stays well below the specialists on code and IFBench.}

\label{fig:joint4-per-domain}
\end{figure}

\begin{figure}[ht]
\centering
\includegraphics[width=\linewidth]{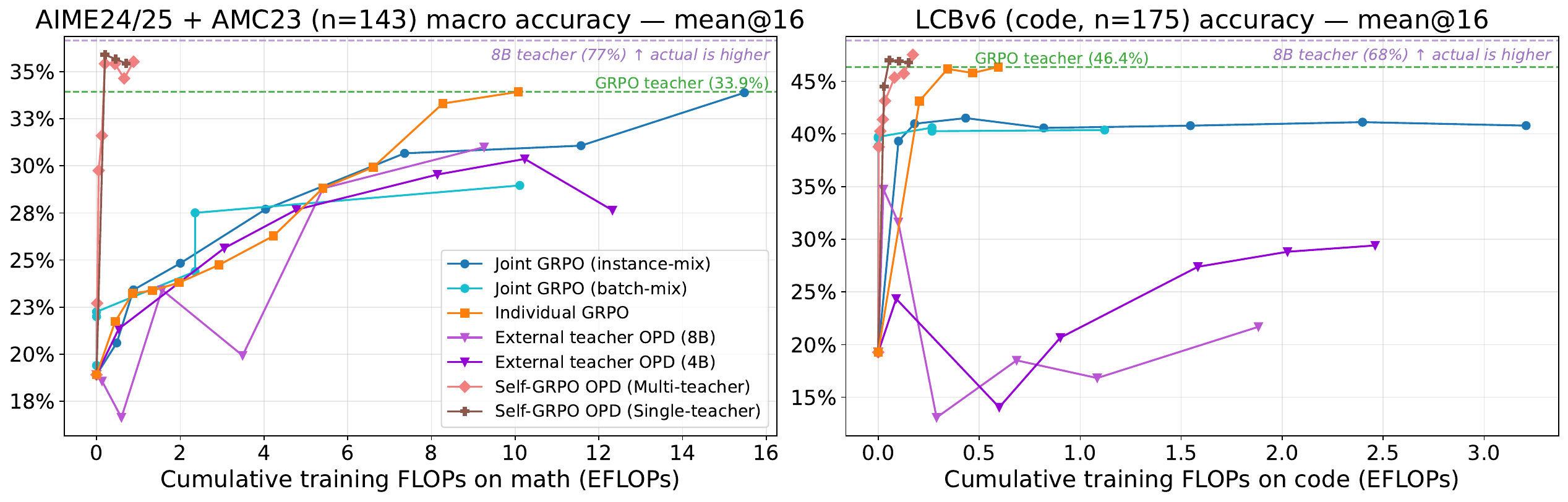}
\caption{\textbf{Per-domain breakdown of the \num{2}-domain math and code mix.} Same methods and conventions as Figure~\ref{fig:joint4-per-domain}. Self-GRPO \opd{} passes the specialists (\pct{33.9} on math, \pct{46.4} on code) within \SI{1}{\exa\flop} on math and \SI{0.2}{\exa\flop} on code. Joint GRPO with instance-level mixing needs about \SI{15}{\exa\flop} to reach the math specialist and stays near \pct{41} on code. \opd{} with an external teacher stays below \pct{32} on math and \pct{30} on code.}

\label{fig:mathcode-per-domain}
\end{figure}

\paragraph{Per-domain results.}
Figures~\ref{fig:joint4-per-domain} and~\ref{fig:mathcode-per-domain} show
per-domain breakdowns for the \num{4}-domain and \num{2}-domain experiments
respectively.

\end{document}